\documentclass[runningheads]{llncs}
\usepackage[skip=5pt]{caption} 
\usepackage[T1]{fontenc}
\usepackage{setspace}
\let\oldbibliography\bibliography
\renewcommand{\bibliography}[1]{%
  \begin{spacing}{0.9} 
  \oldbibliography{#1}%
  \end{spacing}
}
\usepackage{cite}
\usepackage{graphicx}
\usepackage{amsmath, amsfonts, mathtools}
\usepackage{booktabs}
\usepackage{multirow}
\usepackage{subcaption}
\usepackage{xcolor}
\usepackage{algorithm}
\usepackage{algorithmic}
\usepackage{listings}
\usepackage{tablefootnote}
\usepackage{textcomp}
\usepackage{enumitem}
\usepackage{tabularx}
\usepackage{xltabular}

\usepackage{amssymb}

\usepackage{float}

\usepackage[export]{adjustbox} 
\usepackage{siunitx} 
\usepackage[title]{appendix}
\usepackage{hyperref}
\usepackage[capitalize,noabbrev]{cleveref}

\newcommand{\vx}{\mathbf{x}}
\newcommand{\mX}{\mathbf{X}}
\newcommand{\mY}{\mathbf{Y}}

\begin{document}
\title{Channel Dependence, Limited Lookback Windows, and the Simplicity of Datasets: How Biased is Time Series Forecasting?}
\titlerunning{How Biased is Time Series Forecasting?}
%
\author{Ibram Abdelmalak\inst{1,2}\thanks{Authors contributed equally to this research.} \and
Kiran Madhusudhanan\inst{1,2}\protect\footnotemark[1] \and
Jungmin Choi\inst{1,2}\protect\footnotemark[1] \and
Christian Klötergens\inst{1,2} \and
Vijaya Krishna Yalavarthi\inst{1,2} \and
Maximilian Stubbemann\inst{1,2} \and
Lars Schmidt-Thieme\inst{1,2}}

\authorrunning{I. Abdelmalak et al.}

\institute{Institue of Computer Science,University of Hildesheim, Hildesheim, Germany \and
VWFS Data Analytics Research Center (VWFS-DARC), Hildesheim, Germany \\
\email{\{abdelmalak, kiranmadhusud, choi, kloetergens, yalavarthi, stubbemann, schmidt-thieme\}@ismll.de}}
\maketitle              

\begin{abstract}

In Long-term Time Series Forecasting (LTSF), the lookback window is a critical hyperparameter often set arbitrarily, undermining the validity of model evaluations. We argue that the lookback window 
must be tuned on a per-task basis to ensure fair comparisons. Our empirical results show that failing to do so can invert 
performance rankings, particularly when comparing univariate and multivariate methods. Experiments on standard benchmarks 
reposition Channel-Independent (CI) models, such as PatchTST, as state-of-the-art methods. However, we reveal this superior 
performance is largely an artifact of weak inter-channel correlations and simplicity of patterns within these 
specific datasets. Using Granger causality analysis and ODE datasets (with implicit channel correlations), 
we demonstrate that the true strength of multivariate Channel-Dependent (CD) models emerges on datasets with strong, 
inherent cross-channel dependencies, where they significantly outperform CI models. We conclude with four key recommendations for improving TSF research: (i) consider the lookback window as a key hyperparameter to tune, (ii) for standard datasets, examining CI architectures is advantageous, (iii) leverage statistical analysis of datasets to guide the choice between CI and CD architectures, and (iv) prefer CD models in scenarios with limited data.


\end{abstract}

\section{Introduction}
\label{sec:introduction}

\begin{figure}[t]
\centering
\includegraphics[width=\linewidth]{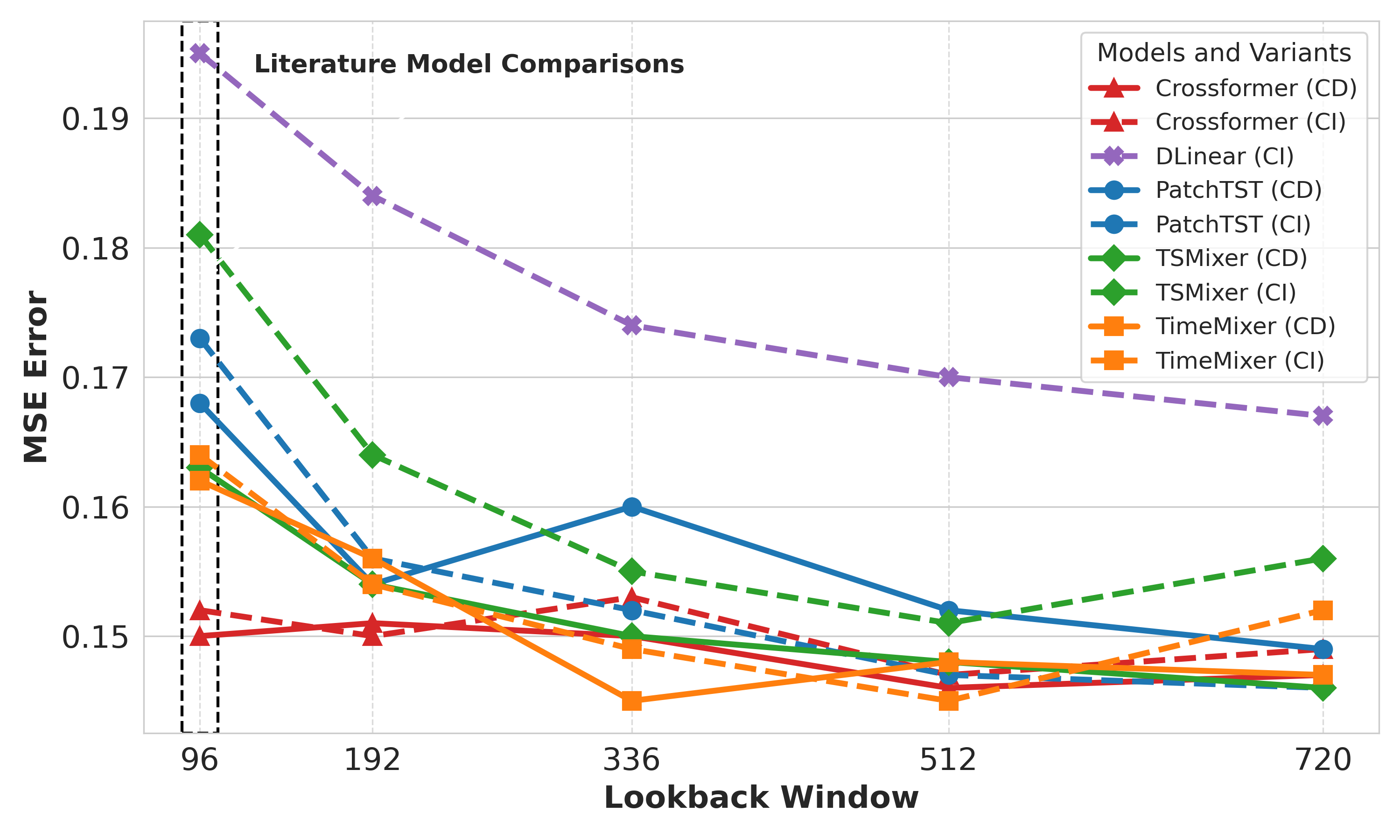}
\caption{Test-error (MSE) on the y-axis for various lookback windows on the x-axis for the
weather dataset and a forecast horizon of 96.
CI denotes the channel-independent variant and CD denotes the channel-dependent variant. The black dotted line shows a lookback window of 96.}
\label{fig:lookback_analysis_motivation}
\end{figure}

Long-term Time Series Forecasting (LTSF) is a core machine learning task that aims to predict future observations by using patterns in historical time-dependent data in various domains such as finance, energy, healthcare, and climate science~\cite{godahewa2021monash}.
Nevertheless, the literature suffers from two significant experimental shortcomings that compromise the
fairness and reliability of model comparisons.

First, LTSF models utilize a fixed-length sequence of past observations, the \emph{lookback window} ($L$), for prediction. Despite its fundamental role, The lookback window is often set arbitrarily (typically $L=96$) across benchmarks without justification~\cite{liu24,zeng23,zhang23,yi23,chen23,chen25,lin24,lin25,huang25}.

While simplifying comparisons, using a fixed lookback window masks a critical reality: the optimal lookback window varies across models and datasets (\Cref{fig:lookback_analysis_motivation}). Fixed-window comparisons may thus obscure which models truly achieve state-of-the-art performance under optimized settings. Unlike \cite{zeng23, tong2025efficiently}, our study explicitly tunes all hyperparameters for every lookback window to ensure reliable evaluation.

The optimal lookback window depends on both dataset properties and model architecture. As shown in \Cref{fig:lookback_analysis_motivation}, Channel-Independent (CI) models typically prefer longer windows (e.g., 336--512) to capture temporal patterns~\cite{nie23,zeng23,chen23}, whereas Channel-Dependent (CD) models often thrive with shorter windows (e.g., 96) by leveraging cross-channel dependencies~\cite{wang24,liu24}. Even within the CD class, benefits vary: Crossformer-CD outperforms TimeMixer-CD at $L=96$, but falls behind at $L=336$. These findings motivate our first claim: \textbf{\textit{Tuning the lookback window as a model-specific hyperparameter is essential for fair evaluation and reliable assessment in time series forecasting.}}

The second limitation concerns the practical relevance of current LTSF benchmarks. Standard multivariate datasets—such as ETT, Electricity, Weather, and Traffic—exhibit very limited dependence among channels. This lack of inter-variable complexity renders these benchmarks overly simplistic, raising a critical question: \textbf{\textit{Does channel dependence truly matter in multivariate LTSF?}}

\begin{table}[t]
    \caption{Frequency of CI and CD wins across models and datasets. 'ODE-6' denotes the 6 ODE datasets with intrinsic channel correlation; 'Std.-7' represents the 7 standard datasets.}
    \centering
    \small 
    \begin{tabular}{l @{\extracolsep{10pt}} cc @{\extracolsep{10pt}} cc}
        \toprule
        \multirow{2}{*}{\textbf{Model}} & \multicolumn{2}{c}{\textbf{ODE-6}} & \multicolumn{2}{c}{\textbf{Std.-7}} \\
        & \textbf{CI} & \textbf{CD} & \textbf{CI} & \textbf{CD} \\
        \midrule
        PatchTST    & \textbf{4} & 2          & \textbf{6} & 1 \\
        TSMixer     & 0          & \textbf{6} & \textbf{7} & 0 \\
        Crossformer & 0          & \textbf{6} & \textbf{4} & 3 \\
        TimeMixer   & \textbf{2} & \textbf{4} & \textbf{4} & 3 \\
        \midrule
        \textbf{Totals} & 6      & \textbf{18} & \textbf{21} & 7 \\
        \bottomrule
    \end{tabular}
    \label{tab:cd_vs_ci}
\end{table}

To address this, we conduct an empirical analysis combining traditional performance metrics with \emph{Granger causality}~\cite{granger}. Our results reveal that multivariate modeling benefits depend on: (1) available historical data and (2) the degree of causal influence among channels. When inter-channel interactions are weak, modeling channel dependence offers limited returns. Consequently, we utilize complex, chaotic Ordinary Differential Equation (ODE) datasets~\cite{gilpin21}, which demonstrate that standard benchmarks fail to capture the cross-channel interactions crucial for multivariate forecasting.

As shown in \Cref{tab:cd_vs_ci}, CD variants consistently outperform CI counterparts on highly coupled ODE datasets. Conversely, standard datasets' simplicity disproportionately favors CI models, limiting the fair evaluation of methods designed for channel dependencies~\cite{nie23,chen23}.

Our main contributions are as follows:
\begin{itemize}[noitemsep,topsep=0pt,leftmargin=*]
\item \textbf{Lookback Evaluation:} We show that tuning the lookback window is a critical, model-specific requirement; neglecting it introduces significant baseline bias.
\item \textbf{Benchmark Simplification:} We reveal that standard LTSF benchmarks lack sufficient channel dependence, unfairly handicapping CD models.
\item \textbf{Causal Analysis:} Using Granger causality and ODE datasets, we prove CD models excel only when complex, verifiable inter-channel interactions are present.
\item \textbf{Guidelines:} We provide actionable recommendations for choosing between CI and CD architectures based on lookback and causal complexity.
\end{itemize}

\section{Related Work}
\label{sec:Related_Works}

LTSF models typically fall into two categories: Channel-Dependent (CD) and Channel-Independent (CI). CI models like DLinear~\cite{zeng23} and PatchTST~\cite{nie23} emphasize temporal dynamics, often outperforming CD models such as TSMixer~\cite{chen23}, iTransformer~\cite{liu24}, and Crossformer~\cite{zhang23}. Recent advancements in CD modeling, including TimeFilter's~\cite{hu2025timefilter} graph filtration and DUET's~\cite{qiu2025duet} hierarchical clustering, attempt to capture complex spatial dependencies, yet the predominantly univariate nature of common benchmarks often masks their potential. Recently, a surge in using Ordinary Differential Equations (ODEs) has generated rigorous, high-causality benchmarks, exemplified by \emph{Physiome}~\cite{physiome2025} and chaotic ODE collections~\cite{gilpin21}. These provide mathematically-defined channel correlations, offering a more robust multivariate testbed than standard datasets.

A critical factor in model evaluation is the lookback window length ($L$). While CI methods typically benefit from $L \in \{336, 512\}$, many studies~\cite{tong2025efficiently,chen25,lin25,huang25} fix $L=96$ for baselines. This practice introduces evaluation bias; for instance, IRPA~\cite{tong2025efficiently} explores variable lookbacks for its own method but maintains $L=96$ for baselines, potentially overstating gains. Although some works acknowledge lookback sensitivity~\cite{wang24TM,nie23}, systematic tuning is often relegated to secondary ablations. While comprehensive benchmarks like TFB~\cite{qiu2024tfb} advocate for fair evaluation across diverse datasets, they do not address the compounding issues of lookback bias and dataset coupling. In contrast, our study systematically tunes $L$ for all models and utilizes ODE-based evaluations to ensure architectural comparisons are based on truly optimal configurations.

\section{Problem Statement}
\label{sec:problem-statement}

To train forecasting models effectively, especially when $T$ is large, we partition the long sequence into multiple shorter subsequences using a sliding window approach. 
Each subsequence is split into an input segment of length $L$ (representing the \emph{lookback window}) and a target segment of length $H$ (the \emph{forecast horizon}). 
A forecasting dataset thus consists of pairs $(\mX, \mY) \in \mathbb{R}^{C \times L} \times \mathbb{R}^{C \times H}$, where $\mX = (\vx^{t}, \dots, \vx^{t+L-1})$ is the \emph{forecasting query} and $\mY = (\vx^{t+L}, \dots, \vx^{t+L+H-1})$ is the corresponding \emph{forecasting answer}.

The goal of the time-series forecasting task is to learn a model $\hat{\mY}: \mathbb{R}^{C \times L} \rightarrow \mathbb{R}^{C \times H}$ using a dataset $\mathcal{D}_{\text{train}} \subseteq \mathbb{R}^{C \times L} \times \mathbb{R}^{C \times H}$, constructed from many such windows sampled from the full sequence. 
The dataset is assumed to be drawn i.i.d.\ from an underlying distribution $p$. The objective is to minimize the expected loss, $\mathbb{E}_{(\mX,\mY) \sim p}\, \ell(\mY, \hat{\mY}(\mX))$, where $\ell$ is typically the mean squared error (MSE). While the underlying time series is intrinsically dependent, this formulation treats the windowed pairs $(\mX, \mY)$ as realizations of a stationary process for the purpose of supervised optimization. This allows the model $\hat{\mY}$ to explicitly focus on capturing the temporal and inter-channel dependencies within each local windowed segment.

\section{Background}
\label{sec:background}

\paragraph{\textbf{Fixing Lookback Windows:}}
The problem formulation in \Cref{sec:problem-statement} does not strictly require the forecasting query $\mX$ to use a fixed lookback length $L$. However, most recent LTSF models adopt a fixed window (commonly $L=96$) during training, primarily for implementation convenience or adherence to benchmarking standards. As argued previously, the lookback window should be treated as a model-specific hyperparameter, integral to the model design.

\paragraph{\textbf{Channel Independence:}}
Recent studies~\cite{nie23,chen23} demonstrate that on widely used benchmarks, CI models often suffice for accurate forecasting. This suggests that cross-channel interactions may be unnecessary for certain tasks. Formally, a model $\hat{\mY}: \mathbb{R}^{C \times L} \to \mathbb{R}^{C \times H}$ is \emph{channel-independent} if there exists a univariate predictor $f: \mathbb{R}^L \to \mathbb{R}^H$ such that for each channel $c \in \{1, \dots, C\}$:
\begin{equation}
    \hat{\mY}(\mX)_{c} = f(\vx_{c, :}) \quad \forall c \in \{1, \dots, C\}.
\end{equation}

\paragraph{Granger Causality.}
Granger causality~\cite{granger} evaluates whether a channel $c_2$ provides predictive information for $c_1$ beyond $c_1$'s own history. We compare the \emph{univariate sum of squared residuals} ($SSR_u$) from a restricted model (CI assumption) and the \emph{multivariate} ($SSR_{mv}$) from an unrestricted model (CD assumption) incorporating $c_2$. Using $x_{c,t}$ to denote the observation of channel $c$ at time $t$, the residuals are:
\begin{align}
    SSR_u &= \sum_{t=L+1}^{L+H} \bigg(x_{c_1, t} - a_0 - \sum_{i=1}^L a_i x_{c_1, t-i} \bigg)^2, \\
    SSR_{mv} &= \sum_{t=L+1}^{L+H} \bigg(x_{c_1, t} - a_0 - \sum_{i=1}^L a_i x_{c_1, t-i} - \sum_{i=1}^L b_i x_{c_2, t-i} \bigg)^2.
\end{align}
where $a_0$ is the intercept, and $a_i, b_i$ are learned coefficients for the lagged values of $c_1$ and $c_2$ respectively. 

The $F$-statistic measures the significance of the error reduction:
\begin{equation}
    F = \frac{(SSR_u - SSR_{mv})/L}{SSR_{mv}/(N - k_{mv})}, 
    \label{eq:granger_fstat}
\end{equation}
where $L$ is the number of added parameters (lags of $c_2$), $N$ the number of observations, and $k_{mv}$ the total parameters in the multivariate model. A high $F$-statistic indicates that $c_2$ Granger-causes $c_1$.

\section{General Experimental Setup}
\label{sec:experimental setup}

\subsection{Models}
\label{subsec:models}

We evaluate several baselines: MLP-based (\textbf{TSMixer}~\cite{chen23}, \textbf{TimeMixer}~\cite{wang24}), attention-based (\textbf{PatchTST}~\cite{nie23}, \textbf{iTransformer}~\cite{liu24}, \textbf{Crossformer}~\cite{zhang23}), and \textbf{DLinear}~\cite{zeng23}. We use \textbf{DLinear-CI} and \textbf{iTransformer-CD} to prevent architectural overlap with TSMixer; others are tested in both variants. Details are in our repository.\footnote{\url{https://github.com/IbramMedhat/How-Biased-is-Time-Series-Forecasting-}} and the appendix.

\subsection{Datasets}
\label{subsec:Dataset}
As noted in the introduction, we evaluate two main dataset categories to examine how increasing complexity particularly channel correlation impacts model performance. 

\paragraph{\textbf{The Standard Datasets:}} 
The first category includes widely used benchmarks from recent state-of-the-art studies and surveys~\cite{nie23,liu24,wang24,chen23,zhang23,zeng23}: \emph{Weather}, \emph{Electricity}, \emph{Traffic}, and the \emph{Electricity Transformer Temperature (ETT)} datasets. 
These benchmarks gained popularity following their adoption in the Informer paper~\cite{informer}.

\begin{figure}[t]
	\centering
	\includegraphics[width=\linewidth]{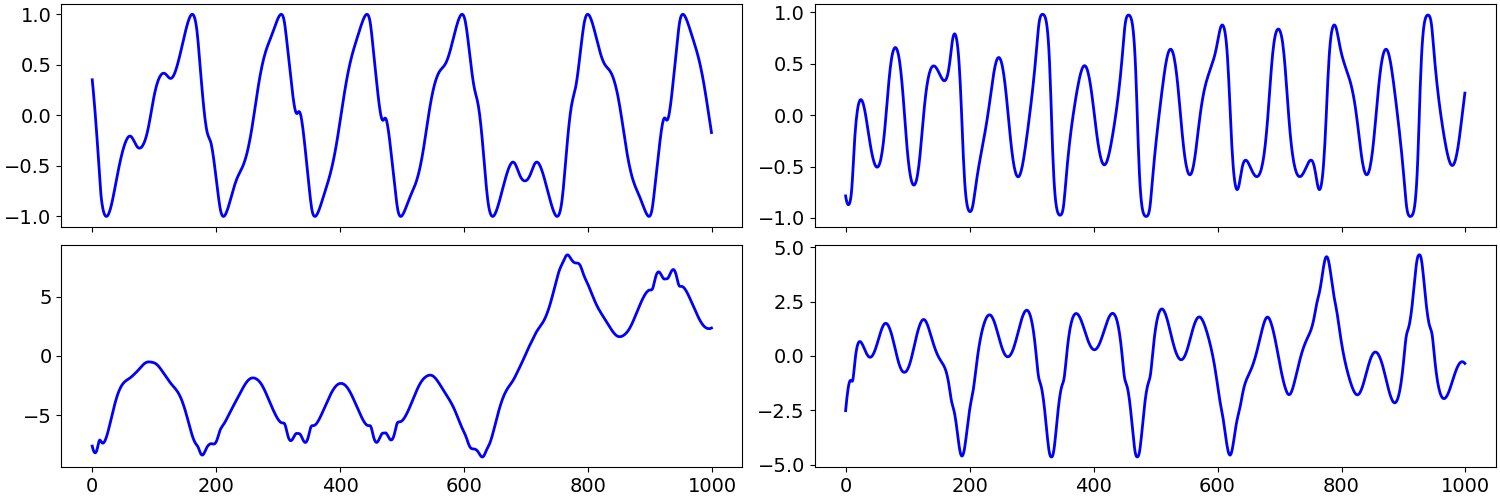}
	\caption{Time series visualization of the Double Pendulum ODE system. Notably, the bottom two channels (angular velocities of the two pendulums) exhibit synchronized behavior in the last 200 time steps, demonstrating time-varying inter-channel correlation.}
	\label{fig:double-pendulum}
\end{figure}

\paragraph{\textbf{The Chaotic ODE Benchmark:}} 
As a second dataset collection, we consider the benchmark proposed by~\cite{gilpin21}. This benchmark was introduced as a time-series collection that features complex time series dynamics with clear real world data properties, providing a challenging and realistic setting for CD time-series forecasting. A demonstration for the \emph{DoublePendulum} dataset and how it simulates a recurrent patterns across different channels can be seen in~\Cref{fig:double-pendulum}. This figure particularly emphasizes periods of synchronization in the angular velocities of the two pendulums during the final 200 timesteps, indicative of inter-channel correlation. 

We select a subset of datasets from this benchmark based on two distinctive criteria: 
\textit{(i)} The datasets follow complex trajectories that challenge most time series models and
\textit{(ii)} they are derived directly from the underlying ODEs, ensuring a clear mathematical correlation between variables. For criterion (i), we use the \textit{Largest Lyapunov exponent}~\cite{gilpin21} to quantify dataset complexity, selecting datasets with high values to ensure forecasting remains non-trivial task that cannot be solved easily based on one channel information. To illustrate criterion (ii), we show below a simplified form of the Double Pendulum system~\cite{elbori2017}, assuming equal masses and rod lengths. The coupling between angular velocities \(\frac{d \theta_1(t)}{dt}, \frac{d \theta_2(t)}{dt}\) and angles \(\theta_1(t), \theta_2(t)\) at time \(t\) from \Cref{eq:theta1,eq:theta2} is evident, with parameters \(g\) (gravity) and \(l\) (rod length) incorporated. Such systems are essential in modeling real-world physics processes, including robotics~\cite{jadlovska2012}.

\begin{equation}
	\label{eq:theta1}
	\frac{d^2 \theta_1(t)}{dt^2} = -\frac{2g}{l} \theta_1(t) + \frac{g}{l} \theta_2(t)
\end{equation}

\begin{equation}
	\label{eq:theta2}
	\frac{d^2 \theta_2(t)}{dt^2} = \frac{2g}{l} \theta_1(t) - \frac{2g}{l} \theta_2(t)
\end{equation}

\subsection{Hyperparameter Tuning}
\label{subsec:HP_tuning}

We ensure fair evaluation via Bayesian optimization (Optuna~\cite{akiba19}) over 20 trials per model--dataset--horizon. The lookback window $L$ is tuned across $\{96, \allowbreak 192, \allowbreak 336, \allowbreak 512, \allowbreak 720\}$, with the best variant selected via validation MSE. We prioritize MSE as the primary metric to remain consistent with LTSF benchmarks and to penalize large outliers common in complex series. Crucially, we address data-loading biases noted in \cite{qiu2024tfb} by setting \texttt{drop\_last=False} during testing. This ensures no trailing points are discarded, providing a complete, unbiased evaluation across the entire distribution and confirming that lookback-CD interactions are not artifacts of data truncation.

\subsection{Granger Causality}

We perform Granger causality analysis following~\cite{granger} with two preprocessing steps: (i) assessing stationarity and applying iterative differencing where necessary, and (ii) removing one channel from any pair with a Pearson correlation above 0.95 to reduce redundancy. Following the $F$-test (\Cref{sec:background}), we report the average $F$-score and the percentage of pairs rejecting the null hypothesis.

\section{Results}
\label{sec:results}

\subsection{Evaluation of Standard Datasets}
\label{subsec:evaluation-standard-datasets}

We reproduce existing time-series forecasting models under our evaluation setup, performing extensive hyperparameter tuning (including lookback window) and averaging results over five random seeds. Reported scores in \Cref{tab:standard_mse} are averaged across horizons 96, 192, 336, and 720. 

PatchTST remains the state-of-the-art model in our fair evaluation. Its CI variant achieves the best performance on three of seven datasets and ranks second or third on another three. Only on the Weather dataset does it fall outside the top three, likely due to stronger inter-channel dependencies indicated by Granger causality. Overall, PatchTST attains an average rank of 2.14, a notable margin (0.57) ahead of the second-best model, CI TSMixer.
Linear models continue to perform competitively. DLinear, for instance, achieves an average rank of 5.43 out of 10 across the six standard datasets and ranks second on ETTh1, slightly behind PatchTST.

\begin{table}[t]
    \centering
    \caption{Mean MSE across four horizons. \textbf{Bold}: best per model; \textcolor{blue}{blue}: overall best. (*): sourced from \cite{wang24TM}. OOM: Out-of-memory.}
    \label{tab:standard_mse}
    \scriptsize 
    \setlength{\tabcolsep}{2.5pt} 
    \begin{tabular}{l cc cc cc c c cc}
        \toprule
        \textbf{Dataset} & \multicolumn{2}{c}{\textbf{PatchTST}} & \multicolumn{2}{c}{\textbf{TSMixer}} & \multicolumn{2}{c}{\textbf{Crossfor.}} & \textbf{DLin.} & \textbf{iTrans.} & \multicolumn{2}{c}{\textbf{TimeMixer}} \\
        & \textbf{CI} & \textbf{CD} & \textbf{CI} & \textbf{CD} & \textbf{CI} & \textbf{CD} & \textbf{CI} & \textbf{CD} & \textbf{CI} & \textbf{CD} \\ 
        \midrule
        ETTh1 & {\color{blue}\textbf{0.422}} & 0.437 & \textbf{0.438} & 0.453 & 0.477 & \textbf{0.456} & 0.423 & 0.511 & \textbf{0.434} & 0.489 \\ 
        Weather & 0.245 & \textbf{0.233} & {\color{blue}\textbf{0.230}} & 0.241 & \textbf{0.236} & 0.296 & 0.289 & 0.253 & 0.247 & \textbf{0.246} \\ 
        Electricity & {\color{blue}\textbf{0.159}} & 0.170 & \textbf{0.162} & 0.170 & \textbf{0.166} & 0.188 & 0.162 & 0.220 & \textbf{0.173} & 0.260 \\ 
        ETTh2 & \textbf{0.365} & 0.382 & \textbf{0.378} & 0.390 & 0.741 & \textbf{0.691} & 0.507 & {\color{blue}\textbf{0.363}} & \textbf{0.371} & 0.378 \\ 
        ETTm1 & \textbf{0.356} & 0.371 & \textbf{0.356} & 0.370 & \textbf{0.393} & 0.426 & 0.359 & 0.374 & {\color{blue}\textbf{0.355}} & 0.408 \\ 
        ETTm2 & \textbf{0.258} & 0.259 & {\color{blue}\textbf{0.257}} & 0.273 & \textbf{0.433} & 0.485 & 0.289 & {\color{blue}\textbf{0.257}} & \textbf{0.275} & 0.342 \\ 
        Traffic & {\color{blue}\textbf{0.388}} & OOM & \textbf{0.407} & 0.417 & OOM & 0.542 & 0.426 & OOM & OOM & {\color{blue}\textbf{0.388}} \\
        \midrule
        \textbf{Wins} & {\color{blue}\textbf{3}} & 0 & 2 & 0 & 0 & 0 & 0 & 2 & 1 & 0 \\ 
        \textbf{Avg. Rank} & {\color{blue}\textbf{2.14}} & 5.43 & 2.71 & 5.00 & 7.29 & 8.14 & 5.43 & 5.86 & 5.29 & 6.57 \\
        \bottomrule
    \end{tabular}
\end{table}

\subsection{Granger Causality for Evaluation of Channel Correlations}
\label{subsec:granger-causality}

Building on previous findings that multivariate models do not consistently outperform univariate models on standard datasets, we perform a statistical analysis of channel correlations across both \emph{standard datasets} and \emph{ODE datasets} to determine where inter-channel dependencies exist and multivariate models might be necessary.

We compare both benchmarks using Granger causality tests with lags (lookback windows) of 30, 96, and 192. 
F-scores and percentages of channel pairs showing causality (i.e. $H_{0}$ rejected) are reported in~\Cref{tab:fscore_results}, with higher values indicating stronger correlations. \emph{The ODE datasets} show significantly stronger inter-channel dependencies with higher F-scores and $H_{0}$ rejection at all lags, while standard datasets have much weaker correlations, with some F-scores dropping below 1.0 at higher lags. This supports and explains earlier observations where univariate models often match or outperform multivariate ones on standard datasets due to weak coupling.

Increasing lag generally reduces correlations in both dataset types, especially for standard datasets, suggesting longer lookback windows favor univariate models that do not use cross-channel signals. In contrast, \emph{ODE datasets} maintain strong correlations even at longer lags (F-scores of 88.5 at lag 96 and 35.29 at lag 192), supporting their use as testbeds for models handling strong cross-channel dynamics.

\begin{table}[t]
    \centering
    \caption{Average F-scores and $H_{0}$ rejection rates ($p=0.05$). Bold indicates the highest correlation at each lag.}
    \label{tab:fscore_results}
    \small 
    \setlength{\tabcolsep}{10pt} 
    \begin{tabular}{l cc}
        \toprule
        \textbf{Dataset Type (lag)} & \textbf{Average F-score} & \textbf{$H_{0}$ rejected (\%)} \\
        \midrule
        ODEs (30)           & \textbf{335.21} & \textbf{73\%} \\
        Standard (30)       & 2.46            & 61\% \\
        \addlinespace[5pt] 
        ODEs (96)           & \textbf{88.50}  & \textbf{68\%} \\
        Standard (96)       & 1.40            & 45\% \\
        \addlinespace[5pt]
        ODEs (192)          & \textbf{35.29}  & \textbf{63\%} \\
        Standard (192)      & 1.23            & 27\% \\
        \bottomrule
    \end{tabular}
\end{table}

\subsection{Evaluation on the ODE Benchmark}
\label{subsec:chaotic-ode-evaluation}

As discussed in~\Cref{subsec:evaluation-standard-datasets}, DLinear remains a competitive baseline on standard datasets, with CI models outperforming CD counterparts. However, on the \emph{ODE datasets}, model performance differs significantly. CD variants consistently outperform CI models in 18 out of 24 experiments as shown in~\Cref{tab:ode_results}, underscoring the importance of modeling inter-channel dependencies in datasets governed by channel-interactions.

Interestingly, Linear models perform poorly on \emph{ODE datasets}. DLinear ranks worst in 5 of 6 \emph{ODE datasets}, with an average rank of 9.83. Compared to Crossformer, DLinear's performance drops between 45\% and 118\%, with a median decline of 58.3\%.

These results challenge the notion that linear models compete well with deep architectures, highlighting the need to evaluate models on datasets with varying structural characteristics. For data with intrinsic inter-channel correlations, like ODE systems, model architectures must explicitly account for these dependencies to achieve strong performance. We note that while Granger causality tests provide a rigorous measure of linear statistical predictability between channels, they serve as a proxy for inter-channel dependency in this context rather than implying structural or physical causality in the strong sense.

\begin{table*}[t]
    \centering
    \caption{Mean MSE on the \textbf{ODE Benchmark}. \textbf{Bold}: best per model; \textcolor{blue}{blue}: overall best. Superscript references statistical details in the extended version.}
    \label{tab:ode_results}
    \scriptsize
    \setlength{\tabcolsep}{3.2pt} 
    \begin{tabular}{l cc cc cc c c cc}
        \toprule
        \textbf{Dataset} & \multicolumn{2}{c}{\textbf{PatchTST}} & \multicolumn{2}{c}{\textbf{TSMixer}} & \multicolumn{2}{c}{\textbf{Crossfor.}} & \textbf{DLin.} & \textbf{iTrans.} & \multicolumn{2}{c}{\textbf{TimeMixer}} \\
        & \textbf{CI} & \textbf{CD} & \textbf{CI} & \textbf{CD} & \textbf{CI} & \textbf{CD} & \textbf{CI} & \textbf{CD} & \textbf{CI} & \textbf{CD} \\ 
        \midrule
        Lorenz & \textbf{0.839} & 0.841 & 0.880 & \textbf{0.868} & 0.667 & {\color[HTML]{0000FF} \textbf{0.643}} & 0.934 & 0.675 & 0.764 & \textbf{0.673} \\
        BlinkingRotlet & \textbf{0.424} & 0.426 & 0.580 & \textbf{0.487} & 0.340 & {\color[HTML]{0000FF} \textbf{0.311}} & 0.522 & 0.426 & \textbf{0.433} & 0.520 \\
        CellCycle & \textbf{0.635} & 0.667 & 0.792 & \textbf{0.771} & 0.429 & {\color[HTML]{0000FF} \textbf{0.428}} & 0.935 & 0.808 & 0.679 & \textbf{0.556} \\
        DoublePendulum & \textbf{0.653} & 0.668 & 0.768 & \textbf{0.737} & 0.553 & \textbf{0.541} & 0.805 & 0.656 & 0.595 & {\color[HTML]{0000FF} \textbf{0.529}} \\
        Hopfield & 0.420 & \textbf{0.346} & 0.507 & \textbf{0.435} & 0.335 & \textbf{0.316} & 0.690 & 0.300 & 0.622 & {\color[HTML]{0000FF} \textbf{0.245}} \\
        LorenzCoupled & 0.881 & \textbf{0.866} & 0.950 & \textbf{0.900} & 0.700 & {\color[HTML]{0000FF} \textbf{0.666}} & 0.963 & 0.857 & \textbf{0.788} & 0.832 \\
        \midrule
        \textbf{Wins} & 0 & 0 & 0 & 0 & 0 & {\color[HTML]{0000FF} \textbf{4}} & 0 & 0 & 0 & 2 \\
        \textbf{Avg. Rank} & 5.17 & 5.67 & 8.83 & 7.50 & 2.50 & {\color[HTML]{0000FF} \textbf{1.50}} & 9.83 & 5.17 & 5.50 & 3.30 \\
        \bottomrule
    \end{tabular}
\end{table*}

\subsection{Lookback Window Analysis}
\label{subsec:lw_analysis}

\begin{figure*}[t]
	\centering
	\begin{subfigure}[b]{0.32\linewidth}
		\centering
		\includegraphics[width=\linewidth]{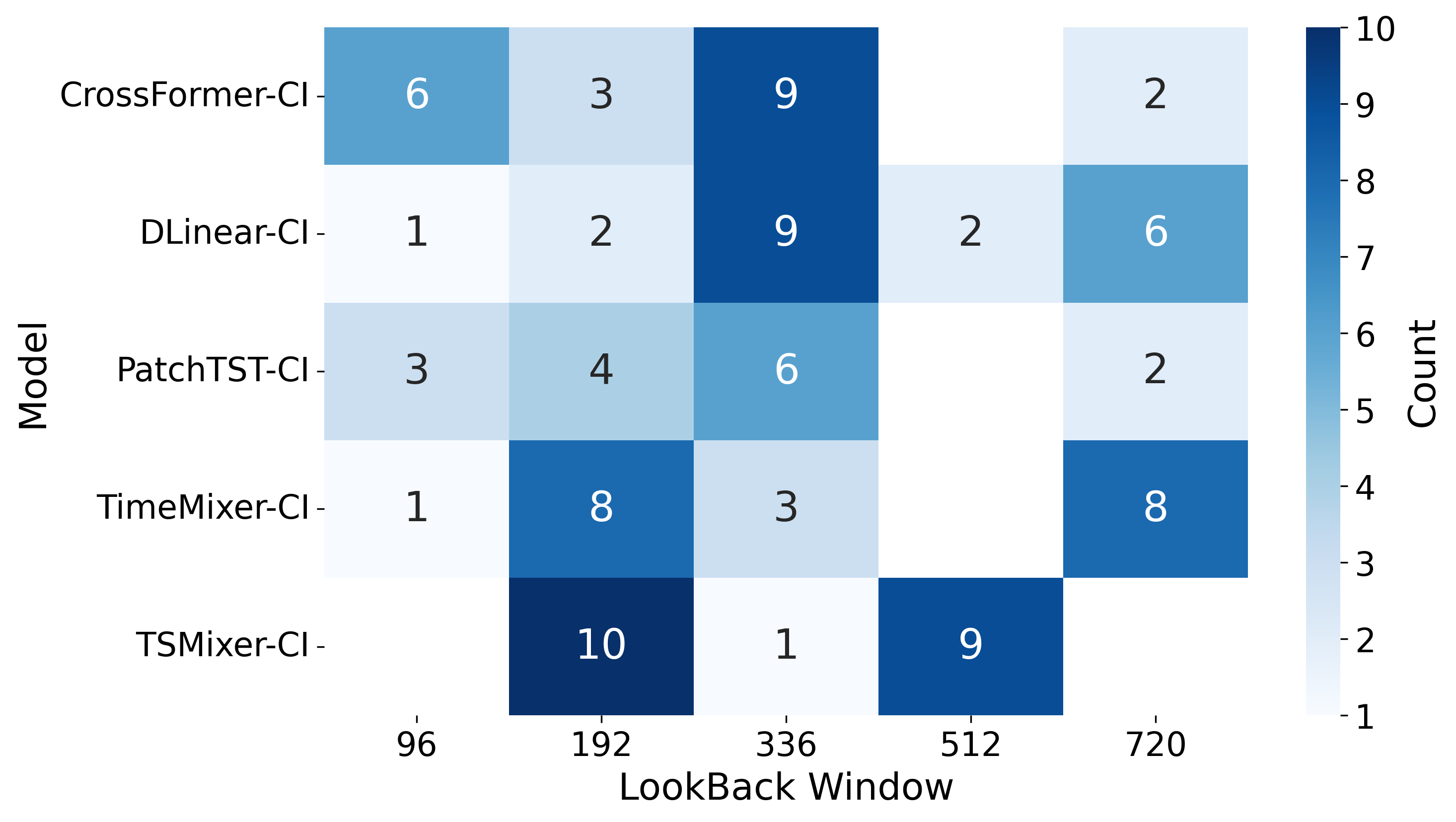}
		\subcaption{CI models\label{fig:lw-heatmap-a}}
	\end{subfigure}
	\hfill
	\begin{subfigure}[b]{0.32\linewidth}
		\centering
		\includegraphics[width=\linewidth]{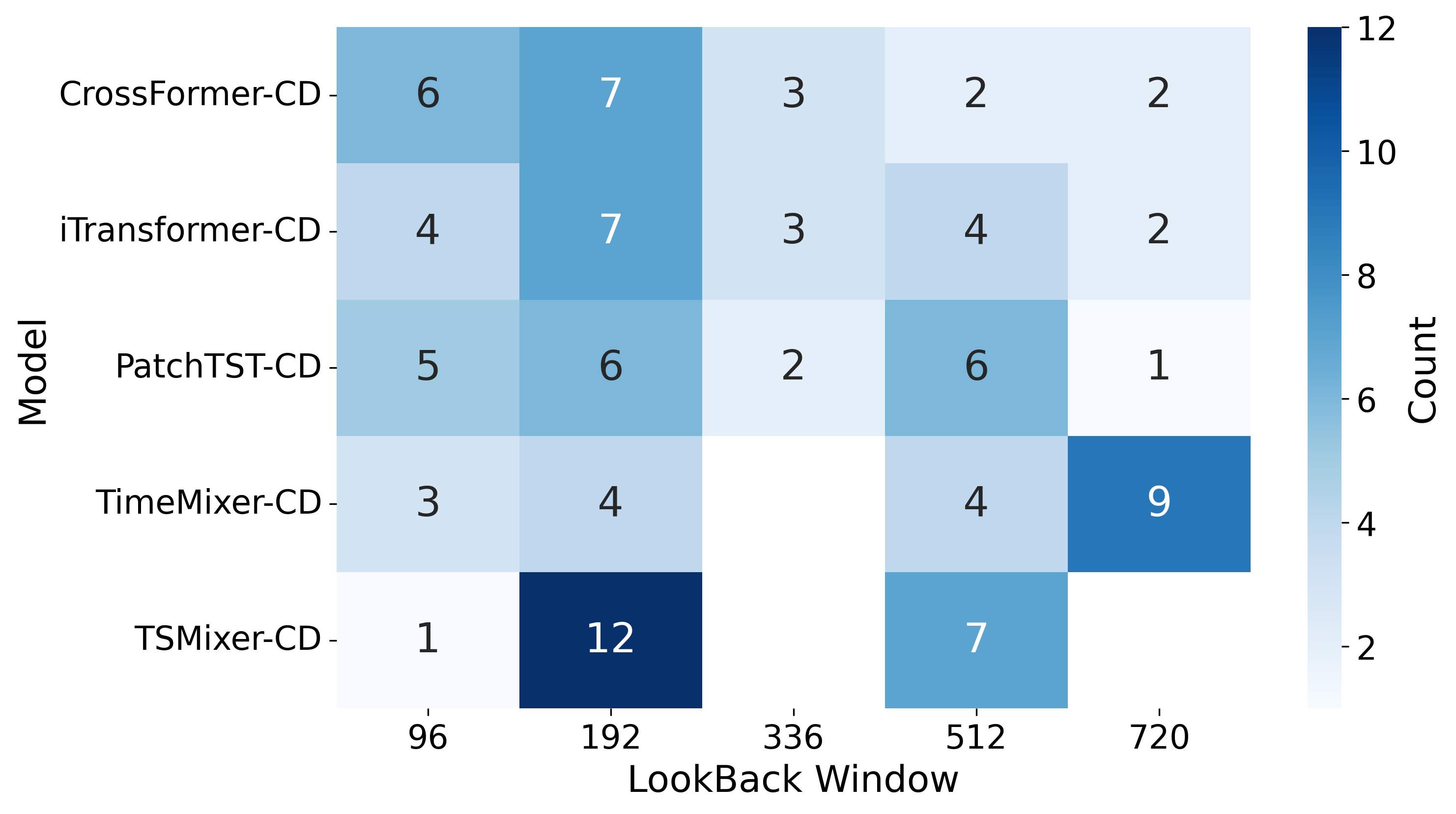}
		\subcaption{CD models\label{fig:lw-heatmap-b}}
	\end{subfigure}
	\hfill
	\begin{subfigure}[b]{0.32\linewidth}
		\centering
		\includegraphics[width=\linewidth]{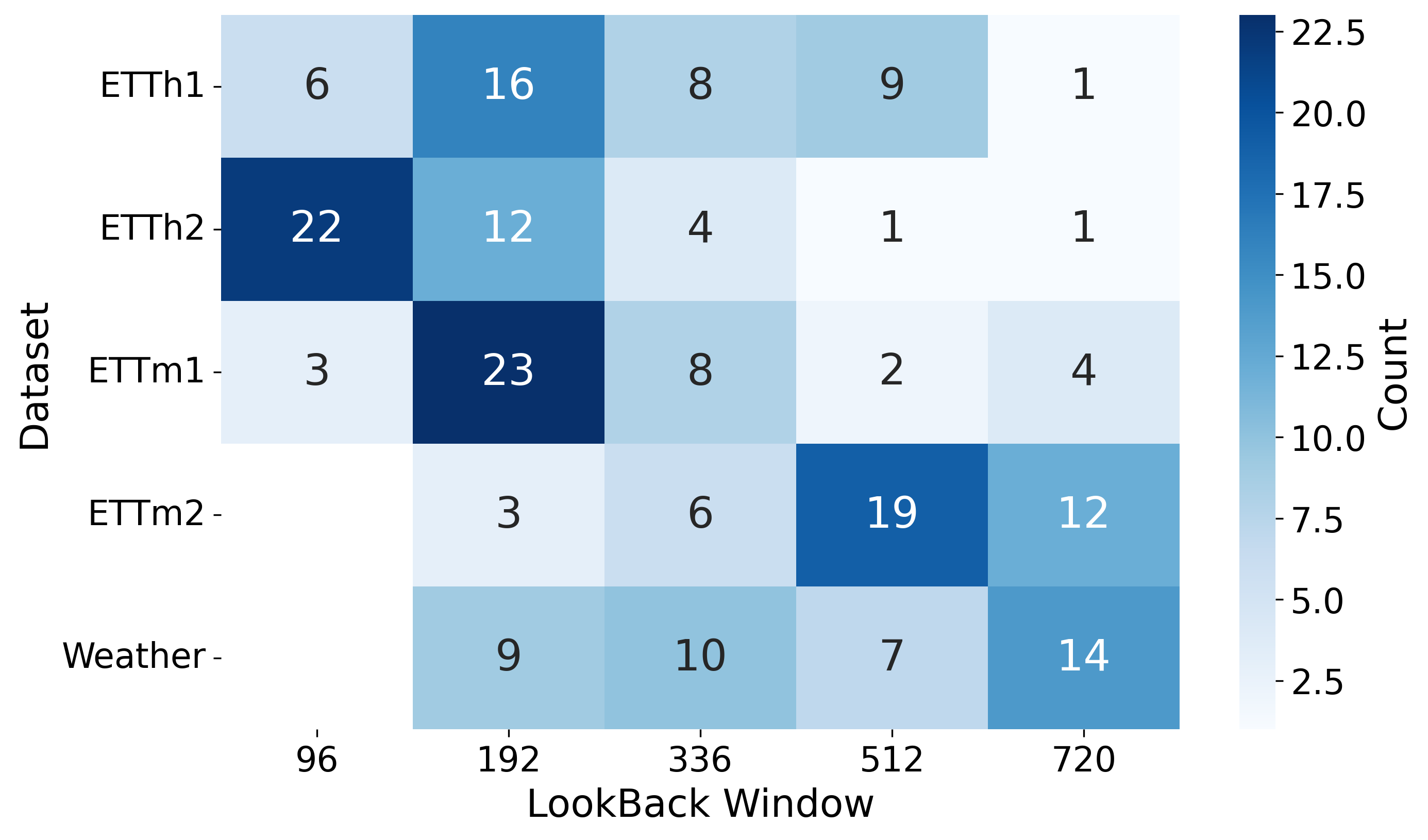}
		\subcaption{Datasets\label{fig:lw-heatmap-c}}
	\end{subfigure}
	\caption{Frequencies of best-performing lookback windows grouped by: (a) CI models, (b) CD models, and (c) datasets.}

	\label{fig:lw-heatmap}
	
\end{figure*}

This section analyzes the distribution of optimal lookback window values by dataset and model, showing significant variation. \Cref{fig:lw-heatmap} shows that CI models favor longer windows (336 or 720), with 96 chosen only 11\% of the time. In contrast, CD models often benefit from shorter windows (96 or 192) in 55\% of cases. 

Theoretically, this divergence suggests a trade-off between temporal depth and inter-channel noise. CI models, treating each channel in isolation, require longer historical contexts to compensate for the lack of cross-variable information. Conversely, CD models extract relevant signals from other channels, allowing them to achieve high accuracy with shorter temporal windows. Furthermore, for CD models, excessively long windows may introduce redundant inter-channel noise, making it harder for the model to distinguish true dependencies from spurious correlations across variables.

Comparing across datasets, hourly datasets (\emph{ETTh1} and \emph{ETTh2}) typically prefer shorter windows around 96, covering approximately four days, while 15-minute interval datasets (\emph{ETTm1} and \emph{ETTm2}) favor longer windows (greater than 336) to capture equivalent temporal spans. This highlights that while temporal span is a physical requirement of the data, the optimal window length is a function of the model's capacity to process spatial (channel) versus temporal information.

\section{Conclusion and Recommendations}
\label{sec:conclusion-recommendation}

Our study establishes a fair evaluation framework for multivariate time series forecasting by tuning lookback windows, analyzing inter-channel correlations, and assessing diverse datasets. From this, we conclude and recommend the following:

\begin{enumerate}
	\item \textbf{Lookback window tuning is essential.} 
	Fixing the lookback window  (commonly at 96) biases results against univariate models and masks true performance differences. 
	\emph{Recommendation:} Always allow models to optimize lookback windows to ensure balanced and meaningful comparisons.
	
	\item \textbf{CD modeling is unnecessary for standard datasets.} 
	CI models generally outperform CD models on \emph{standard datasets} by a margin of 21 to 7, suggesting that mixing information across channels can negatively impact performance. 
\emph{Recommendation:} For \emph{standard datasets}, CI models should be preferred as they typically provide better results.

	\item \textbf{Dataset characteristics dictate model design.} 
	Datasets with strong inter-channel correlations, such as \emph{ODE datasets}, require CD models capturing these dependencies, supported by superior CD performance and statistical tests. 
	\emph{Recommendation:} Pre-evaluate channel correlations using statistical tests to select appropriate model architectures and use CD models on datasets exhibiting strong inter-channel dynamics.
	
	\item \textbf{Lookback window length interacts with channel dependency.} 
	CD models benefit from shorter lookback windows, while CI models may favor longer ones, reflecting differing temporal dependency needs. 
	\emph{Recommendation:} Tailor lookback window length jointly with model type and dataset temporal resolution to maximize forecasting accuracy.

\end{enumerate}

\paragraph{\textbf{Limitations and Future Work:}}
While our framework ensures a fair comparison, it is subject to several limitations. First, our evaluation relies on a predefined discrete set of lookback windows $\{96, 192, 336, 512, 720\}$; while this aligns with standard practice in LTSF literature, it remains a pragmatic subset of all possible temporal spans. Second, our primary reliance on MSE, while standard for penalizing large outliers, may not fully capture model performance in contexts where distributional or scale-invariant metrics are prioritized. 

Furthermore, while this study leverages chaotic ODE simulations to provide a mathematically verifiable ground truth for channel dependency, future work will explore whether these advantages transfer to highly stochastic real-world domains—such as high-frequency financial markets or complex human behavior analysis—to further validate the broader applicability of our findings. Finally, our observations regarding CI and CD model performance are conditioned on the current state-of-the-art model families; the emergence of hybrid architectures may shift these dynamics in future research.

\bibliographystyle{splncs04}
\bibliography{literature}
\clearpage

\appendix

\section{Description of All Datasets}
\label{sec:dataset-description}

In this section, we provide a detailed description of the various datasets used in our experiments. 

\paragraph{Standard Datasets}
We first utilize standard benchmarks widely recognized in the time series forecasting literature. The \textbf{ETT} (Electricity Transformer Training) datasets represent long-term power deployment indicators, recorded at two granularities: 15-minute intervals (denoted by 'm') and 1-hour intervals (denoted by 'h') across two different counties (1 and 2). The \textbf{Electricity} dataset tracks hourly consumption for 321 households over three years. The \textbf{Traffic} dataset represents traffic flow recorded by 862 sensors at an hourly frequency over two years. Finally, the \textbf{Weather} dataset includes 21 meteorological features recorded every 10 minutes. Statistics for these datasets are summarized in Table~\ref{tab:standard-datasets}.

\begin{table}[ht]
    \centering
    \caption{Basic statistics about the standard datasets used in the time series forecasting literature.}
    \label{tab:standard-datasets}
    \small
    \begin{tabular}{lcccc}
        \toprule
        \textbf{Dataset} & \textbf{Channels} & \textbf{Granularity} & \textbf{Split (T/V/T)} & \textbf{Total Steps} \\
        \midrule
        ETTh (1/2)  & 7   & 1 hour  & 12/4/4 (months) & 17,420 \\
        ETTm (1/2)  & 7   & 15 mins & 12/4/4 (months) & 69,680 \\
        Electricity & 321 & 1 hour  & 7:1:2           & 26,304 \\
        Traffic     & 862 & 1 hour  & 7:1:2           & 17,544 \\
        Weather     & 21  & 10 mins & 7:1:2           & 52,696 \\
        \bottomrule
    \end{tabular}
\end{table}

\paragraph{Chaotic ODE Benchmark}
The second category of datasets is based on the Chaotic ODE Benchmark. We follow a common protocol for the six chosen attractors, generating 20 points per time unit and utilizing a 7:1:2 train:validation:test split. We generate a long time series of 60,000 timesteps, divided into forecasting samples via a rolling window. For reproducibility, we provide the code and random seeds required to generate these datasets. For further details on the dimensionality and underlying processes, we refer to the Chaotic ODE Benchmark paper \cite{gilpin21}.

\section{Hyperparameter Tuning}
\label{sec:hyperparameter-tuning}

We evaluate various Transformer-based and MLP-based models under a unified hyperparameter search space (Table~\ref{tab:common-hyperparameters}).

\begin{table}[ht]
    \centering
    \caption{Common Hyperparameter Search Ranges. Values are integers except for learning rate and dropout (floats).}
    \label{tab:common-hyperparameters}
    \small
    \begin{tabular}{ll}
        \toprule
        \textbf{Hyperparameter} & \textbf{Search Range} \\ 
        \midrule
        Learning Rate           & $[10^{-7}, 10^{-2}]$ \\ 
        Hidden Dimension ($d_{model}$)      & $\{128, 256, 512, 1024\}$ \\
        Feedforward Dimension ($d_{ff}$)    & $\{128, 256, 512, 1024\}$ \\
        \# Encoder/Mixer Layers & $[1, 10]$ \\
        Dropout                 & $[0, 0.9]$ \\
        Sequence Length ($seq\_len$)       & $\{96, 192, 336, 512, 720\}$ \\
        \bottomrule
    \end{tabular}
\end{table}

Model-specific parameters were tuned within the ranges suggested by their respective original papers (Table~\ref{tab:model-specific-hyperparameters}). Best-performing hyperparameter sets for each model-dataset pair are included in the supplementary materials to ensure full reproducibility.

\begin{table}[ht]
    \centering
    \caption{Model Specific Hyperparameter Search Ranges (all values are integers).}
    \label{tab:model-specific-hyperparameters}
    \small
    \begin{tabular}{lll}
        \toprule
        \textbf{Model} & \textbf{Hyperparameter} & \textbf{Range} \\
        \midrule
        PatchTST             & Patch Size      & $\{8, 16\}$ \\
                             & Stride          & $\{4, 8\}$ \\
        TSMixer              & Hidden Size     & $\{32, 64, 256, 1024\}$ \\
        Crossformer/FaCT     & Segment Length  & $[3, 12]$ \\
                             & Baseline        & $\{0, 1\}$ \\
                             & Cross Factor    & $[3, 20]$ \\     
        \bottomrule       
    \end{tabular}
\end{table}

\section{Implementation Details}
\label{sec:implementation}

\subsection{Libraries and Hardware}
All models were implemented using PyTorch (v2.4) in Python (v3.12.1). Hyperparameter optimization was conducted via Optuna (v3.6.1). Experiments were performed on machines equipped with NVIDIA RTX 4090 and RTX 3090 GPUs.

\subsection{Training Procedure}
Models were trained using the Adam optimizer with Mean Squared Error (MSE) loss. The default mini-batch size was 32. Training lasted for a maximum of 100 epochs, with early stopping triggered if validation loss did not improve for 10 consecutive epochs.

\subsection{Granger Causality}
\label{subsec:granger-supplementary}
We utilize a Granger causality framework to analyze inter-channel dependencies. We extract the first 1,000 timesteps for this statistical test. Initially, we filter out highly redundant channels (Pearson correlation $> 0.95$). We then apply the Augmented Dickey-Fuller (ADF) test to ensure stationarity, applying differencing where necessary. The causality tests are performed using the \texttt{statsmodels} package \cite{seabold2010statsmodels}.

\section{Derivation of CI/CD Model Variants}
\label{sec:CI-CD-Derivation}

We evaluated six recent baselines, creating both Channel-Independent (CI) and Channel-Dependent (CD) versions where applicable.

\paragraph{Simpler Linear/MLP Models}
\textbf{DLinear} remains a CI baseline due to its decomposition approach. For \textbf{TSMixer}, we implemented a CI version by removing the channel-mixing component (referred to as TMix-Only in the original work). Both versions are denoted as TSMixer (CI) and TSMixer (CD).

\paragraph{PatchTST}
The original PatchTST is CI. For our CD variant, we apply the transformer backbone to a flattened dimension of both patches and channels, enabling joint attention across both.

\paragraph{Crossformer}
The standard CD version uses Two-Stage Attention (TSA) over temporal segments and channels. Our CI version removes the second stage of TSA to restrict attention to the temporal dimension only. For \textbf{iTransformer}, we utilize only the CD version, as a CI version would simplify the model into a basic linear projection already covered by other baselines.

\paragraph{TimeMixer}
Originally a CD model utilizing multi-resolution MLP-mixing. We derived the CI version by removing the channel-mixing components while maintaining the multi-resolution temporal mixing structure.

\section{Detailed Results}
\label{sec:detailed-results}

Full forecasting results across all horizons are presented in Table~\ref{tab:Full-results-ODE-datasets} (ODE datasets) and Table~\ref{tab:Full-results-standard-datasets} (Standard datasets). Results represent the mean MSE over five random seeds: $\{3001, 3002, 3003, 3004, 3005\}$.

\begin{table*}[ht]
    \centering
    \caption{Full forecasting results (MSE) over standard datasets. Best result per model variant in \textbf{bold}; best overall result highlighted in \textcolor{blue}{blue}.}
    \label{tab:Full-results-standard-datasets}
    \scriptsize
    \setlength{\tabcolsep}{3.2pt} 
    \begin{tabular}{lr cccccccc cc}
        \toprule
        & & \multicolumn{2}{c}{\textbf{PatchTST}} & \multicolumn{2}{c}{\textbf{TSMixer}} & \multicolumn{2}{c}{\textbf{CrossFormer}} & \textbf{DLin.} & \textbf{iTrans.} & \multicolumn{2}{c}{\textbf{TimeMixer}} \\
        \cmidrule(r){3-4} \cmidrule(r){5-6} \cmidrule(r){7-8} \cmidrule(r){9-9} \cmidrule(r){10-10} \cmidrule(r){11-12}
        \textbf{Dataset} & \textbf{H} & \textbf{CI} & \textbf{CD} & \textbf{CI} & \textbf{CD} & \textbf{CI} & \textbf{CD} & \textbf{CI} & \textbf{CD} & \textbf{CI} & \textbf{CD} \\
        \midrule
        ETTh1 & 96  & \textbf{0.375} & 0.381 & \textbf{0.374} & 0.404 & 0.396 & \textbf{0.400} & \textcolor{blue}{\textbf{0.372}} & 0.413 & \textbf{0.391} & 0.426 \\
              & 192 & 0.419 & \textbf{0.407} & \textbf{0.432} & 0.438 & \textbf{0.452} & 0.446 & \textcolor{blue}{\textbf{0.406}} & 0.457 & \textbf{0.434} & 0.580 \\
              & 336 & \textbf{0.459} & 0.460 & \textbf{0.447} & 0.471 & \textbf{0.459} & 0.480 & \textcolor{blue}{\textbf{0.435}} & 0.500 & \textbf{0.444} & 0.500 \\
              & 720 & \textcolor{blue}{\textbf{0.437}} & 0.499 & 0.500 & \textbf{0.499} & \textbf{0.601} & 0.497 & 0.481 & 0.541 & \textbf{0.498} & 0.473 \\
        \midrule
        Weather & 96  & 0.151 & \textcolor{blue}{\textbf{0.150}} & \textbf{0.152} & 0.155 & \textcolor{blue}{\textbf{0.150}} & 0.153 & 0.374 & 0.156 & \textbf{0.158} & 0.199 \\
                & 192 & \textcolor{blue}{\textbf{0.193}} & 0.197 & \textbf{0.196} & 0.202 & 0.199 & \textcolor{blue}{\textbf{0.193}} & 0.210 & 0.201 & \textbf{0.198} & 0.289 \\
                & 336 & 0.278 & \textcolor{blue}{\textbf{0.247}} & \textcolor{blue}{\textbf{0.247}} & 0.255 & 0.261 & 0.495* & 0.255 & 0.263 & \textbf{0.266} & 0.343 \\
                & 720 & 0.359 & \textbf{0.336} & \textbf{0.330} & 0.354 & 0.335 & 0.342 & \textcolor{blue}{\textbf{0.316}} & 0.319 & \textbf{0.342} & 0.442 \\
        \midrule
        Elect. & 96  & \textcolor{blue}{\textbf{0.129*}} & 0.141 & \textbf{0.131} & 0.137 & 0.139 & 0.150 & 0.135 & 0.135 & \textbf{0.130} & 0.143 \\
                    & 192 & \textcolor{blue}{\textbf{0.147*}} & 0.156 & \textbf{0.149} & 0.156 & 0.177 & 0.168 & \textbf{0.149} & 0.151 & \textbf{0.149} & 0.191 \\
                    & 336 & \textcolor{blue}{\textbf{0.163*}} & 0.173 & \textbf{0.165} & 0.176 & 0.194 & 0.182* & \textbf{0.164} & 0.175 & \textbf{0.171} & 0.174 \\
                    & 720 & \textbf{0.197*} & 0.210 & \textbf{0.203} & 0.210 & 0.261 & 0.251* & 0.199 & \textcolor{blue}{\textbf{0.196}} & 0.212 & \textbf{0.197} \\
        \midrule
        ETTh2 & 96  & 0.286 & \textcolor{blue}{\textbf{0.285}} & \textbf{0.291} & 0.304 & 0.397 & 0.537 & 0.303 & 0.324 & \textcolor{blue}{\textbf{0.285}} & 0.377 \\
              & 192 & \textcolor{blue}{\textbf{0.352}} & 0.381 & \textbf{0.376} & 0.392 & \textbf{0.713} & 0.794 & 0.397 & 0.396 & \textbf{0.360} & 0.459 \\
              & 336 & \textcolor{blue}{\textbf{0.392}} & 0.422 & \textbf{0.402} & 0.421 & 0.727 & 0.553 & 0.518 & 0.447 & \textbf{0.410} & 0.476 \\
              & 720 & \textcolor{blue}{\textbf{0.431}} & 0.438 & 0.444 & \textbf{0.441} & 1.126 & 0.880 & 0.811 & 0.441 & \textcolor{blue}{\textbf{0.431}} & 0.603 \\
        \midrule
        ETTm1 & 96  & \textbf{0.293} & 0.306 & \textcolor{blue}{\textbf{0.292}} & 0.294 & 0.309 & 0.364 & 0.299 & 0.314 & \textbf{0.299} & 0.366 \\
              & 192 & \textbf{0.334} & 0.339 & \textbf{0.333} & 0.344 & \textbf{0.398} & 0.411 & 0.334 & 0.359 & \textcolor{blue}{\textbf{0.330}} & 0.340 \\
              & 336 & \textbf{0.373} & 0.400 & \textbf{0.374} & 0.389 & 0.399 & 0.463 & 0.368 & 0.405 & \textcolor{blue}{\textbf{0.363}} & 0.468 \\
              & 720 & \textbf{0.425} & 0.441 & \textbf{0.426} & 0.452 & 0.466 & 0.465 & 0.434 & 0.437 & \textcolor{blue}{\textbf{0.418}} & 0.607 \\
        \midrule
        ETTm2 & 96  & \textcolor{blue}{\textbf{0.161}} & 0.166 & \textbf{0.163} & 0.176 & 0.199 & 0.188 & 0.166 & 0.180 & \textbf{0.166} & 0.167 \\
              & 192 & 0.221 & \textcolor{blue}{\textbf{0.220}} & \textbf{0.234} & 0.235 & \textbf{0.270} & 0.305 & 0.236 & 0.230 & \textcolor{blue}{\textbf{0.220}} & 0.242 \\
              & 336 & 0.287 & \textbf{0.273} & \textcolor{blue}{\textbf{0.271}} & 0.275 & 0.820 & 0.660 & 0.296 & 0.287 & \textbf{0.272} & 0.292 \\
              & 720 & \textbf{0.363} & 0.375 & \textcolor{blue}{\textbf{0.360}} & 0.414 & 0.442 & 0.787 & 0.458 & 0.369 & \textbf{0.361} & 0.384 \\
        \midrule
        Traffic & 96  & 0.360* & OOM & 0.386 & \textcolor{blue}{\textbf{0.384}} & OOM & 0.514* & 0.395 & OOM & OOM & 0.360* \\
                & 192 & 0.375* & OOM & \textcolor{blue}{\textbf{0.392}} & 0.405 & OOM & 0.549* & 0.406 & OOM & OOM & 0.375* \\
                & 336 & 0.385* & OOM & \textcolor{blue}{\textbf{0.407}} & 0.423 & OOM & 0.530* & 0.436 & OOM & OOM & 0.385* \\
                & 720 & 0.430* & OOM & \textcolor{blue}{\textbf{0.443}} & 0.457 & OOM & 0.573* & 0.466 & OOM & OOM & 0.430* \\
        \bottomrule
    \end{tabular}
\end{table*}

\begin{table*}[ht]
    \centering
    \caption{Full forecasting results (MSE) for Chaotic ODE datasets.}
    \label{tab:Full-results-ODE-datasets}
    \scriptsize
    \setlength{\tabcolsep}{3.2pt}
    \begin{tabular}{lr cccccccc cc}
        \toprule
        & & \multicolumn{2}{c}{\textbf{PatchTST}} & \multicolumn{2}{c}{\textbf{TSMixer}} & \multicolumn{2}{c}{\textbf{CrossFormer}} & \textbf{DLin.} & \textbf{iTrans.} & \multicolumn{2}{c}{\textbf{TimeMixer}} \\
        \cmidrule(r){3-4} \cmidrule(r){5-6} \cmidrule(r){7-8} \cmidrule(r){9-9} \cmidrule(r){10-10} \cmidrule(r){11-12}
        \textbf{Dataset} & \textbf{H} & \textbf{CI} & \textbf{CD} & \textbf{CI} & \textbf{CD} & \textbf{CI} & \textbf{CD} & \textbf{CI} & \textbf{CD} & \textbf{CI} & \textbf{CD} \\
        \midrule
        Lorenz & 96  & 0.658 & \textbf{0.658} & 0.747 & \textbf{0.729} & 0.331 & \textcolor{blue}{\textbf{0.266}} & 0.884 & 0.433 & 0.398 & \textbf{0.338} \\
               & 192 & \textbf{0.825} & 0.832 & 0.862 & \textbf{0.847} & \textcolor{blue}{\textbf{0.623}} & 0.631 & 0.922 & 0.712 & \textbf{0.662} & 0.884 \\
               & 336 & \textbf{0.906} & \textbf{0.906} & 0.927 & \textbf{0.926} & 0.792 & \textcolor{blue}{\textbf{0.762}} & 0.950 & 0.852 & \textbf{0.878} & 0.895 \\
               & 720 & \textbf{0.965} & 0.970 & 0.983 & \textbf{0.971} & 0.920 & \textcolor{blue}{\textbf{0.914}} & 0.978 & 0.954 & 0.962 & \textbf{0.929} \\
        \midrule
        Blink.Rot. & 96  & 0.162 & \textbf{0.156} & 0.322 & \textbf{0.228} & 0.064 & \textcolor{blue}{\textbf{0.045}} & 0.374 & 0.112 & 0.116 & \textbf{0.076} \\
                       & 192 & 0.322 & \textbf{0.321} & 0.510 & \textbf{0.429} & 0.230 & \textcolor{blue}{\textbf{0.210}} & 0.504 & 0.299 & \textbf{0.272} & 0.315 \\
                       & 336 & \textbf{0.511} & 0.540 & 0.727 & \textbf{0.587} & 0.491 & \textcolor{blue}{\textbf{0.466}} & 0.576 & 0.481 & 0.599 & \textbf{0.500} \\
                       & 720 & 0.703 & \textbf{0.688} & 0.759 & \textbf{0.703} & 0.576 & \textcolor{blue}{\textbf{0.524}} & 0.635 & 0.623 & 0.710 & \textbf{0.622} \\
        \midrule
        CellCycle & 96  & \textbf{0.263} & 0.311 & 0.513 & \textbf{0.505} & 0.036 & \textcolor{blue}{\textbf{0.035}} & 0.866 & 0.227 & 0.110 & \textbf{0.071} \\
                  & 192 & \textbf{0.580} & 0.624 & 0.791 & \textbf{0.740} & \textcolor{blue}{\textbf{0.244}} & 0.275 & 0.932 & 0.515 & 0.475 & \textbf{0.370} \\
                  & 336 & \textbf{0.768} & 0.795 & 0.894 & \textbf{0.876} & 0.612 & \textcolor{blue}{\textbf{0.556}} & 0.957 & 0.688 & 0.742 & \textbf{0.719} \\
                  & 720 & \textbf{0.931} & 0.939 & 0.970 & \textbf{0.963} & \textcolor{blue}{\textbf{0.825}} & 0.844 & 0.984 & 0.894 & \textbf{0.867} & 0.892 \\
        \midrule
        Doub.Pend. & 96  & 0.322 & \textbf{0.278} & 0.541 & \textbf{0.461} & 0.090 & \textcolor{blue}{\textbf{0.083}} & 0.667 & 0.314 & 0.250 & \textbf{0.110} \\
                        & 192 & \textbf{0.551} & 0.594 & \textbf{0.713} & 0.753 & \textbf{0.418} & 0.466 & 0.762 & 0.534 & 0.440 & \textcolor{blue}{\textbf{0.380}} \\
                        & 336 & \textbf{0.806} & 0.825 & 0.847 & \textbf{0.808} & 0.802 & \textbf{0.712} & 0.856 & 0.807 & 0.702 & \textcolor{blue}{\textbf{0.650}} \\
                        & 720 & \textbf{0.933} & 0.975 & 0.973 & \textbf{0.926} & 0.902 & \textbf{0.904} & 0.933 & 0.952 & \textcolor{blue}{\textbf{0.901}} & 1.018 \\
        \midrule
        Hopfield & 96  & 0.156 & \textbf{0.073} & 0.268 & \textbf{0.185} & 0.059 & \textbf{0.049} & 0.472 & 0.046 & 0.054 & \textcolor{blue}{\textbf{0.036}} \\
                 & 192 & 0.311 & \textbf{0.216} & 0.410 & \textbf{0.321} & 0.162 & \textbf{0.155} & 0.641 & 0.128 & 0.118 & \textcolor{blue}{\textbf{0.102}} \\
                 & 336 & 0.486 & \textbf{0.439} & 0.571 & \textbf{0.494} & \textbf{0.399} & 0.404 & 0.759 & 0.280 & 0.567 & \textcolor{blue}{\textbf{0.199}} \\
                 & 720 & 0.727 & \textcolor{blue}{\textbf{0.654}} & 0.778 & \textbf{0.742} & 0.721 & \textbf{0.655} & 0.886 & 0.663 & \textbf{0.699} & 0.717 \\
        \midrule
        Lor.Coupl. & 96  & 0.696 & \textbf{0.587} & 0.840 & \textbf{0.745} & 0.372 & \textcolor{blue}{\textbf{0.276}} & 0.919 & 0.610 & \textbf{0.458} & 0.818 \\
                        & 192 & \textbf{0.883} & 0.901 & 0.949 & \textbf{0.895} & 0.687 & \textcolor{blue}{\textbf{0.635}} & 0.959 & 0.840 & \textbf{0.795} & 0.947 \\
                        & 336 & \textbf{0.957} & 0.971 & 0.997 & \textbf{0.966} & 0.813 & \textcolor{blue}{\textbf{0.810}} & 0.978 & 0.924 & \textbf{0.893} & 0.984 \\
                        & 720 & \textbf{0.990} & 1.007 & 1.013 & \textbf{0.994} & \textcolor{blue}{\textbf{0.926}} & 0.942 & 0.996 & 0.984 & \textbf{0.969} & 1.003 \\                         
        \bottomrule                        
    \end{tabular}
\end{table*}

\begin{table*}[ht]
    \centering
    \caption{Standard deviation values averaged over the 4 used horizons. "---" denotes Out of Memory (OOM) errors.}
    \label{tab:std-values}
    \scriptsize
    \setlength{\tabcolsep}{4.5pt}
    \begin{tabular}{l cccccccccc}
        \toprule
        & \multicolumn{2}{c}{\textbf{PatchTST}} & \multicolumn{2}{c}{\textbf{TSMixer}} & \multicolumn{2}{c}{\textbf{CrossFormer}} & \textbf{DLin.} & \textbf{iTrans.} & \multicolumn{2}{c}{\textbf{TimeMixer}} \\
        \cmidrule(r){2-3} \cmidrule(r){4-5} \cmidrule(r){6-7} \cmidrule(r){8-8} \cmidrule(r){9-9} \cmidrule(r){10-11}
        \textbf{Dataset} & \textbf{CI} & \textbf{CD} & \textbf{CI} & \textbf{CD} & \textbf{CI} & \textbf{CD} & \textbf{CI} & \textbf{CD} & \textbf{CI} & \textbf{CD} \\
        \midrule
        Lorenz         & 0.0033 & 0.0025 & 0.0020 & 0.0115 & 0.0002 & 0.0095 & 0.0065 & 0.0303 & 0.0267 & 0.0066 \\
        Blink.Rotlet   & 0.0063 & 0.0072 & 0.0063 & 0.0083 & 0.0023 & 0.0171 & 0.0085 & 0.0580 & 0.0146 & 0.0294 \\
        CellCycle      & 0.0030 & 0.0040 & 0.0030 & 0.0047 & 0.0000 & 0.0093 & 0.0163 & 0.1127 & 0.0126 & 0.0818 \\
        Doub.Pend.     & 0.0070 & 0.0067 & 0.0070 & 0.0305 & 0.0005 & 0.0110 & 0.0088 & 0.0136 & 0.0026 & 0.0052 \\
        Hopfield       & 0.0057 & 0.0070 & 0.0057 & 0.0075 & 0.0007 & 0.0049 & 0.0129 & 0.0262 & 0.3732 & 0.0076 \\
        Lor.Coupled    & 0.0030 & 0.0045 & 0.0030 & 0.0075 & 0.0000 & 0.0127 & 0.0056 & 0.0565 & 0.0088 & 0.0088 \\
        ETTh1          & 0.0030 & 0.0092 & 0.0030 & 0.0015 & 0.0002 & 0.0425 & 0.0222 & 0.0156 & 0.0012 & 0.0175 \\
        Weather        & 0.0058 & 0.0010 & 0.0033 & 0.0053 & 0.0004 & 0.0061 & 0.0049 & 0.0172 & 0.0037 & 0.0039 \\
        Electricity    & ---    & ---    & 0.0003 & 0.0020 & 0.0000 & 0.0041 & 0.0124 & 0.0053 & 0.0017 & 0.0037 \\
        ETTh2          & 0.0040 & 0.0047 & 0.0040 & 0.0088 & 0.0011 & 0.1265 & 0.1307 & 0.0153 & 0.0036 & 0.0389 \\
        ETTm1          & 0.0023 & 0.0047 & 0.0023 & 0.0033 & 0.0000 & 0.0131 & 0.0367 & 0.0059 & 0.0023 & 0.0161 \\
        ETTm2          & 0.0010 & 0.0020 & 0.0010 & 0.0040 & 0.0004 & 0.0928 & 0.0364 & 0.0007 & 0.0022 & 0.1017 \\
        Traffic        & ---    & ---    & 0.0007 & 0.0025 & 0.0000 & ---    & ---    & ---    & ---    & ---    \\
        \bottomrule                           
    \end{tabular}
\end{table*}

\section{Detailed Granger Causality Analysis}
\label{sec:detailed-granger-results}

We evaluate inter-channel dependencies across three distinct dataset groups:
\begin{enumerate}[label=(\roman*)]
    \item Standard benchmark datasets (e.g., ETT, Weather);
    \item Chaotic ODE datasets (e.g., Lorenz, Double Pendulum);
    \item The large-scale Monash forecasting collection \cite{godahewa2021monash}.
\end{enumerate}

For datasets with high dimensionality, we restrict the Granger causality analysis to pairwise comparisons among 20 randomly selected channels to maintain computational efficiency.

\paragraph{Standard Benchmarks}
As shown in Table~\ref{tab:unified_fscore_results}, datasets such as Electricity and Traffic exhibit moderate inter-channel correlation. However, higher causality ($F$-scores) is predominantly concentrated at shorter lookback windows (lag=30). As the lag increases to 192, the predictive power of neighboring channels diminishes significantly. This quantitative decline supports the observation that Channel-Independent (CI) univariate models often perform optimally at longer lookback settings, as the marginal information gain from other channels becomes negligible over time.

\paragraph{Chaotic ODE Datasets}
In contrast, chaotic ODE datasets consistently display high $F$-scores across all lag settings. Even at a lag of 192, systems like the Lorenz and Double Pendulum maintain robust and persistent dependencies between channels. This is characteristic of deterministic physical systems where the state of one variable is intrinsically and permanently linked to the evolution of others. These results underscore why multivariate models (CD) are particularly well-suited for physical or simulated dynamics.

\paragraph{Monash Forecasting Collection}
We analyzed 40 datasets from the Monash repository after filtering for missing values. As detailed in Table~\ref{tab:Monash_granger}, the vast majority of these real-world datasets lack strong lateral dependencies. Only six of the forty datasets show an inter-channel $F$-score above 10, with most scores declining sharply as lag increases. Many of these datasets (e.g., Hospital or Australian Electricity) behave more like collections of independent univariate series rather than integrated multivariate systems.

\paragraph{Summary}
Overall, Granger causality serves as a powerful diagnostic tool for disentangling the nature of time series dependencies. While standard benchmarks permit effective univariate treatment at larger lags, ODE datasets demand explicit modeling of channel interactions. These findings suggest that research into multivariate dependencies should prioritize genuinely integrated datasets, such as chaotic ODEs, rather than simple aggregations.
\begin{table*}[ht]
    \centering
    \caption{Average $F$-scores and percentage of $F$-tests passed ($p=0.05$) across different lag settings. Asterisks (*) indicate analysis performed on 20 randomly selected channels.}
    \label{tab:unified_fscore_results}
    \small
    \setlength{\tabcolsep}{4pt}
    \begin{tabular}{l rr rr rr}
        \toprule
        & \multicolumn{2}{c}{\textbf{Lag = 30}} & \multicolumn{2}{c}{\textbf{Lag = 96}} & \multicolumn{2}{c}{\textbf{Lag = 192}} \\
        \cmidrule(r){2-3} \cmidrule(r){4-5} \cmidrule(r){6-7}
        \textbf{Dataset} & \textbf{F-score} & \textbf{Passed \%} & \textbf{F-score} & \textbf{Passed \%} & \textbf{F-score} & \textbf{Passed \%} \\
        \midrule
        Double Pendulum   & 39.60   & 100\% & 10.21  & 100\% & 6.35   & 100\% \\
        Lorenz            & 1378.54 & 83\%  & 353.25 & 83\%  & 126.38 & 83\%  \\
        Lorenz Coupled    & 579.23  & 97\%  & 160.95 & 66\%  & 72.09  & 57\%  \\
        Cell Cycle        & 6.78    & 50\%  & 2.61   & 53\%  & 1.58   & 53\%  \\
        Blinking Rotlet   & 3.98    & 33\%  & 2.59   & 33\%  & 3.76   & 33\%  \\
        HopField          & 2.86    & 80\%  & 1.92   & 70\%  & 1.55   & 50\%  \\
        \midrule
        ETTm1             & 1.48    & 50\%  & 1.13   & 33\%  & 1.04   & 23\%  \\
        ETTm2             & 1.31    & 30\%  & 1.11   & 15\%  & 0.99   & 0\%   \\
        ETTh1             & 1.87    & 60\%  & 1.12   & 20\%  & 1.07   & 7\%   \\
        ETTh2             & 1.94    & 70\%  & 1.19   & 36\%  & 1.06   & 17\%  \\
        Weather           & 1.55    & 33\%  & 1.12   & 30\%  & 1.82   & 32\%  \\
        Electricity* & 4.34    & 93\%  & 1.92   & 84\%  & 1.21   & 43\%  \\
        Traffic* & 4.71    & 94\%  & 2.22   & 94\%  & 1.39   & 69\%  \\
        \bottomrule
    \end{tabular}
\end{table*}

\begin{table*}[ht]
    \centering
    \caption{Average $F$-scores across 40 Monash datasets for lags 7, 30, 96, and 192. Empty cells or "---" indicate insufficient time points for statistical validity at that lag.}
    \label{tab:Monash_granger}
    \footnotesize
    \begin{tabular}{l cccc}
        \toprule
        \textbf{Monash Dataset} & \textbf{7} & \textbf{30} & \textbf{96} & \textbf{192} \\
        \midrule
        Australian electricity demand         & 29.38 & 18.06 & 2.71  & 1.62  \\
        Bitcoin (no missing)                  & 1.88  & 1.44  & 1.66  & 2.47  \\
        Car parts (no missing)                & 1.25  & ---   & ---   & ---   \\
        CIF 2016                              & 1.90  & ---   & ---   & ---   \\
        Dominick                              & 1.26  & ---   & ---   & ---   \\
        Electricity hourly                    & 32.55 & 4.66  & 1.96  & 1.21  \\
        Electricity weekly                    & 1.18  & 1.11  & ---   & ---   \\
        Fred MD                               & 1.92  & 1.49  & 1.32  & 1.14  \\
        Hospital                              & 1.38  & ---   & ---   & ---   \\
        Kaggle web traffic (no missing)       & 3.02  & 3.25  & 12.96 & 45.63 \\
        Kaggle web traffic weekly             & 1.14  & 1.56  & ---   & ---   \\
        KDD Cup 2018 (no missing)             & 2.48  & 1.53  & 1.18  & 1.08  \\
        London smart meters (no missing)      & 1.69  & 1.26  & 1.07  & 1.03  \\
        M1 Monthly / Quarterly / Yearly       & 2.02  & ---   & ---   & ---   \\
        M3 Monthly / Other / Quarterly / Yearly & 1.59  & ---   & ---   & ---   \\
        M4 Daily                              & 1.10  & 1.05  & ---   & ---   \\
        M4 Hourly                             & 18.06 & 3.48  & 1.96  & 1.20  \\
        M4 Monthly / Quarterly / Weekly       & 1.76  & ---   & ---   & ---   \\
        NN5 Daily (no missing)                & 9.45  & 2.10  & 1.17  & 1.02  \\
        NN5 Weekly                            & 1.17  & 1.03  & ---   & ---   \\
        Pedestrian counts                     & 10.85 & 3.14  & 2.65  & ---   \\
        Rideshare (no missing)                & 3.73  & 2.15  & 1.31  & ---   \\
        Solar 10 minutes                      & 8.63  & 3.87  & 1.52  & 1.56  \\
        Solar weekly                          & 0.78  & ---   & ---   & ---   \\
        Temperature rain (no missing)         & 2.85  & 1.54  & 1.20  & 1.06  \\
        Tourism Monthly / Quarterly / Yearly  & 12.12 & 1.58  & ---   & ---   \\
        Traffic hourly                        & 16.00 & 4.44  & 2.08  & 1.25  \\
        Traffic weekly                        & 1.12  & 1.21  & ---   & ---   \\
        Vehicle trips (no missing)            & 1.38  & ---   & ---   & ---   \\
        Weather                               & 1.22  & 1.17  & 1.12  & 1.03  \\
        \bottomrule
    \end{tabular}
\end{table*}

\section{Full Lookback Window Analysis}
\label{sec:full-lw-analysis}

\begin{figure*}[t]
    \centering
    \begin{subfigure}[b]{0.32\linewidth}
        \centering
        \includegraphics[width=\linewidth]{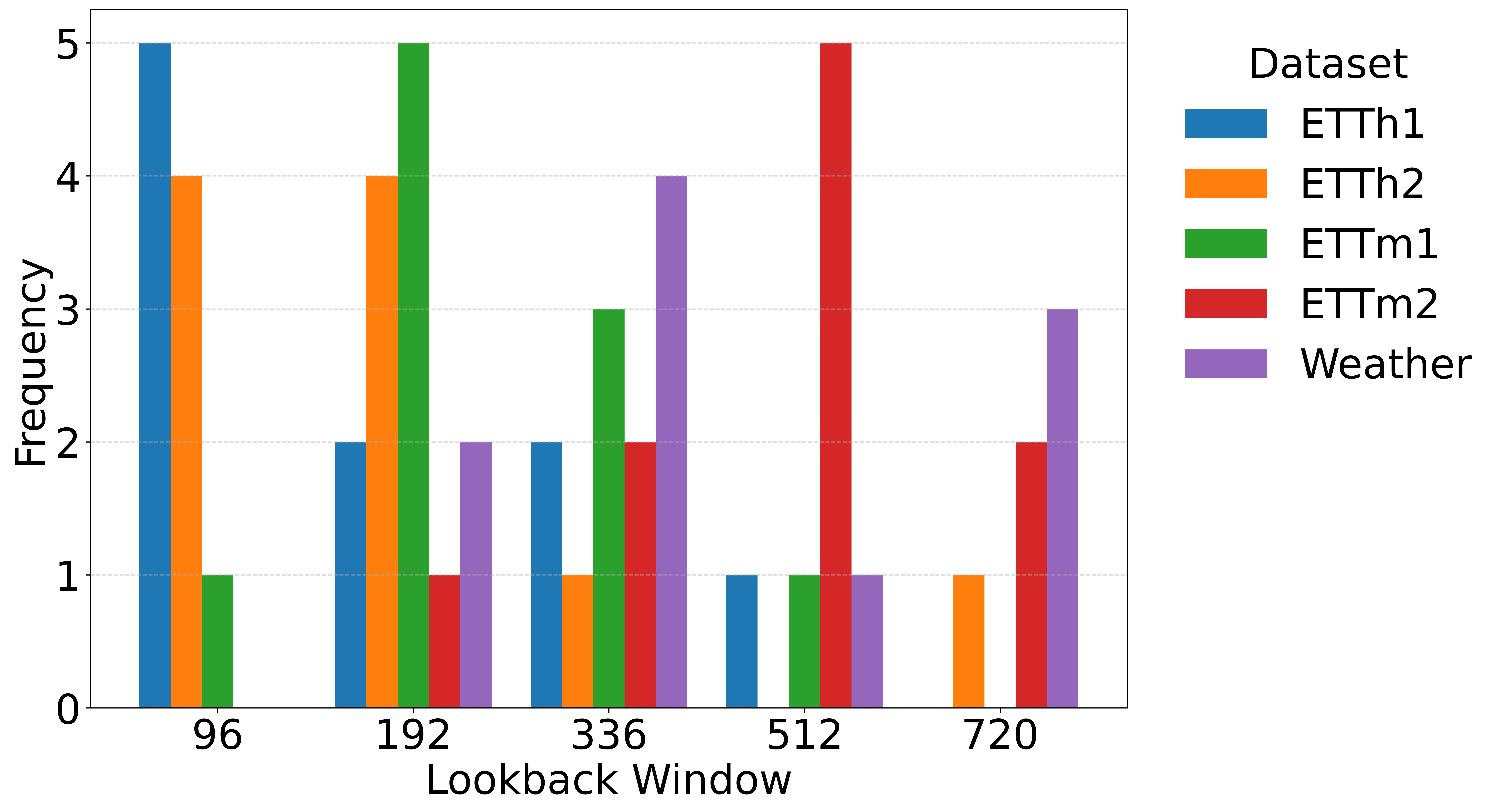}
        \caption{By Dataset}
    \end{subfigure}
    \hfill
    \begin{subfigure}[b]{0.32\linewidth}
        \centering
        \includegraphics[width=\linewidth]{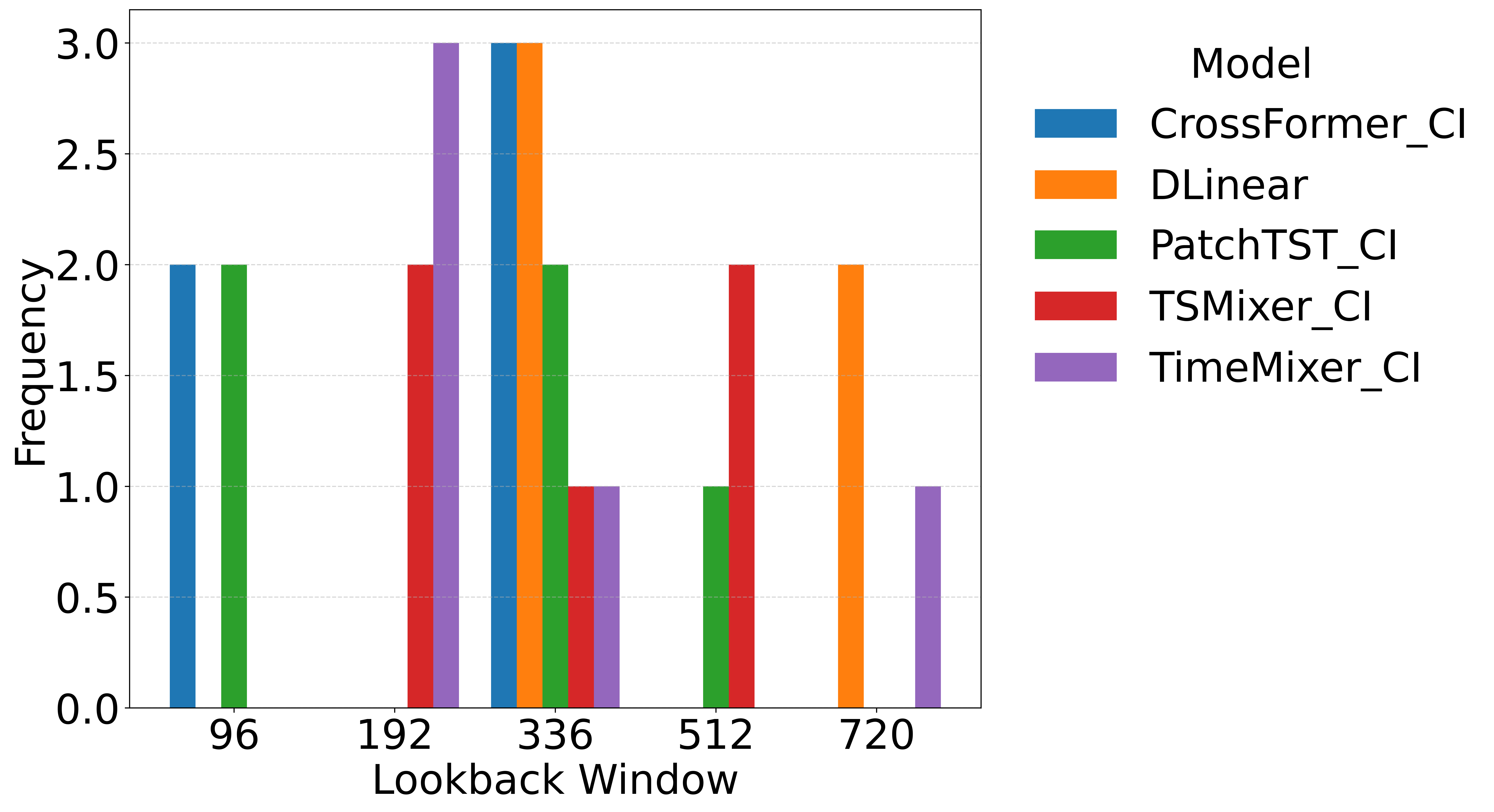}
        \caption{By CI Models}
    \end{subfigure}
    \hfill
    \begin{subfigure}[b]{0.32\linewidth}
        \centering
        \includegraphics[width=\linewidth]{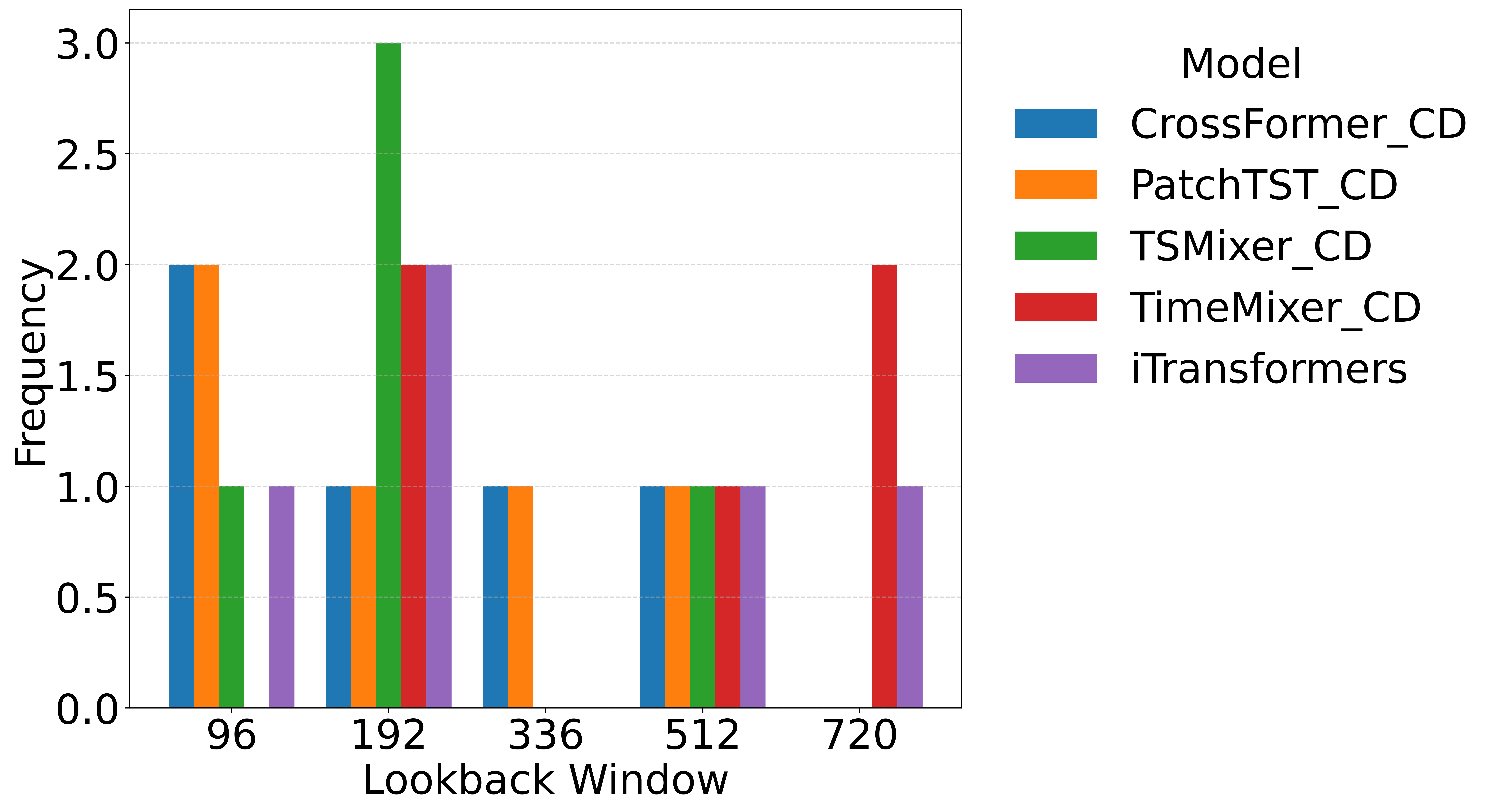}
        \caption{By CD Models}
    \end{subfigure}
    \caption{Frequency analysis of best-performing lookback windows for \textbf{forecasting horizon 96}.}
    \label{fig:lw-analysis-96}
\end{figure*}

\begin{figure*}[t]
    \centering
    \begin{subfigure}[b]{0.32\linewidth}
        \centering
        \includegraphics[width=\linewidth]{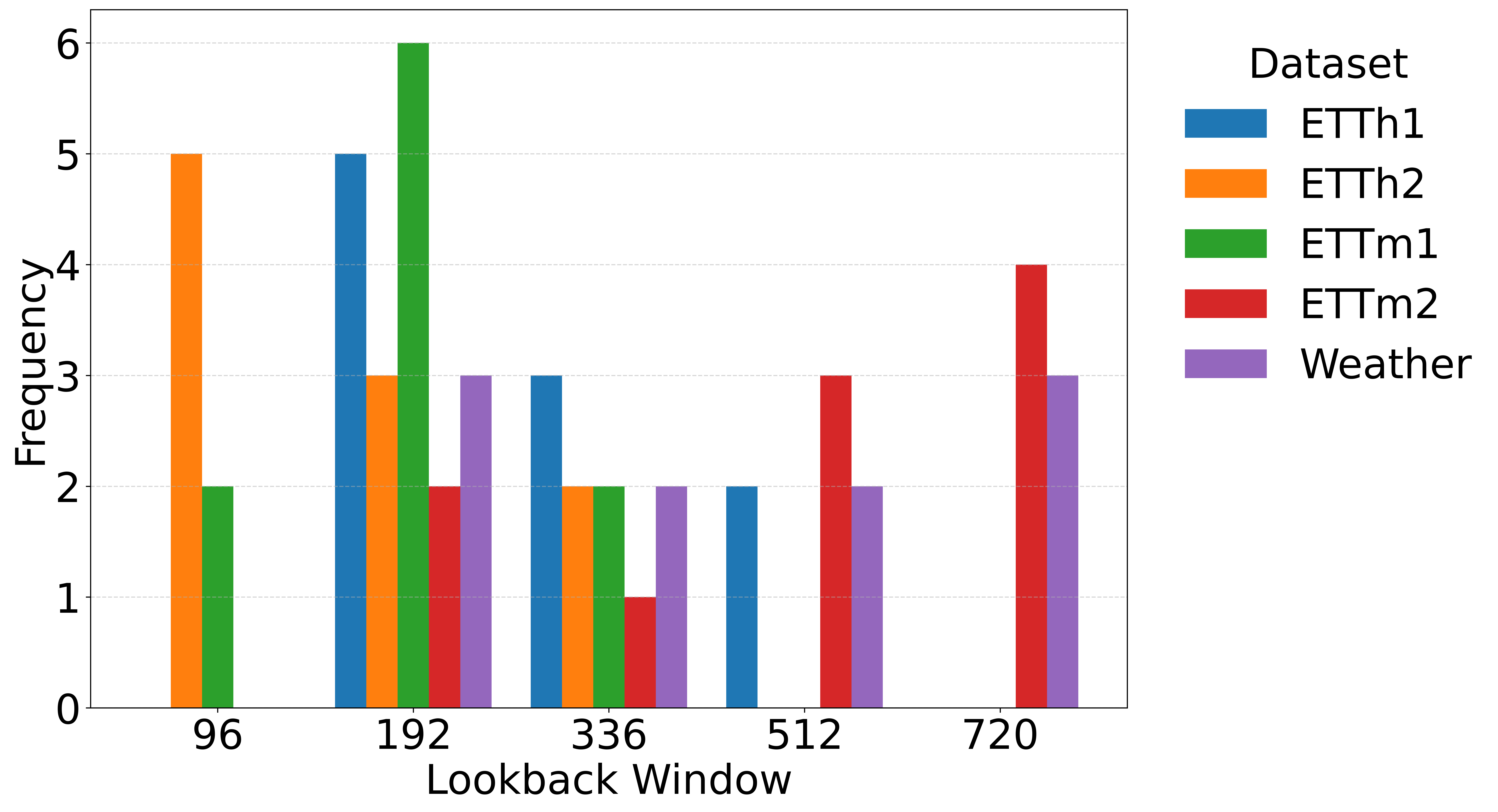}
        \caption{By Dataset}
    \end{subfigure}
    \hfill
    \begin{subfigure}[b]{0.32\linewidth}
        \centering
        \includegraphics[width=\linewidth]{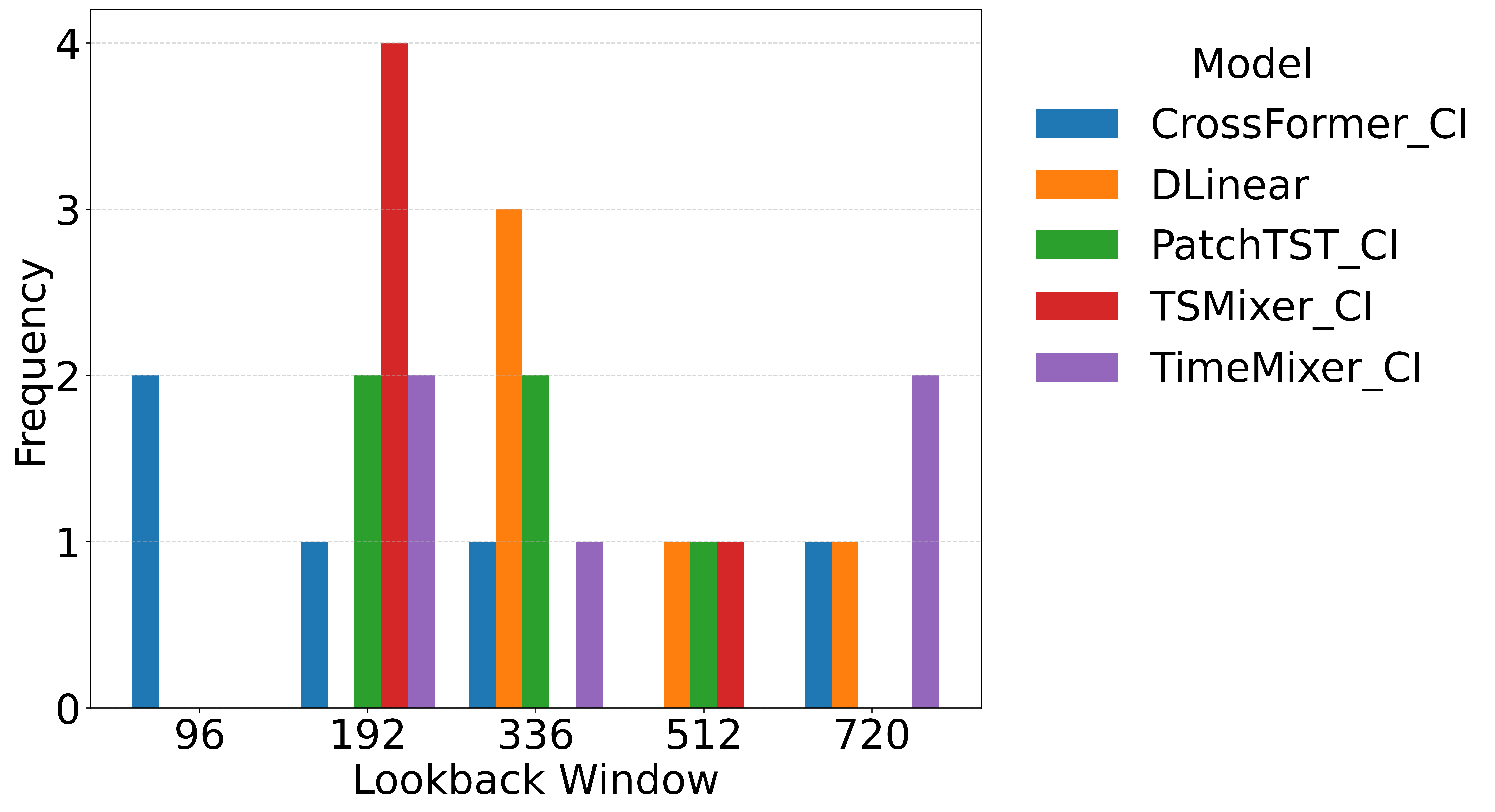}
        \caption{By CI Models}
    \end{subfigure}
    \hfill
    \begin{subfigure}[b]{0.32\linewidth}
        \centering
        \includegraphics[width=\linewidth]{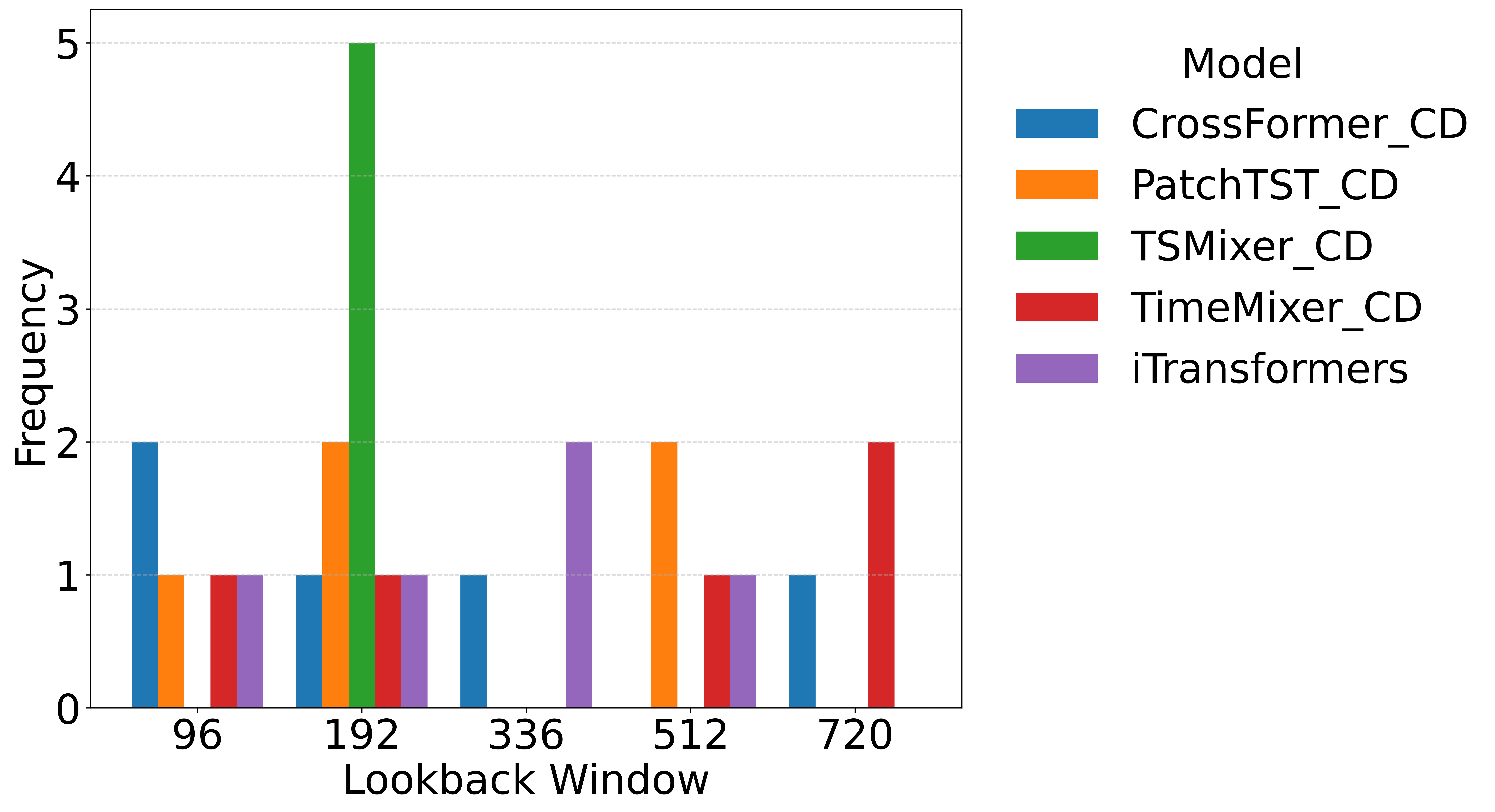}
        \caption{By CD Models}
    \end{subfigure}
    \caption{Frequency analysis of best-performing lookback windows for \textbf{forecasting horizon 192}.}
    \label{fig:lw-analysis-192}
\end{figure*}

\begin{figure*}[t]
    \centering
    \begin{subfigure}[b]{0.32\linewidth}
        \centering
        \includegraphics[width=\linewidth]{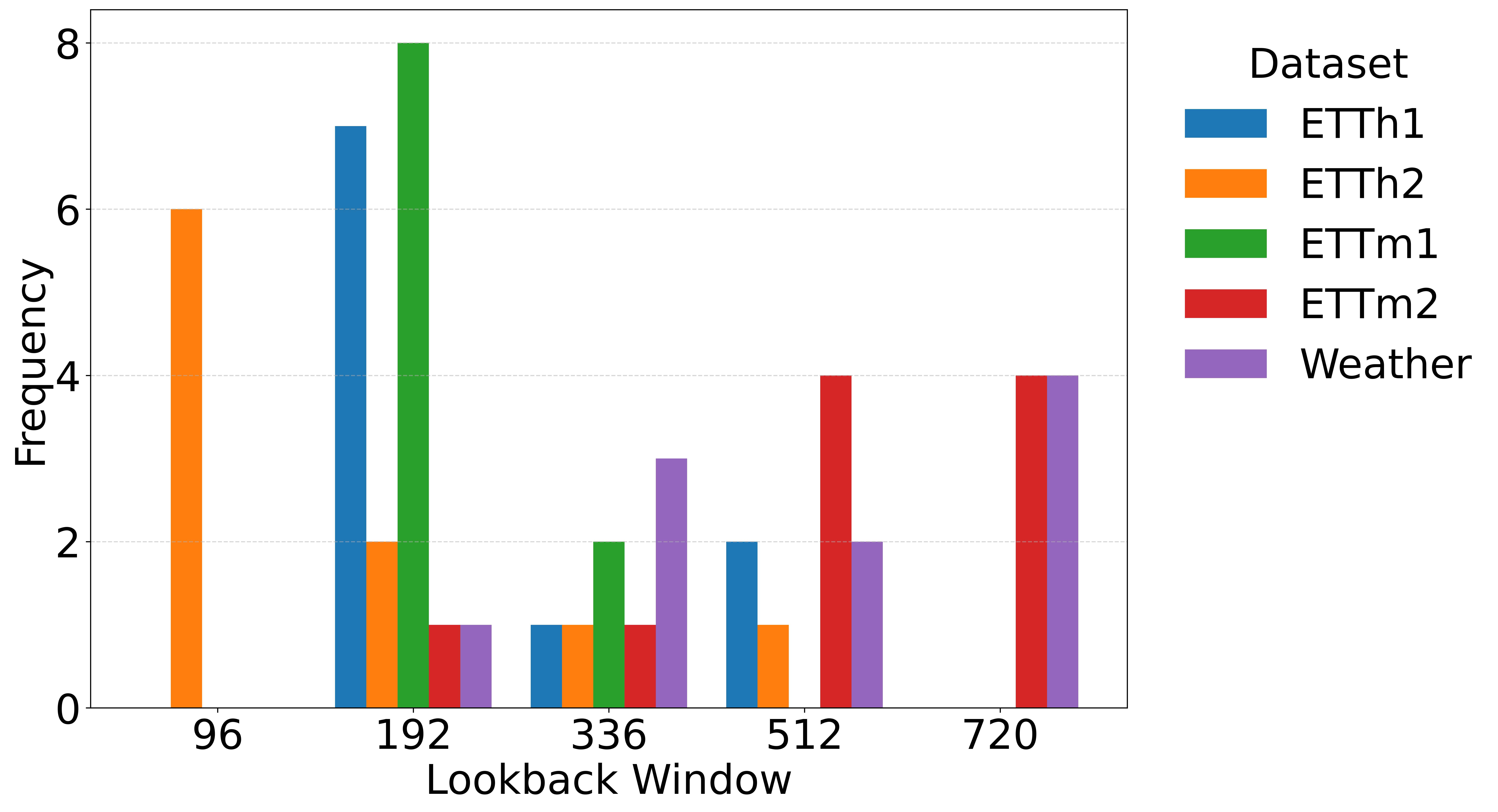}
        \caption{By Dataset}
    \end{subfigure}
    \hfill
    \begin{subfigure}[b]{0.32\linewidth}
        \centering
        \includegraphics[width=\linewidth]{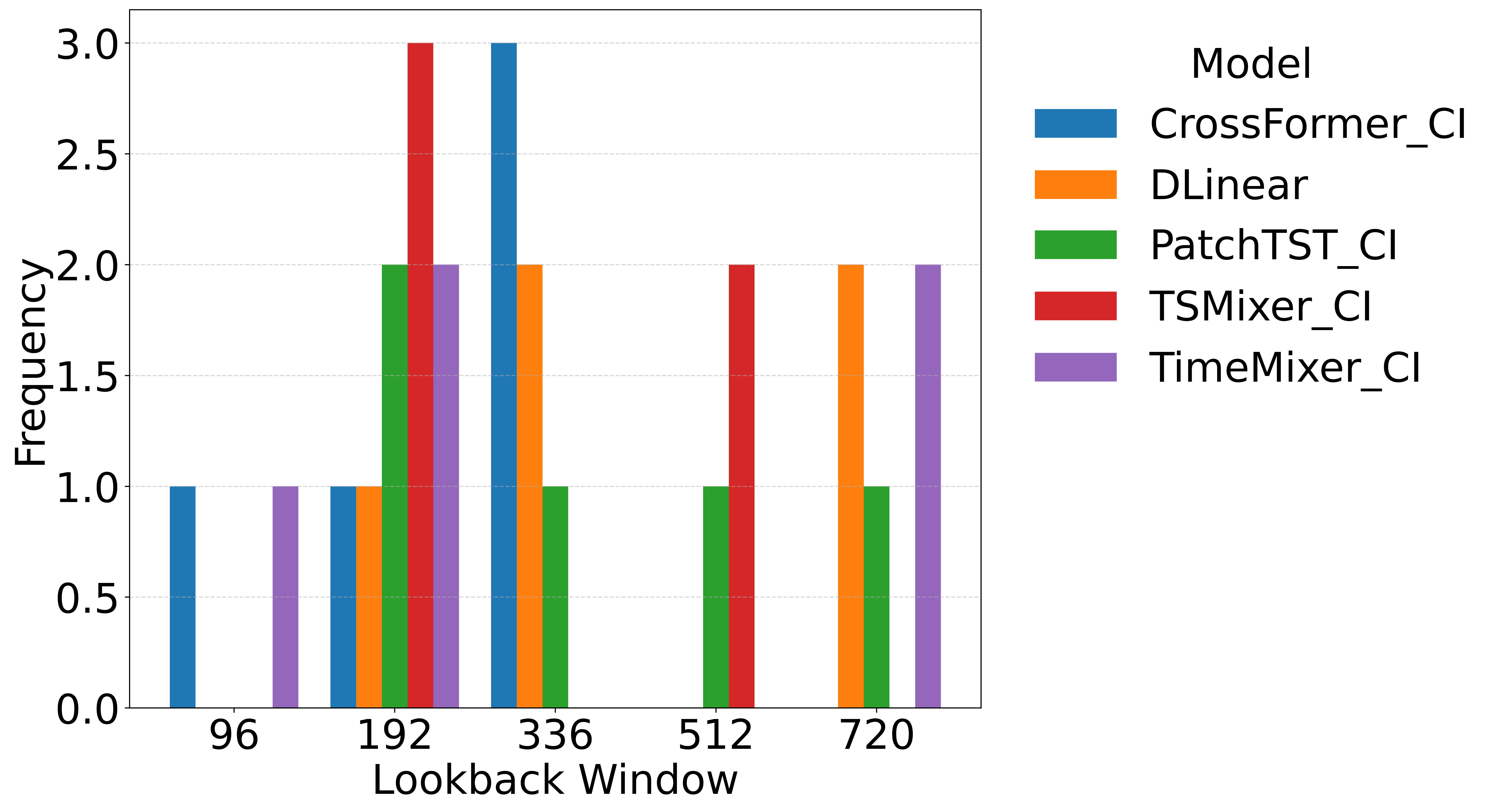}
        \caption{By CI Models}
    \end{subfigure}
    \hfill
    \begin{subfigure}[b]{0.32\linewidth}
        \centering
        \includegraphics[width=\linewidth]{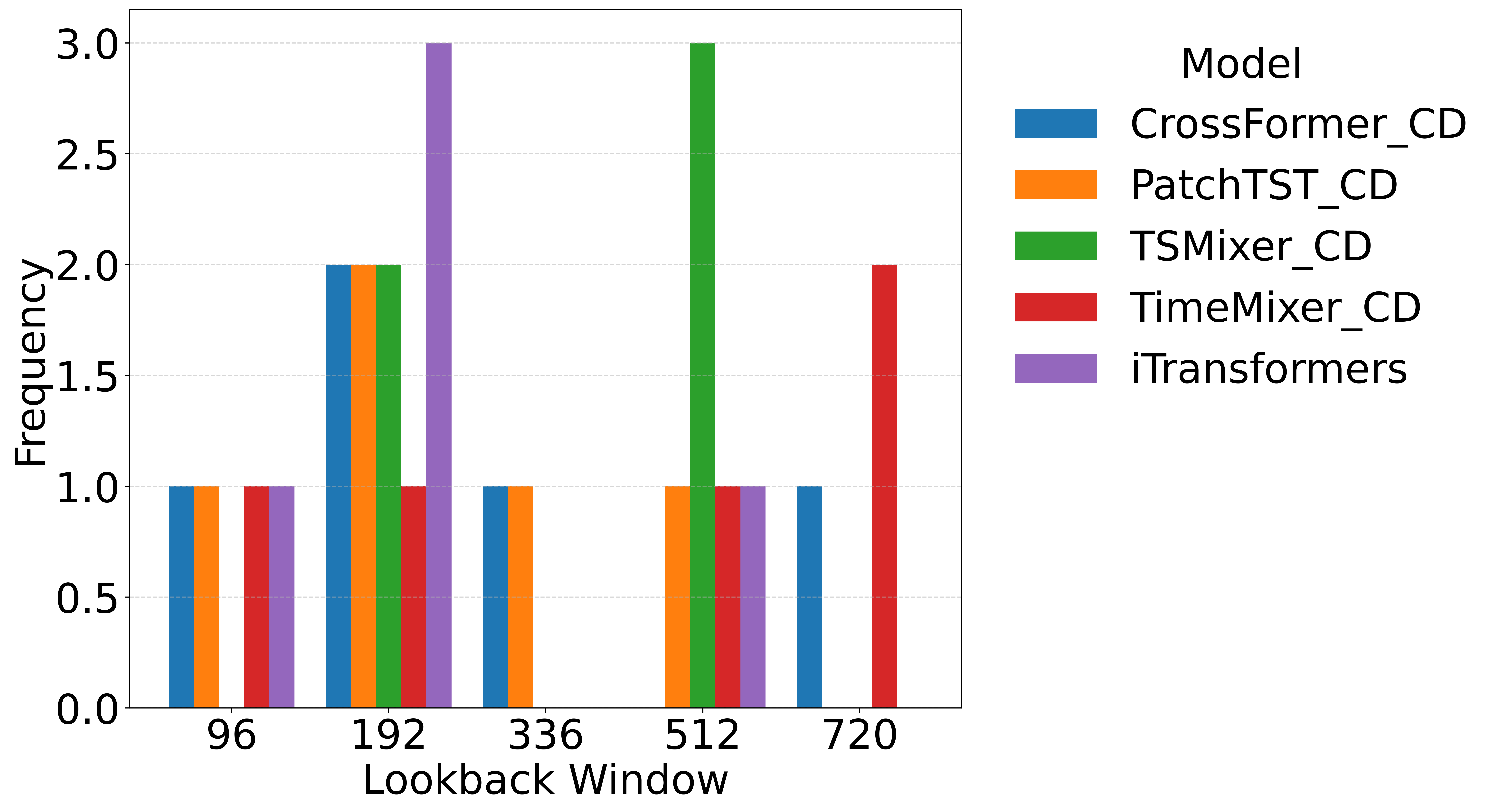}
        \caption{By CD Models}
    \end{subfigure}
    \caption{Frequency analysis of best-performing lookback windows for \textbf{forecasting horizon 336}.}
    \label{fig:lw-analysis-336}
\end{figure*}

\begin{figure*}[t]
    \centering
    \begin{subfigure}[b]{0.32\linewidth}
        \centering
        \includegraphics[width=\linewidth]{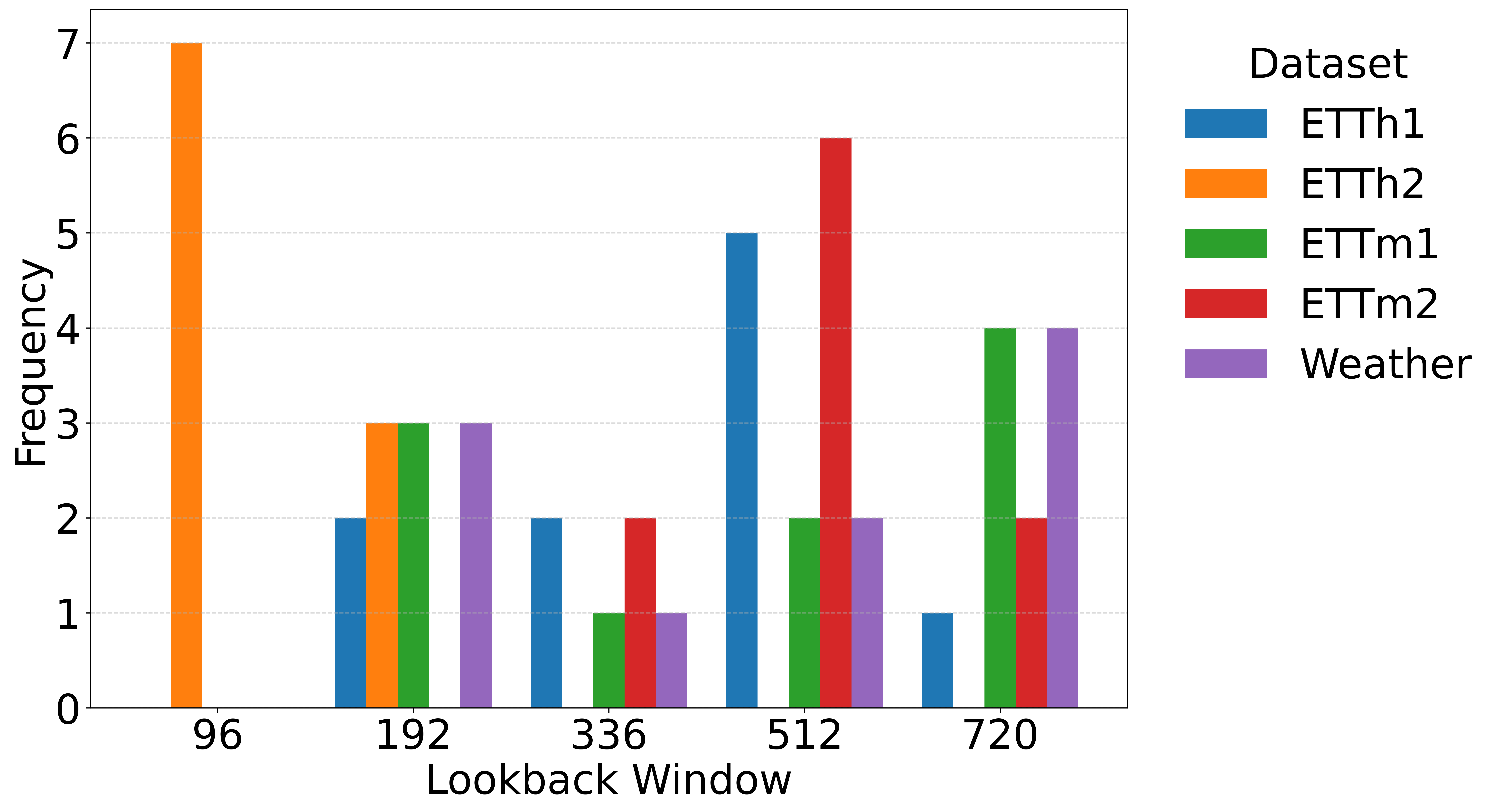}
        \caption{By Dataset}
    \end{subfigure}
    \hfill
    \begin{subfigure}[b]{0.32\linewidth}
        \centering
        \includegraphics[width=\linewidth]{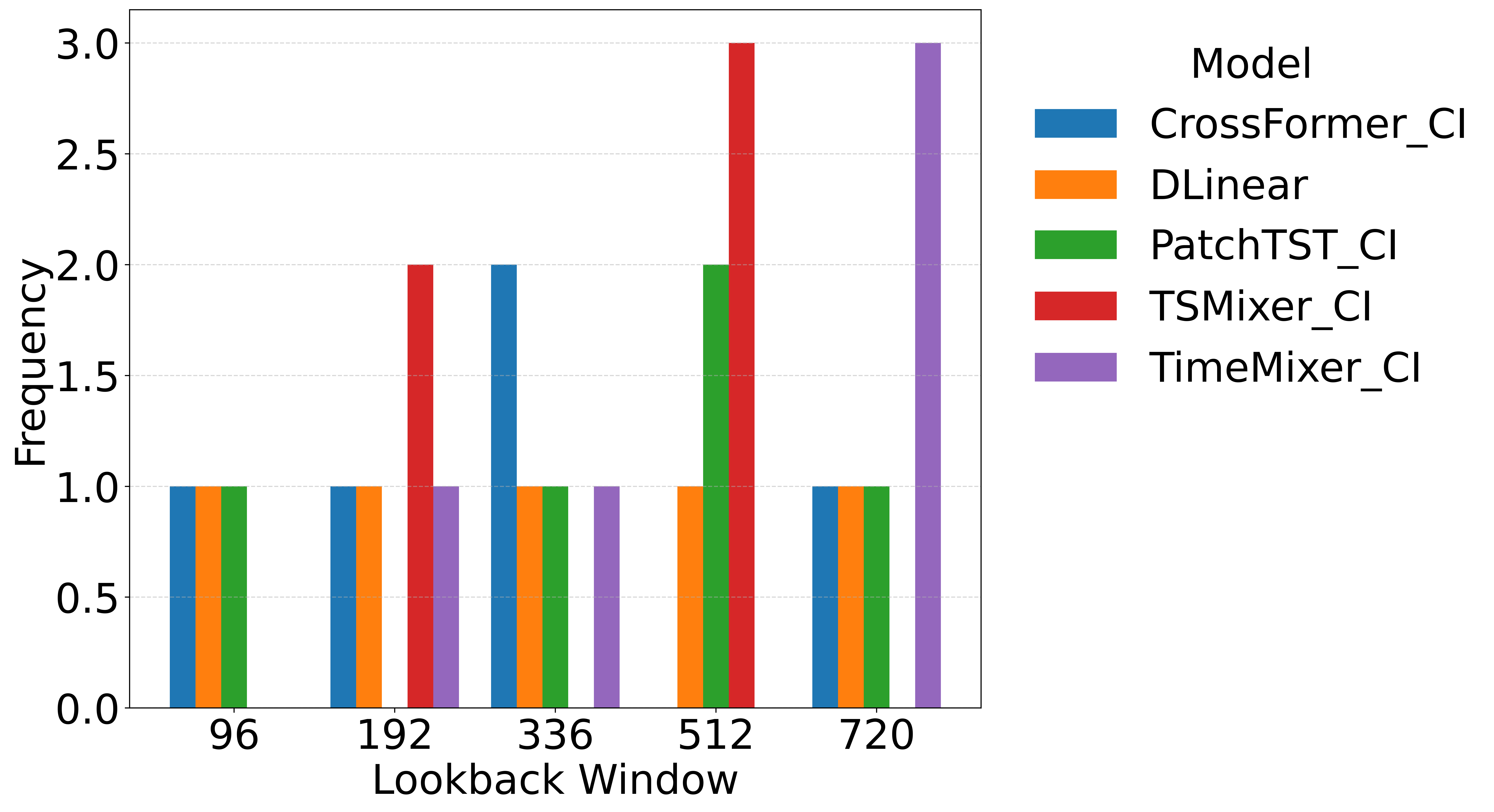}
        \caption{By CI Models}
    \end{subfigure}
    \hfill
    \begin{subfigure}[b]{0.32\linewidth}
        \centering
        \includegraphics[width=\linewidth]{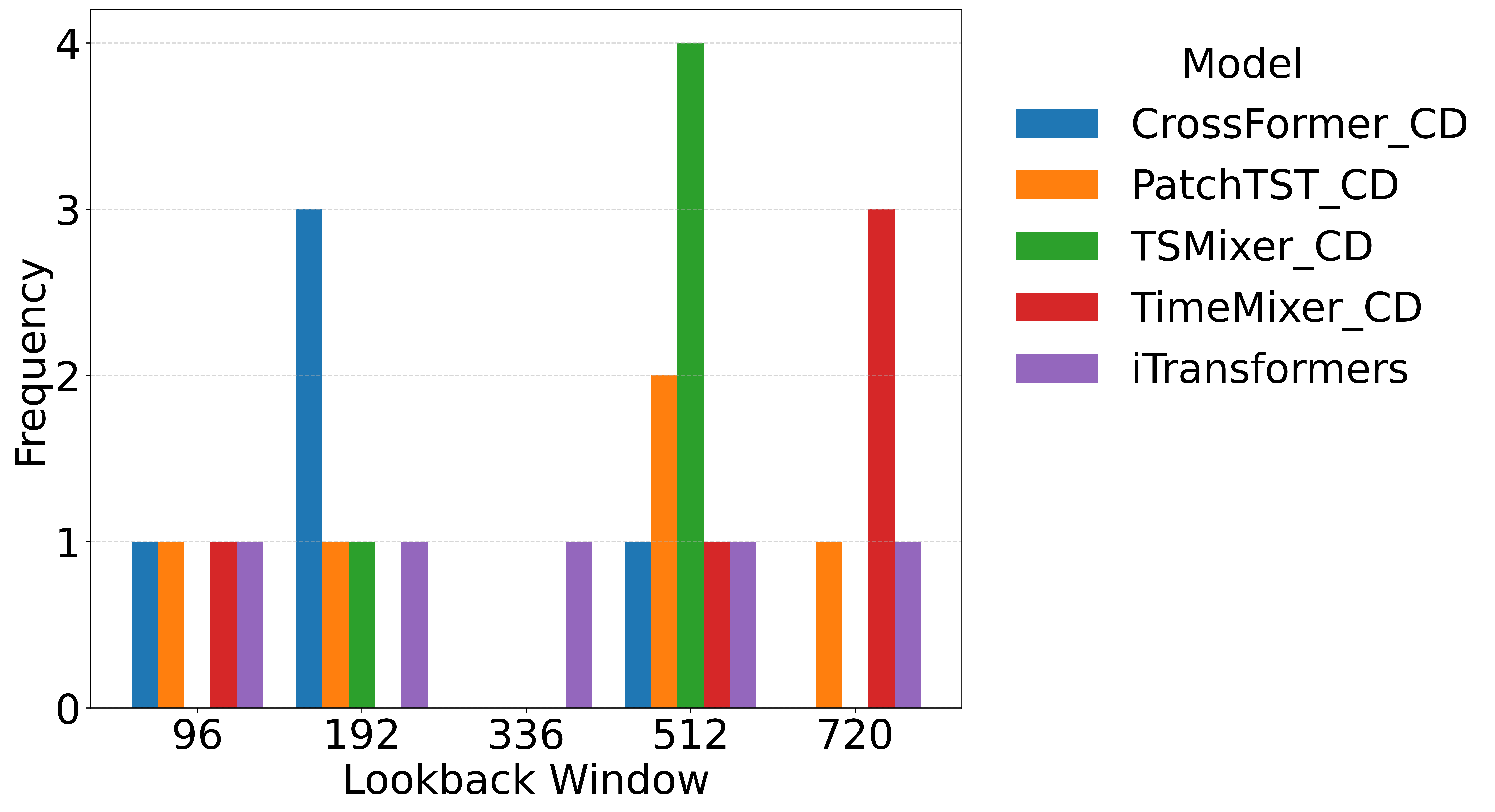}
        \caption{By CD Models}
    \end{subfigure}
    \caption{Frequency analysis of best-performing lookback windows for \textbf{forecasting horizon 720}.}
    \label{fig:lw-analysis-720}
\end{figure*}

This section provides a granular analysis of optimal lookback window frequencies across all evaluated forecasting horizons ($H \in \{96, 192, 336, 720\}$). Figures~\ref{fig:lw-analysis-96} through \ref{fig:lw-analysis-720} detail these results, categorized by dataset, Channel-Independent (CI) models, and Channel-Dependent (CD) models. 

On average, the insights derived from the $H=96$ horizon remain consistent across the extended experiments. For instance, the tendency of ETTh1 and ETTh2 to utilize smaller lookback windows persists for horizons 192 and 336. However, at $H=720$, we observe a systemic shift toward larger lookback windows. This transition is intuitive; the increased difficulty of forecasting across an extended horizon promotes the use of broader historical context for all datasets.

Furthermore, the observation that CI models typically benefit more from larger lookback windows than CD models holds true across horizons. As the forecasting horizon expands to 720, a greater number of models across all categories favor larger windows. These patterns underscore the complexity of multivariate time series forecasting and confirm that the systematic tuning of lookback windows is essential for optimal performance across diverse horizons.

\section{ODE Datasets Sample Visualization}
\label{sec:ode-viz-samples}

\begin{figure}[ht]
    \centering
    \includegraphics[width=\linewidth]{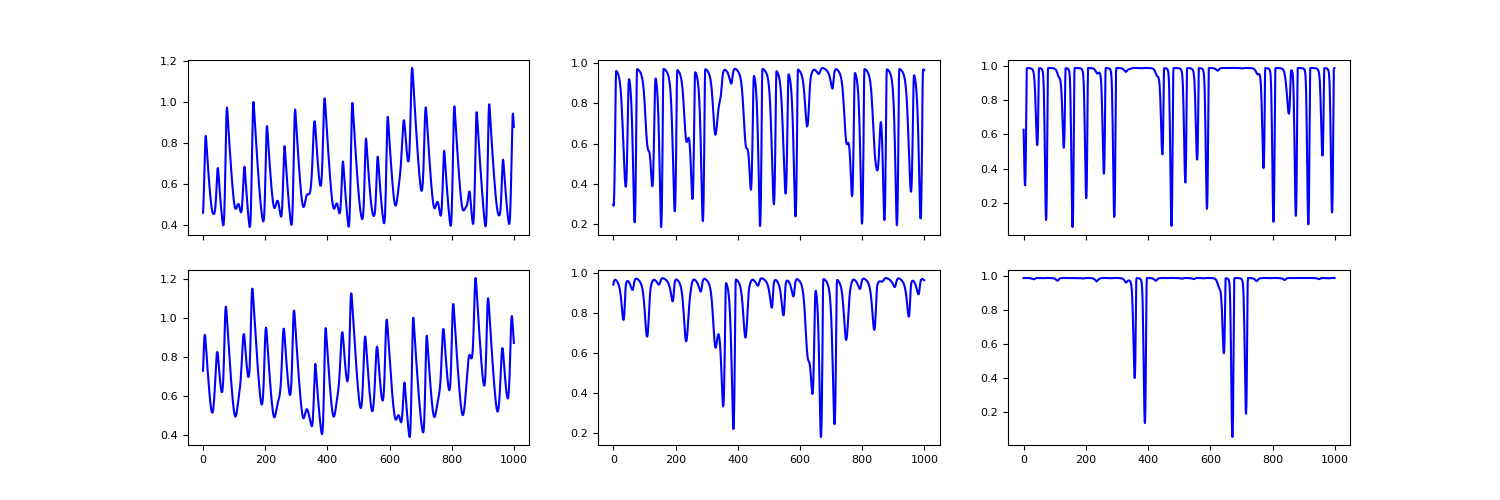}
    \caption{Time series visualization of the cell cycle ODE-based time series sample.}
    \label{fig:cell-cycle}
\end{figure}

In this section, we provide visualizations for the remaining six ODE datasets utilized in this research. Figure~\ref{fig:cell-cycle} illustrates 1,000 timesteps from the Cell Cycle-based ODE. Specifically, two distinct cyclin degradation activations are observable: channel 2 (middle-top) activates the degradation of cyclin C1 levels (middle-bottom). This phenomenon is replicated on the rightmost two channels for cyclin C2 levels.

\begin{figure}[ht]
    \centering
    \includegraphics[width=\linewidth]{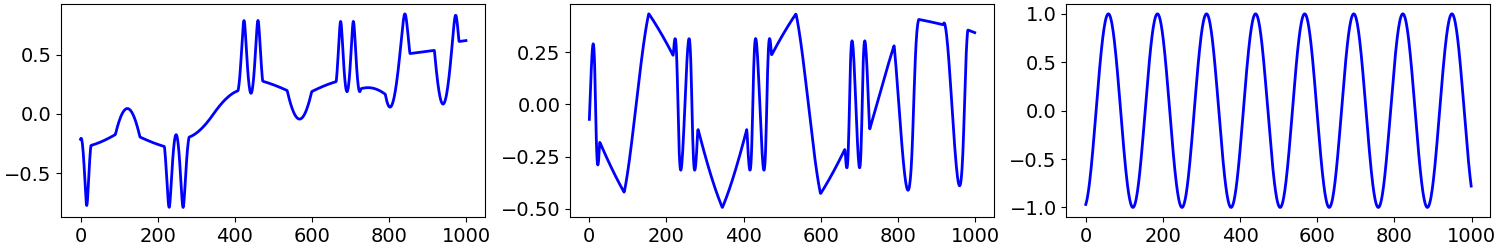}
    \caption{Time series visualization of the blinking rotlet ODE-based time series sample.}
    \label{fig:blinking-rotlet}
\end{figure}

\begin{figure}[ht]
    \centering
    \includegraphics[width=\linewidth]{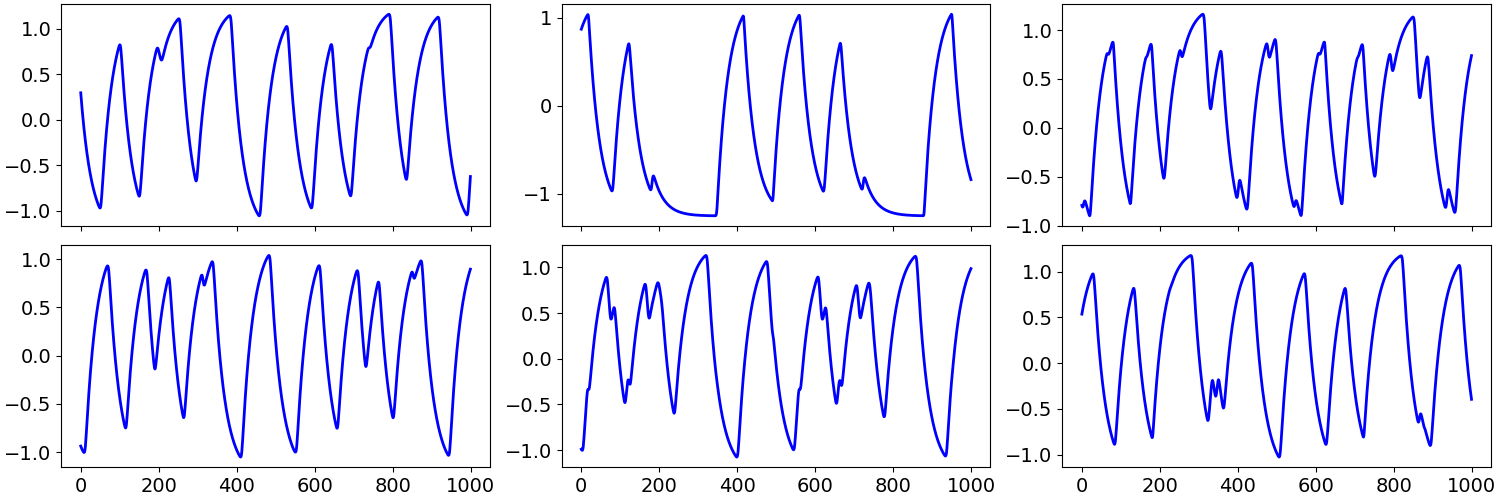}
    \caption{Time series visualization of the hopfield ODE-based time series sample.}
    \label{fig:hopfield}
\end{figure}

\begin{figure}[ht]
    \centering
    \includegraphics[width=\linewidth]{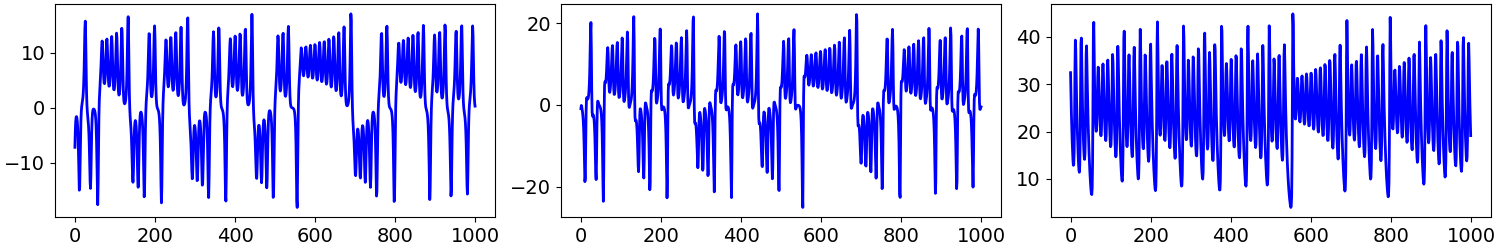}
    \caption{Time series visualization of the lorenz ODE-based time series sample.}
    \label{fig:lorenz}
\end{figure}

\begin{figure}[ht]
    \centering
    \includegraphics[width=\linewidth]{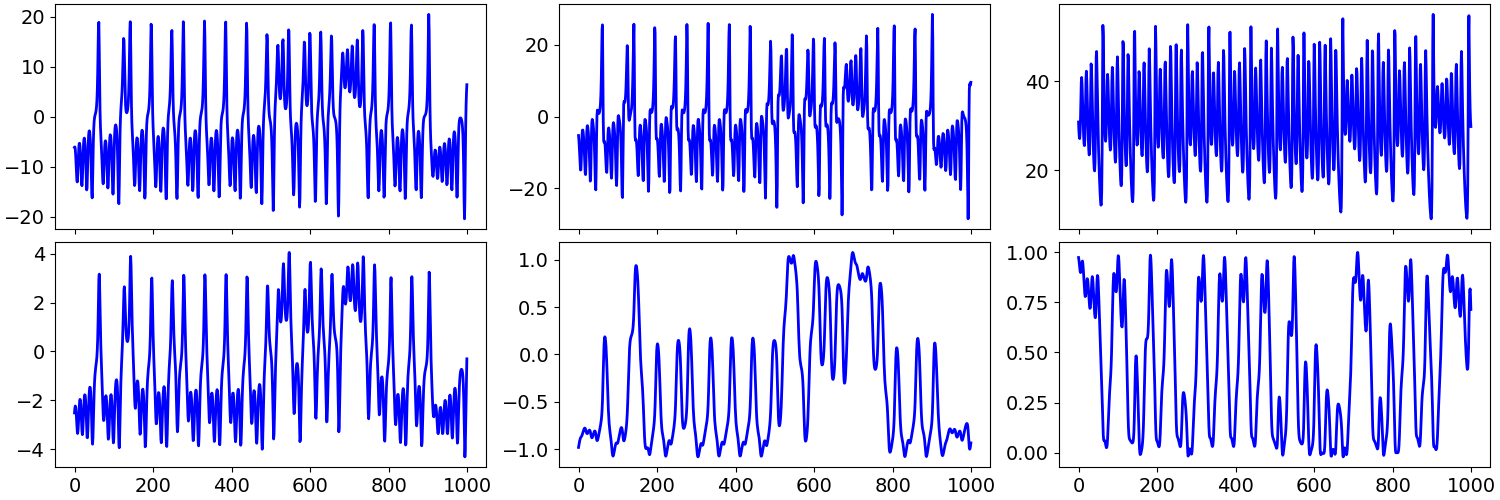}
    \caption{Time series visualization of the Lorenz Coupled ODE-based time series sample.}
    \label{fig:lorenz-coupled}
\end{figure}

Figures~\ref{fig:blinking-rotlet}, \ref{fig:hopfield}, \ref{fig:lorenz}, and \ref{fig:lorenz-coupled} display the first 1,000 timesteps for the Blinking Rotlet, Hopfield, Lorenz, and Lorenz-Coupled datasets, respectively. These visualizations demonstrate how deterministic dynamics generate complex, real-world-like multivariate patterns, where temporal fluctuations in one channel directly influence or trigger corresponding shifts in associated channels.

\clearpage

\section{Cell Cycle ODE Discussion}

We have already discussed the interesting properties of Double Pendulum ODEs. In this part of the appendix, we would like to further extend the discussion on the coupling between the ODEs and how variables interact across time. Equations 1--6 present the system of ODEs we used to generate the cell cycle dataset \cite{romond1999}. These six equations show two different oscillators based on the concentration levels of cyclin ($C_1$ and $C_2$) and how they motivate the progression of the cell cycle. 

The cyclin concentration reflects directly on the activated Cyclin-Dependent Kinases (cdk1, cdk2), which are represented in the equations by $M_1$ and $M_2$, respectively; it also directly affects enzymes $X_1$ and $X_2$. The terms $v_{ij}$ and $v_{dj}$ are constants multiplied in Equation 1 (for $j=1$) and Equation 4 (for $j=2$) as a rate of cyclin synthesis. In Equation 1, we can see clearly the coupling between the ODE variables $C_1$ and $M_2$, whereas in Equation 4, we see the coupling between $C_2$ and $M_1$. 

All of this demonstrates how information from one channel can cause spikes or changes of trend in other channels, showing clear channel dependence in the generated time-series data. Furthermore, this section is valuable for demonstrating the important application of such equations to a real-life biomedical process, which further elevates the importance of our study. Please note that these equations were taken from \cite{romond1999}; for more details, we encourage the reader to review their work.

\begin{equation}
    \frac{dC_1}{dt} = \frac{v_{i1} K_{im1}}{K_{im1} + M_2} - \frac{v_{d1} X_1 C_1}{K_{d1} + C_1} - k_{d1} C_1
\end{equation}

\begin{equation}
    \frac{dM_1}{dt} = \frac{V_1 (1 - M_1)}{K_1 + (1 - M_1)} - \frac{V_2 M_1}{K_2 + M_1}
\end{equation}

\begin{equation}
    \frac{dX_1}{dt} = \frac{V_3 (1 - X_1)}{K_3 + (1 - X_1)} - \frac{V_4 X_1}{K_4 + X_1}
\end{equation}

\begin{equation}
    \frac{dC_2}{dt} = \frac{v_{i2} K_{im2}}{K_{im2} + M_1} - \frac{v_{d2} X_2 C_2}{K_{d2} + C_2} - k_{d2} C_2
\end{equation}

\begin{equation}
    \frac{dM_2}{dt} = \frac{U_1 (1 - M_2)}{H_1 + (1 - M_2)} - \frac{U_2 M_2}{H_2 + M_2}
\end{equation}

\begin{equation}
    \frac{dX_2}{dt} = \frac{U_3 (1 - X_2)}{H_3 + (1 - X_2)} - \frac{U_4 X_2}{H_4 + X_2}
\end{equation}

\begin{equation}
    V_1 = \frac{C_1}{K_{c1} + C_1}V_{M_1}, \qquad V_3 = M_1 \cdot V_{M_3}
\end{equation}

\begin{equation}
    U_1 = \frac{C_2}{K_{c2} + C_2}U_{M_1}, \qquad U_3 = M_2 \cdot U_{M_3}
\end{equation}

\section{Ablation Study on RevIN}
\label{sec:revin-ablation}

In this section, we present an ablation study evaluating the impact of Reversible Instance Normalization (RevIN) \cite{kim21} on two standard datasets and two datasets from the chaotic benchmark. Since Crossformer was originally proposed without RevIN, we integrated the RevIN component into both the Channel-Independent (CI) and Channel-Dependent (CD) versions to measure its effect across different dataset types. Additionally, we included PatchTST as a strong baseline to measure the effect of RevIN on its primary CI version.

We observed two distinct patterns in the results, as detailed in Table \ref{tab:revin-ablation}:
\begin{itemize}
    \item The introduction of RevIN improves the performance of both Crossformer variants on the standard datasets.
    \item Conversely, on the chaotic ODE datasets, the models without RevIN outperform their counterparts that include the normalization technique.
\end{itemize}

These findings confirm that, as established in the literature \cite{chen23}, RevIN generally enhances model performance on standard datasets. While this was not universally true for PatchTST in our tests, it holds for most models in existing studies. Importantly, as emphasized throughout this paper, generalizing across different types of datasets is a non-trivial task; RevIN is no exception, as it deteriorates performance when applied to Crossformer on chaotic ODE datasets. This suggests that the complex, non-stationary patterns inherent in chaotic ODEs may not benefit from the normalization-denormalization technique introduced by RevIN.

\begin{table*}[t]
    \centering
    \caption{Ablation study on Reversible Instance Normalization (RevIN). The results demonstrate that while RevIN improves performance on standard datasets (ETTh2, ETTm1), it can be detrimental to chaotic ODE datasets (Hopfield, LorenzCoupled). The best variant for each model is in \textbf{bold}, and the best overall result for each dataset is highlighted in \textcolor{blue}{blue}. Forecasting horizon $L=96$.}
    \label{tab:revin-ablation}
    \small
    \begin{tabularx}{\textwidth}{l @{\extracolsep{\fill}} cc cc cc}
        \toprule
        \textbf{Dataset / Model} & \multicolumn{2}{c}{\textbf{PatchTST (CI)}} & \multicolumn{2}{c}{\textbf{CrossFormer (CD)}} & \multicolumn{2}{c}{\textbf{CrossFormer (CI)}} \\
        \cmidrule(lr){2-3} \cmidrule(lr){4-5} \cmidrule(lr){6-7}
        & \textbf{+RevIN} & \textbf{--RevIN} & \textbf{+RevIN} & \textbf{--RevIN} & \textbf{+RevIN} & \textbf{--RevIN} \\
        \midrule
        \textbf{Hopfield}        & \textbf{0.151} & 0.156 & 0.058 & \textcolor{blue}{\textbf{0.049}} & 0.073 & \textbf{0.059} \\
        \textbf{LorenzCoupled}   & \textbf{0.689} & 0.696 & 0.453 & \textcolor{blue}{\textbf{0.276}} & 0.515 & \textbf{0.372} \\
        \textbf{ETTh2}           & 0.298 & \textbf{0.287} & \textbf{0.322} & 0.537 & \textbf{0.318} & 0.397 \\
        \textbf{ETTm1}           & 0.314 & \textbf{0.313} & \textbf{0.339} & 0.364 & \textbf{0.297} & 0.309 \\
        \bottomrule               
    \end{tabularx}
\end{table*}


\begin{xltabular}{\textwidth}{l l X}
    \caption{Hyperparameter settings for the \textbf{BlinkingRotlet} dataset across different forecasting horizons. The table is split across pages for readability.} \label{tab:hyperparams_blinkingrotlet} \\
    \toprule
    \textbf{Model} & \textbf{Setting} & \textbf{Best Hyperparameters} \\
    \midrule
    \endfirsthead
    
    \multicolumn{3}{c}{{\bfseries \tablename\ \thetable{} -- continued from previous page}} \\
    \toprule
    \textbf{Model} & \textbf{Setting} & \textbf{Best Hyperparameters} \\
    \midrule
    \endhead
    PatchTST & CI-96 & lr: 1.44e-05, e\_layers: 10, d\_ff: 256, d\_model: 1024, dropout: 0.014, fc\_dropout: 0.539, patch\_size: 16, stride: 8, seq\_len: 192 \\
    PatchTST & CD-96 & lr: 1.44e-05, e\_layers: 10, d\_ff: 256, d\_model: 1024, dropout: 0.014, fc\_dropout: 0.539, patch\_size: 16, stride: 8, seq\_len: 192 \\
    TSMixer & CI-96 & lr: 0.0001, num\_blocks: 5, hidden\_size: 32, dropout: 0.336, activation: relu, seq\_len: 512 \\
    TSMixer & CD-96 & lr: 0.0010, num\_blocks: 1, hidden\_size: 64, dropout: 0.033, activation: relu, seq\_len: 192 \\
    CrossFormer & CI-96 & lr: 0.0002, e\_layers: 7, d\_ff: 256, d\_model: 512, dropout: 0.248, seg\_len: 7, baseline: 0, cross\_factor: 20, seq\_len: 336 \\
    CrossFormer & CD-96 & lr: 0.0003, e\_layers: 7, d\_ff: 256, d\_model: 512, dropout: 0.285, seg\_len: 7, baseline: 0, cross\_factor: 5, seq\_len: 336 \\
    DLinear & CI-96 & lr: 0.0026, seq\_len: 720 \\
    iTransformers & CD-96 & lr: 1.60e-05, e\_layers: 9, d\_ff: 2048, d\_model: 1024, dropout: 0.012, seq\_len: 192 \\
    TimeMixer & CI-96 & lr: 0.0028, d\_ff: 256, d\_model: 128, e\_layers: 4, seq\_len: 192 \\
    TimeMixer & CD-96 & lr: 0.0023, d\_ff: 256, d\_model: 128, e\_layers: 5, seq\_len: 192 \\
    \midrule
    PatchTST & CI-192 & lr: 1.44e-05, e\_layers: 10, d\_ff: 256, d\_model: 1024, dropout: 0.014, fc\_dropout: 0.539, patch\_size: 16, stride: 8, seq\_len: 192 \\
    PatchTST & CD-192 & lr: 1.44e-05, e\_layers: 10, d\_ff: 256, d\_model: 1024, dropout: 0.014, fc\_dropout: 0.539, patch\_size: 16, stride: 8, seq\_len: 192 \\
    TSMixer & CI-192 & lr: 0.0005, num\_blocks: 8, hidden\_size: 1024, dropout: 0.705, activation: relu, seq\_len: 512 \\
    TSMixer & CD-192 & lr: 0.0003, num\_blocks: 1, hidden\_size: 64, dropout: 0.205, activation: gelu, seq\_len: 512 \\
    CrossFormer & CI-192 & lr: 0.0003, e\_layers: 7, d\_ff: 256, d\_model: 512, dropout: 0.312, seg\_len: 7, baseline: 0, cross\_factor: 18, seq\_len: 336 \\
    CrossFormer & CD-192 & lr: 0.0003, e\_layers: 1, d\_ff: 128, d\_model: 128, dropout: 0.266, seg\_len: 7, baseline: 0, cross\_factor: 7, seq\_len: 192 \\
    DLinear & CI-192 & lr: 0.0055, seq\_len: 512 \\
    iTransformers & CD-192 & lr: 0.0006, e\_layers: 9, d\_ff: 2048, d\_model: 128, dropout: 0.286, seq\_len: 720 \\
    TimeMixer & CI-192 & lr: 0.0023, d\_ff: 256, d\_model: 128, e\_layers: 5, seq\_len: 192 \\
    TimeMixer & CD-192 & lr: 0.0028, d\_ff: 256, d\_model: 128, e\_layers: 4, seq\_len: 192 \\
    \midrule
    PatchTST & CI-336 & lr: 1.96e-07, e\_layers: 5, d\_ff: 256, d\_model: 1024, dropout: 0.295, fc\_dropout: 0.622, patch\_size: 8, stride: 8, seq\_len: 720 \\
    PatchTST & CD-336 & lr: 9.16e-05, e\_layers: 6, d\_ff: 1024, d\_model: 1024, dropout: 0.086, fc\_dropout: 0.509, patch\_size: 16, stride: 8, seq\_len: 192 \\
    TSMixer & CI-336 & lr: 0.0012, num\_blocks: 6, hidden\_size: 32, dropout: 0.264, activation: relu, seq\_len: 512 \\
    TSMixer & CD-336 & lr: 0.0002, num\_blocks: 5, hidden\_size: 64, dropout: 0.287, activation: relu, seq\_len: 512 \\
    CrossFormer & CI-336 & lr: 0.0002, e\_layers: 7, d\_ff: 256, d\_model: 256, dropout: 0.309, seg\_len: 8, baseline: 0, cross\_factor: 5, seq\_len: 336 \\
    CrossFormer & CD-336 & lr: 0.0002, e\_layers: 7, d\_ff: 256, d\_model: 256, dropout: 0.309, seg\_len: 8, baseline: 0, cross\_factor: 18, seq\_len: 336 \\
    DLinear & CI-336 & lr: 0.0004, seq\_len: 720 \\
    iTransformers & CD-336 & lr: 0.0006, e\_layers: 9, d\_ff: 2048, d\_model: 128, dropout: 0.286, seq\_len: 720 \\
    TimeMixer & CI-336 & lr: 0.0028, d\_ff: 256, d\_model: 128, e\_layers: 4, seq\_len: 192 \\
    TimeMixer & CD-336 & lr: 0.0010, d\_ff: 1024, d\_model: 1024, e\_layers: 10, seq\_len: 192 \\
    \midrule
    PatchTST & CI-720 & lr: 1.44e-05, e\_layers: 10, d\_ff: 256, d\_model: 1024, dropout: 0.014, fc\_dropout: 0.539, patch\_size: 16, stride: 8, seq\_len: 192 \\
    PatchTST & CD-720 & lr: 0.0003, e\_layers: 9, d\_ff: 256, d\_model: 128, dropout: 0.228, fc\_dropout: 0.072, patch\_size: 8, stride: 8, seq\_len: 336 \\
    TSMixer & CI-720 & lr: 0.0004, num\_blocks: 8, hidden\_size: 1024, dropout: 0.754, activation: relu, seq\_len: 512 \\
    TSMixer & CD-720 & lr: 0.0001, num\_blocks: 2, hidden\_size: 32, dropout: 0.297, activation: relu, seq\_len: 512 \\
    CrossFormer & CI-720 & lr: 0.0002, e\_layers: 7, d\_ff: 512, d\_model: 256, dropout: 0.317, seg\_len: 7, baseline: 1, cross\_factor: 17, seq\_len: 96 \\
    CrossFormer & CD-720 & lr: 2.20e-05, e\_layers: 6, d\_ff: 256, d\_model: 512, dropout: 0.085, seg\_len: 12, baseline: 1, cross\_factor: 11, seq\_len: 96 \\
    DLinear & CI-720 & lr: 0.0042, seq\_len: 512 \\
    iTransformers & CD-720 & lr: 0.0001, e\_layers: 8, d\_ff: 2048, d\_model: 1024, dropout: 0.356, seq\_len: 192 \\
    TimeMixer & CI-720 & lr: 2.30e-05, d\_ff: 512, d\_model: 128, e\_layers: 1, seq\_len: 512 \\
    TimeMixer & CD-720 & lr: 0.0021, d\_ff: 256, d\_model: 128, e\_layers: 4, seq\_len: 192 \\
\end{xltabular}

\begin{xltabular}{\textwidth}{l l X}
    \caption{Hyperparameter settings for the \textbf{CellCycle} dataset. The table is split across pages for readability.} \label{tab:hyperparams_cellcycle} \\
    \toprule
    \textbf{Model} & \textbf{Setting} & \textbf{Best Hyperparameters} \\
    \midrule
    \endfirsthead
    
    \multicolumn{3}{c}{{\bfseries \tablename\ \thetable{} -- continued from previous page}} \\
    \toprule
    \textbf{Model} & \textbf{Setting} & \textbf{Best Hyperparameters} \\
    \midrule
    \endhead
    PatchTST & CI-96 & lr: 0.0003, e\_layers: 10, d\_ff: 256, d\_model: 128, dropout: 0.081, fc\_dropout: 0.390, patch\_size: 8, stride: 8, seq\_len: 336 \\
    PatchTST & CD-96 & lr: 0.0003, e\_layers: 9, d\_ff: 256, d\_model: 128, dropout: 0.228, fc\_dropout: 0.072, patch\_size: 8, stride: 8, seq\_len: 336 \\
    TSMixer & CI-96 & lr: 0.0055, num\_blocks: 10, hidden\_size: 256, dropout: 0.197, activation: gelu, seq\_len: 512 \\
    TSMixer & CD-96 & lr: 0.0041, num\_blocks: 10, hidden\_size: 256, dropout: 0.415, activation: gelu, seq\_len: 96 \\
    CrossFormer & CI-96 & lr: 0.0002, e\_layers: 7, d\_ff: 256, d\_model: 512, dropout: 0.248, seg\_len: 7, baseline: 0, cross\_factor: 5, seq\_len: 336 \\
    CrossFormer & CD-96 & lr: 0.0002, e\_layers: 7, d\_ff: 256, d\_model: 512, dropout: 0.248, seg\_len: 7, baseline: 0, cross\_factor: 5, seq\_len: 336 \\
    DLinear & CI-96 & lr: 0.0012, seq\_len: 336 \\
    iTransformers & CD-96 & lr: 0.0001, e\_layers: 8, d\_ff: 2048, d\_model: 1024, dropout: 0.356, seq\_len: 192 \\
    TimeMixer & CI-96 & lr: 0.0027, d\_ff: 256, d\_model: 256, e\_layers: 8, seq\_len: 96 \\
    TimeMixer & CD-96 & lr: 0.0017, d\_ff: 256, d\_model: 512, e\_layers: 4, seq\_len: 96 \\
    \midrule

    PatchTST & CI-192 & lr: 0.0003, e\_layers: 10, d\_ff: 256, d\_model: 128, dropout: 0.081, fc\_dropout: 0.390, patch\_size: 8, stride: 8, seq\_len: 336 \\
    PatchTST & CD-192 & lr: 0.0003, e\_layers: 9, d\_ff: 256, d\_model: 128, dropout: 0.228, fc\_dropout: 0.072, patch\_size: 8, stride: 8, seq\_len: 336 \\
    TSMixer & CI-192 & lr: 0.0041, num\_blocks: 10, hidden\_size: 256, dropout: 0.415, activation: gelu, seq\_len: 96 \\
    TSMixer & CD-192 & lr: 0.0041, num\_blocks: 10, hidden\_size: 256, dropout: 0.415, activation: gelu, seq\_len: 96 \\
    CrossFormer & CI-192 & lr: 0.0002, e\_layers: 7, d\_ff: 256, d\_model: 512, dropout: 0.248, seg\_len: 7, baseline: 0, cross\_factor: 5, seq\_len: 336 \\
    CrossFormer & CD-192 & lr: 0.0015, e\_layers: 5, d\_ff: 512, d\_model: 128, dropout: 0.228, seg\_len: 12, baseline: 1, cross\_factor: 17, seq\_len: 336 \\
    DLinear & CI-192 & lr: 0.0066, seq\_len: 720 \\
    iTransformers & CD-192 & lr: 1.30e-05, e\_layers: 4, d\_ff: 512, d\_model: 1024, dropout: 0.440, seq\_len: 336 \\
    TimeMixer & CI-192 & lr: 0.0009, d\_ff: 1024, d\_model: 1024, e\_layers: 10, seq\_len: 192 \\
    TimeMixer & CD-192 & lr: 0.0021, d\_ff: 256, d\_model: 128, e\_layers: 4, seq\_len: 192 \\
    \midrule

    PatchTST & CI-336 & lr: 0.00027, e\_layers: 10, d\_ff: 256, d\_model: 128, dropout: 0.081, fc\_dropout: 0.390, patch\_size: 8, stride: 8, seq\_len: 336 \\
    PatchTST & CD-336 & lr: 1.44e-05, e\_layers: 10, d\_ff: 256, d\_model: 1024, dropout: 0.014, fc\_dropout: 0.539, patch\_size: 16, stride: 8, seq\_len: 192 \\
    TSMixer & CI-336 & lr: 0.0041, num\_blocks: 10, hidden\_size: 256, dropout: 0.415, activation: gelu, seq\_len: 96 \\
    TSMixer & CD-336 & lr: 0.0041, num\_blocks: 10, hidden\_size: 256, dropout: 0.415, activation: gelu, seq\_len: 96 \\
    CrossFormer & CI-336 & lr: 0.0006, e\_layers: 8, d\_ff: 512, d\_model: 128, dropout: 0.361, seg\_len: 6, baseline: 1, cross\_factor: 8, seq\_len: 512 \\
    CrossFormer & CD-336 & lr: 0.0015, e\_layers: 5, d\_ff: 512, d\_model: 128, dropout: 0.228, seg\_len: 12, baseline: 1, cross\_factor: 17, seq\_len: 336 \\
    DLinear & CI-336 & lr: 0.0049, seq\_len: 512 \\
    iTransformers & CD-336 & lr: 0.0004, e\_layers: 1, d\_ff: 2048, d\_model: 1024, dropout: 0.388, seq\_len: 192 \\
    TimeMixer & CI-336 & lr: 0.0020, d\_ff: 512, d\_model: 256, e\_layers: 7, seq\_len: 192 \\
    TimeMixer & CD-336 & lr: 0.0021, d\_ff: 512, d\_model: 1024, e\_layers: 8, seq\_len: 96 \\
    \midrule

    PatchTST & CI-720 & lr: 1.44e-05, e\_layers: 10, d\_ff: 256, d\_model: 1024, dropout: 0.014, fc\_dropout: 0.539, patch\_size: 16, stride: 8, seq\_len: 192 \\
    PatchTST & CD-720 & lr: 1.44e-05, e\_layers: 10, d\_ff: 256, d\_model: 1024, dropout: 0.014, fc\_dropout: 0.539, patch\_size: 16, stride: 8, seq\_len: 192 \\
    TSMixer & CI-720 & lr: 0.0006, num\_blocks: 8, hidden\_size: 64, dropout: 0.181, activation: relu, seq\_len: 192 \\
    TSMixer & CD-720 & lr: 0.0010, num\_blocks: 1, hidden\_size: 64, dropout: 0.033, activation: relu, seq\_len: 192 \\
    CrossFormer & CI-720 & lr: 0.0006, e\_layers: 7, d\_ff: 128, d\_model: 256, dropout: 0.309, seg\_len: 7, baseline: 1, cross\_factor: 18, seq\_len: 336 \\
    CrossFormer & CD-720 & lr: 0.0002, e\_layers: 6, d\_ff: 128, d\_model: 256, dropout: 0.302, seg\_len: 7, baseline: 1, cross\_factor: 5, seq\_len: 512 \\
    DLinear & CI-720 & lr: 0.0033, seq\_len: 336 \\
    iTransformers & CD-720 & lr: 0.0001, e\_layers: 8, d\_ff: 2048, d\_model: 1024, dropout: 0.356, seq\_len: 192 \\
    TimeMixer & CI-720 & lr: 0.0019, d\_ff: 512, d\_model: 512, e\_layers: 8, seq\_len: 96 \\
    TimeMixer & CD-720 & lr: 0.0022, d\_ff: 1024, d\_model: 512, e\_layers: 8, seq\_len: 96 \\
\end{xltabular}

\begin{xltabular}{\textwidth}{l l X}
    \caption{Hyperparameter settings for the \textbf{DoublePendulum} dataset. The table is split across pages for readability.} \label{tab:hyperparams_doublependulum} \\
    \toprule
    \textbf{Model} & \textbf{Setting} & \textbf{Best Hyperparameters} \\
    \midrule
    \endfirsthead
    
    \multicolumn{3}{c}{{\bfseries \tablename\ \thetable{} -- continued from previous page}} \\
    \toprule
    \textbf{Model} & \textbf{Setting} & \textbf{Best Hyperparameters} \\
    \midrule
    \endhead
    PatchTST & CI-96 & lr: 1.44e-05, e\_layers: 10, d\_ff: 256, d\_model: 1024, dropout: 0.014, fc\_dropout: 0.539, patch\_size: 16, stride: 8, seq\_len: 192 \\
    PatchTST & CD-96 & lr: 0.0003, e\_layers: 9, d\_ff: 256, d\_model: 128, dropout: 0.228, fc\_dropout: 0.072, patch\_size: 8, stride: 8, seq\_len: 336 \\
    TSMixer & CI-96 & lr: 0.0012, num\_blocks: 6, hidden\_size: 64, dropout: 0.302, activation: relu, seq\_len: 192 \\
    TSMixer & CD-96 & lr: 0.0010, num\_blocks: 1, hidden\_size: 64, dropout: 0.033, activation: relu, seq\_len: 192 \\
    CrossFormer & CI-96 & lr: 0.0015, e\_layers: 9, d\_ff: 512, d\_model: 128, dropout: 0.228, seg\_len: 6, baseline: 0, cross\_factor: 20, seq\_len: 336 \\
    CrossFormer & CD-96 & lr: 0.0003, e\_layers: 7, d\_ff: 256, d\_model: 512, dropout: 0.285, seg\_len: 7, baseline: 0, cross\_factor: 5, seq\_len: 336 \\
    DLinear & CI-96 & lr: 0.0080, seq\_len: 192 \\
    iTransformers & CD-96 & lr: 0.0001, e\_layers: 8, d\_ff: 2048, d\_model: 1024, dropout: 0.356, seq\_len: 192 \\
    TimeMixer & CI-96 & lr: 0.0026, d\_ff: 256, d\_model: 256, e\_layers: 8, seq\_len: 192 \\
    TimeMixer & CD-96 & lr: 0.0018, d\_ff: 1024, d\_model: 128, e\_layers: 6, seq\_len: 192 \\
    \midrule

    PatchTST & CI-192 & lr: 1.53e-05, e\_layers: 10, d\_ff: 256, d\_model: 1024, dropout: 0.112, fc\_dropout: 0.457, patch\_size: 16, stride: 4, seq\_len: 192 \\
    PatchTST & CD-192 & lr: 1.44e-05, e\_layers: 10, d\_ff: 256, d\_model: 1024, dropout: 0.014, fc\_dropout: 0.539, patch\_size: 16, stride: 8, seq\_len: 192 \\
    TSMixer & CI-192 & lr: 0.0012, num\_blocks: 4, hidden\_size: 64, dropout: 0.323, activation: relu, seq\_len: 192 \\
    TSMixer & CD-192 & lr: 0.0032, num\_blocks: 7, hidden\_size: 1024, dropout: 0.006, activation: relu, seq\_len: 192 \\
    CrossFormer & CI-192 & lr: 0.0006, e\_layers: 7, d\_ff: 128, d\_model: 256, dropout: 0.309, seg\_len: 7, baseline: 1, cross\_factor: 18, seq\_len: 336 \\
    CrossFormer & CD-192 & lr: 0.0009, e\_layers: 8, d\_ff: 512, d\_model: 128, dropout: 0.193, seg\_len: 9, baseline: 0, cross\_factor: 7, seq\_len: 336 \\
    DLinear & CI-192 & lr: 0.0097, seq\_len: 192 \\
    iTransformers & CD-192 & lr: 0.0002, e\_layers: 8, d\_ff: 2048, d\_model: 512, dropout: 0.114, seq\_len: 192 \\
    TimeMixer & CI-192 & lr: 0.0027, d\_ff: 256, d\_model: 256, e\_layers: 8, seq\_len: 96 \\
    TimeMixer & CD-192 & lr: 0.0006, d\_ff: 256, d\_model: 256, e\_layers: 7, seq\_len: 96 \\
    \midrule

    PatchTST & CI-336 & lr: 1.44e-05, e\_layers: 10, d\_ff: 256, d\_model: 1024, dropout: 0.014, fc\_dropout: 0.539, patch\_size: 16, stride: 8, seq\_len: 192 \\
    PatchTST & CD-336 & lr: 1.44e-05, e\_layers: 10, d\_ff: 256, d\_model: 1024, dropout: 0.014, fc\_dropout: 0.539, patch\_size: 16, stride: 8, seq\_len: 192 \\
    TSMixer & CI-336 & lr: 0.0012, num\_blocks: 8, hidden\_size: 1024, dropout: 0.307, activation: relu, seq\_len: 192 \\
    TSMixer & CD-336 & lr: 0.0010, num\_blocks: 1, hidden\_size: 64, dropout: 0.033, activation: relu, seq\_len: 192 \\
    CrossFormer & CI-336 & lr: 0.0005, e\_layers: 6, d\_ff: 512, d\_model: 256, dropout: 0.326, seg\_len: 7, baseline: 1, cross\_factor: 8, seq\_len: 512 \\
    CrossFormer & CD-336 & lr: 0.0002, e\_layers: 7, d\_ff: 512, d\_model: 128, dropout: 0.309, seg\_len: 11, baseline: 0, cross\_factor: 8, seq\_len: 336 \\
    DLinear & CI-336 & lr: 0.0011, seq\_len: 192 \\
    iTransformers & CD-336 & lr: 0.0001, e\_layers: 8, d\_ff: 2048, d\_model: 1024, dropout: 0.356, seq\_len: 192 \\
    TimeMixer & CI-336 & lr: 0.0022, d\_ff: 1024, d\_model: 512, e\_layers: 8, seq\_len: 96 \\
    TimeMixer & CD-336 & lr: 0.0019, d\_ff: 256, d\_model: 256, e\_layers: 8, seq\_len: 96 \\
    \midrule

    PatchTST & CI-720 & lr: 0.0001, e\_layers: 9, d\_ff: 512, d\_model: 512, dropout: 0.065, fc\_dropout: 0.539, patch\_size: 8, stride: 4, seq\_len: 96 \\
    PatchTST & CD-720 & lr: 0.0002, e\_layers: 1, d\_ff: 1024, d\_model: 512, dropout: 0.471, fc\_dropout: 0.710, patch\_size: 8, stride: 8, seq\_len: 336 \\
    TSMixer & CI-720 & lr: 0.0005, num\_blocks: 7, hidden\_size: 1024, dropout: 0.169, activation: relu, seq\_len: 192 \\
    TSMixer & CD-720 & lr: 0.0010, num\_blocks: 1, hidden\_size: 64, dropout: 0.033, activation: relu, seq\_len: 192 \\
    CrossFormer & CI-720 & lr: 0.0002, e\_layers: 6, d\_ff: 512, d\_model: 128, dropout: 0.266, seg\_len: 8, baseline: 1, cross\_factor: 5, seq\_len: 336 \\
    CrossFormer & CD-720 & lr: 0.0001, e\_layers: 3, d\_ff: 256, d\_model: 512, dropout: 0.123, seg\_len: 8, baseline: 0, cross\_factor: 7, seq\_len: 336 \\
    DLinear & CI-720 & lr: 0.0032, seq\_len: 512 \\
    iTransformers & CD-720 & lr: 8.00e-06, e\_layers: 2, d\_ff: 2048, d\_model: 1024, dropout: 0.012, seq\_len: 192 \\
    TimeMixer & CI-720 & lr: 0.0018, d\_ff: 512, d\_model: 512, e\_layers: 7, seq\_len: 96 \\
    TimeMixer & CD-720 & lr: 0.0020, d\_ff: 256, d\_model: 256, e\_layers: 8, seq\_len: 96 \\
\end{xltabular}

\begin{xltabular}{\textwidth}{l l X}
    \caption{Hyperparameter settings for the \textbf{ETTh1} dataset across different forecasting horizons.} \label{tab:hyperparams_etth1} \\
    \toprule
    \textbf{Model} & \textbf{Setting} & \textbf{Best Hyperparameters} \\
    \midrule
    \endfirsthead
    
    \multicolumn{3}{c}{{\bfseries \tablename\ \thetable{} -- continued from previous page}} \\
    \toprule
    \textbf{Model} & \textbf{Setting} & \textbf{Best Hyperparameters} \\
    \midrule
    \endhead
    PatchTST & CI-96 & lr: 0.0001, e\_layers: 1, d\_ff: 512, d\_model: 256, dropout: 0.572, fc\_dropout: 0.631, patch\_size: 16, stride: 4, seq\_len: 96 \\
    PatchTST & CD-96 & lr: 0.0001, e\_layers: 3, d\_ff: 512, d\_model: 512, dropout: 0.518, fc\_dropout: 0.802, patch\_size: 16, stride: 4, seq\_len: 96 \\
    TSMixer & CI-96 & lr: 0.0097, num\_blocks: 9, hidden\_size: 32, dropout: 0.748, activation: gelu, seq\_len: 336 \\
    TSMixer & CD-96 & lr: 0.0002, num\_blocks: 6, hidden\_size: 256, dropout: 0.139, activation: gelu, seq\_len: 96 \\
    CrossFormer & CI-96 & lr: 0.0002, e\_layers: 5, d\_ff: 512, d\_model: 256, dropout: 0.212, seg\_len: 8, baseline: 0, cross\_factor: 10, seq\_len: 96 \\
    CrossFormer & CD-96 & lr: 0.0080, e\_layers: 1, d\_ff: 128, d\_model: 128, dropout: 0.189, seg\_len: 8, baseline: 0, cross\_factor: 9, seq\_len: 192 \\
    DLinear & CI-96 & lr: 0.0094, seq\_len: 336 \\
    iTransformers & CD-96 & lr: 0.0007, e\_layers: 3, d\_ff: 1024, d\_model: 512, dropout: 0.539, seq\_len: 96 \\
    TimeMixer & CI-96 & lr: 0.0004, d\_ff: 256, d\_model: 256, e\_layers: 6, seq\_len: 192 \\
    TimeMixer & CD-96 & lr: 0.0002, d\_ff: 1024, d\_model: 512, e\_layers: 4, seq\_len: 512 \\
    \midrule

    PatchTST & CI-192 & lr: 7.27e-06, e\_layers: 4, d\_ff: 1024, d\_model: 512, dropout: 0.053, fc\_dropout: 0.183, patch\_size: 16, stride: 4, seq\_len: 192 \\
    PatchTST & CD-192 & lr: 0.0001, e\_layers: 3, d\_ff: 1024, d\_model: 128, dropout: 0.705, fc\_dropout: 0.640, patch\_size: 16, stride: 8, seq\_len: 512 \\
    TSMixer & CI-192 & lr: 0.0005, num\_blocks: 1, hidden\_size: 64, dropout: 0.173, activation: relu, seq\_len: 192 \\
    TSMixer & CD-192 & lr: 0.0003, num\_blocks: 1, hidden\_size: 64, dropout: 0.166, activation: relu, seq\_len: 192 \\
    CrossFormer & CI-192 & lr: 0.0080, e\_layers: 1, d\_ff: 128, d\_model: 128, dropout: 0.189, seg\_len: 8, baseline: 0, cross\_factor: 9, seq\_len: 192 \\
    CrossFormer & CD-192 & lr: 0.0080, e\_layers: 1, d\_ff: 128, d\_model: 128, dropout: 0.189, seg\_len: 8, baseline: 0, cross\_factor: 9, seq\_len: 192 \\
    DLinear & CI-192 & lr: 0.0010, seq\_len: 336 \\
    iTransformers & CD-192 & lr: 1.60e-05, e\_layers: 9, d\_ff: 2048, d\_model: 1024, dropout: 0.012, seq\_len: 336 \\
    TimeMixer & CI-192 & lr: 1.60e-05, d\_ff: 512, d\_model: 256, e\_layers: 3, seq\_len: 336 \\
    TimeMixer & CD-192 & lr: 0.0001, d\_ff: 1024, d\_model: 256, e\_layers: 9, seq\_len: 512 \\
    \midrule

    PatchTST & CI-336 & lr: 5.21e-05, e\_layers: 1, d\_ff: 512, d\_model: 256, dropout: 0.572, fc\_dropout: 0.631, patch\_size: 16, stride: 4, seq\_len: 96 \\
    PatchTST & CD-336 & lr: 2.92e-06, e\_layers: 6, d\_ff: 1024, d\_model: 512, dropout: 0.342, fc\_dropout: 0.404, patch\_size: 16, stride: 4, seq\_len: 192 \\
    TSMixer & CI-336 & lr: 0.0002, num\_blocks: 2, hidden\_size: 64, dropout: 0.205, activation: relu, seq\_len: 512 \\
    TSMixer & CD-336 & lr: 0.0002, num\_blocks: 1, hidden\_size: 64, dropout: 0.207, activation: relu, seq\_len: 192 \\
    CrossFormer & CI-336 & lr: 0.0080, e\_layers: 1, d\_ff: 128, d\_model: 128, dropout: 0.189, seg\_len: 8, baseline: 0, cross\_factor: 9, seq\_len: 192 \\
    CrossFormer & CD-336 & lr: 0.0080, e\_layers: 1, d\_ff: 128, d\_model: 128, dropout: 0.189, seg\_len: 8, baseline: 0, cross\_factor: 9, seq\_len: 192 \\
    DLinear & CI-336 & lr: 0.0011, seq\_len: 336 \\
    iTransformers & CD-336 & lr: 1.60e-05, e\_layers: 9, d\_ff: 2048, d\_model: 1024, dropout: 0.012, seq\_len: 192 \\
    TimeMixer & CI-336 & lr: 1.50e-05, d\_ff: 256, d\_model: 256, e\_layers: 2, seq\_len: 512 \\
    TimeMixer & CD-336 & lr: 0.0009, d\_ff: 1024, d\_model: 512, e\_layers: 8, seq\_len: 512 \\
    \midrule

    PatchTST & CI-720 & lr: 1.78e-06, e\_layers: 2, d\_ff: 512, d\_model: 512, dropout: 0.438, fc\_dropout: 0.405, patch\_size: 16, stride: 4, seq\_len: 336 \\
    PatchTST & CD-720 & lr: 0.0024, e\_layers: 2, d\_ff: 1024, d\_model: 512, dropout: 0.450, fc\_dropout: 0.374, patch\_size: 16, stride: 4, seq\_len: 512 \\
    TSMixer & CI-720 & lr: 0.0002, num\_blocks: 2, hidden\_size: 64, dropout: 0.205, activation: relu, seq\_len: 512 \\
    TSMixer & CD-720 & lr: 0.0003, num\_blocks: 1, hidden\_size: 64, dropout: 0.169, activation: relu, seq\_len: 512 \\
    CrossFormer & CI-720 & lr: 0.0080, e\_layers: 1, d\_ff: 128, d\_model: 128, dropout: 0.189, seg\_len: 8, baseline: 0, cross\_factor: 9, seq\_len: 192 \\
    CrossFormer & CD-720 & lr: 0.0080, e\_layers: 1, d\_ff: 128, d\_model: 128, dropout: 0.189, seg\_len: 8, baseline: 0, cross\_factor: 9, seq\_len: 192 \\
    DLinear & CI-720 & lr: 0.0078, seq\_len: 512 \\
    iTransformers & CD-720 & lr: 1.30e-05, e\_layers: 8, d\_ff: 512, d\_model: 512, dropout: 0.017, seq\_len: 336 \\
    TimeMixer & CI-720 & lr: 1.00e-05, d\_ff: 512, d\_model: 512, e\_layers: 1, seq\_len: 720 \\
    TimeMixer & CD-720 & lr: 0.0005, d\_ff: 512, d\_model: 256, e\_layers: 8, seq\_len: 512 \\
\end{xltabular}

\begin{xltabular}{\textwidth}{l l X}
    \caption{Hyperparameter settings for the \textbf{ETTh2} dataset across different forecasting horizons.} \label{tab:hyperparams_etth2} \\
    \toprule
    \textbf{Model} & \textbf{Setting} & \textbf{Best Hyperparameters} \\
    \midrule
    \endfirsthead
    
    \multicolumn{3}{c}{{\bfseries \tablename\ \thetable{} -- continued from previous page}} \\
    \toprule
    \textbf{Model} & \textbf{Setting} & \textbf{Best Hyperparameters} \\
    \midrule
    \endhead
    PatchTST & CI-96 & lr: 3.50e-05, e\_layers: 1, d\_ff: 512, d\_model: 256, dropout: 0.571, fc\_dropout: 0.486, patch\_size: 16, stride: 4, seq\_len: 96 \\
    PatchTST & CD-96 & lr: 0.0001, e\_layers: 3, d\_ff: 512, d\_model: 512, dropout: 0.688, fc\_dropout: 0.393, patch\_size: 16, stride: 4, seq\_len: 96 \\
    TSMixer & CI-96 & lr: 0.0005, num\_blocks: 1, hidden\_size: 64, dropout: 0.166, activation: relu, seq\_len: 192 \\
    TSMixer & CD-96 & lr: 0.0011, num\_blocks: 6, hidden\_size: 1024, dropout: 0.746, activation: gelu, seq\_len: 192 \\
    CrossFormer & CI-96 & lr: 0.0003, e\_layers: 5, d\_ff: 512, d\_model: 128, dropout: 0.707, seg\_len: 3, baseline: 1, cross\_factor: 10, seq\_len: 96 \\
    CrossFormer & CD-96 & lr: 0.0076, e\_layers: 2, d\_ff: 512, d\_model: 128, dropout: 0.494, seg\_len: 5, baseline: 1, cross\_factor: 13, seq\_len: 96 \\
    DLinear & CI-96 & lr: 0.0010, seq\_len: 336 \\
    iTransformers & CD-96 & lr: 0.0001, e\_layers: 3, d\_ff: 1024, d\_model: 512, dropout: 0.518, seq\_len: 96 \\
    TimeMixer & CI-96 & lr: 0.0001, d\_ff: 256, d\_model: 256, e\_layers: 3, seq\_len: 192 \\
    TimeMixer & CD-96 & lr: 0.0002, d\_ff: 256, d\_model: 256, e\_layers: 6, seq\_len: 192 \\
    \midrule

    PatchTST & CI-192 & lr: 0.0003, e\_layers: 1, d\_ff: 256, d\_model: 128, dropout: 0.553, fc\_dropout: 0.631, patch\_size: 8, stride: 4, seq\_len: 336 \\
    PatchTST & CD-192 & lr: 0.0002, e\_layers: 3, d\_ff: 512, d\_model: 512, dropout: 0.688, fc\_dropout: 0.594, patch\_size: 16, stride: 4, seq\_len: 96 \\
    TSMixer & CI-192 & lr: 0.0097, num\_blocks: 5, hidden\_size: 32, dropout: 0.114, activation: relu, seq\_len: 192 \\
    TSMixer & CD-192 & lr: 0.0001, num\_blocks: 2, hidden\_size: 64, dropout: 0.252, activation: relu, seq\_len: 192 \\
    CrossFormer & CI-192 & lr: 0.0076, e\_layers: 2, d\_ff: 512, d\_model: 128, dropout: 0.494, seg\_len: 5, baseline: 1, cross\_factor: 13, seq\_len: 96 \\
    CrossFormer & CD-192 & lr: 0.0076, e\_layers: 2, d\_ff: 512, d\_model: 128, dropout: 0.494, seg\_len: 5, baseline: 1, cross\_factor: 13, seq\_len: 96 \\
    DLinear & CI-192 & lr: 0.0034, seq\_len: 336 \\
    iTransformers & CD-192 & lr: 0.0001, e\_layers: 3, d\_ff: 1024, d\_model: 512, dropout: 0.518, seq\_len: 96 \\
    TimeMixer & CI-192 & lr: 0.0001, d\_ff: 256, d\_model: 256, e\_layers: 3, seq\_len: 192 \\
    TimeMixer & CD-192 & lr: 0.0004, d\_ff: 256, d\_model: 256, e\_layers: 4, seq\_len: 96 \\
    \midrule

    PatchTST & CI-336 & lr: 3.50e-05, e\_layers: 1, d\_ff: 512, d\_model: 256, dropout: 0.571, fc\_dropout: 0.486, patch\_size: 16, stride: 4, seq\_len: 96 \\
    PatchTST & CD-336 & lr: 0.0002, e\_layers: 3, d\_ff: 512, d\_model: 512, dropout: 0.688, fc\_dropout: 0.594, patch\_size: 16, stride: 4, seq\_len: 96 \\
    TSMixer & CI-336 & lr: 0.0097, num\_blocks: 5, hidden\_size: 32, dropout: 0.331, activation: relu, seq\_len: 192 \\
    TSMixer & CD-336 & lr: 0.0082, num\_blocks: 1, hidden\_size: 256, dropout: 0.155, activation: relu, seq\_len: 512 \\
    CrossFormer & CI-336 & lr: 0.0017, e\_layers: 2, d\_ff: 256, d\_model: 512, dropout: 0.512, seg\_len: 4, baseline: 1, cross\_factor: 8, seq\_len: 96 \\
    CrossFormer & CD-336 & lr: 0.0076, e\_layers: 2, d\_ff: 512, d\_model: 128, dropout: 0.494, seg\_len: 5, baseline: 1, cross\_factor: 13, seq\_len: 96 \\
    DLinear & CI-336 & lr: 0.0036, seq\_len: 192 \\
    iTransformers & CD-336 & lr: 0.0001, e\_layers: 3, d\_ff: 1024, d\_model: 512, dropout: 0.518, seq\_len: 96 \\
    TimeMixer & CI-336 & lr: 0.0072, d\_ff: 256, d\_model: 256, e\_layers: 8, seq\_len: 96 \\
    TimeMixer & CD-336 & lr: 0.0011, d\_ff: 256, d\_model: 256, e\_layers: 4, seq\_len: 96 \\
    \midrule

    PatchTST & CI-720 & lr: 0.0002, e\_layers: 3, d\_ff: 512, d\_model: 512, dropout: 0.714, fc\_dropout: 0.650, patch\_size: 16, stride: 4, seq\_len: 96 \\
    PatchTST & CD-720 & lr: 0.0001, e\_layers: 3, d\_ff: 512, d\_model: 512, dropout: 0.518, fc\_dropout: 0.802, patch\_size: 16, stride: 4, seq\_len: 96 \\
    TSMixer & CI-720 & lr: 0.0021, num\_blocks: 6, hidden\_size: 1024, dropout: 0.141, activation: relu, seq\_len: 192 \\
    TSMixer & CD-720 & lr: 0.0010, num\_blocks: 1, hidden\_size: 64, dropout: 0.033, activation: relu, seq\_len: 192 \\
    CrossFormer & CI-720 & lr: 0.0083, e\_layers: 2, d\_ff: 512, d\_model: 128, dropout: 0.646, seg\_len: 6, baseline: 1, cross\_factor: 17, seq\_len: 96 \\
    CrossFormer & CD-720 & lr: 0.0076, e\_layers: 2, d\_ff: 512, d\_model: 128, dropout: 0.494, seg\_len: 5, baseline: 1, cross\_factor: 13, seq\_len: 96 \\
    DLinear & CI-720 & lr: 0.0093, seq\_len: 96 \\
    iTransformers & CD-720 & lr: 0.0001, e\_layers: 3, d\_ff: 1024, d\_model: 512, dropout: 0.518, seq\_len: 96 \\
    TimeMixer & CI-720 & lr: 0.0001, d\_ff: 512, d\_model: 128, e\_layers: 3, seq\_len: 192 \\
    TimeMixer & CD-720 & lr: 0.0033, d\_ff: 512, d\_model: 128, e\_layers: 8, seq\_len: 96 \\
\end{xltabular}

\begin{xltabular}{\textwidth}{l l X}
    \caption{Hyperparameter settings for the \textbf{ETTm1} dataset across different forecasting horizons.} \label{tab:hyperparams_ettm1} \\
    \toprule
    \textbf{Model} & \textbf{Setting} & \textbf{Best Hyperparameters} \\
    \midrule
    \endfirsthead
    
    \multicolumn{3}{c}{{\bfseries \tablename\ \thetable{} -- continued from previous page}} \\
    \toprule
    \textbf{Model} & \textbf{Setting} & \textbf{Best Hyperparameters} \\
    \midrule
    \endhead
    PatchTST & CI-96 & lr: 0.0003, e\_layers: 9, d\_ff: 256, d\_model: 128, dropout: 0.228, fc\_dropout: 0.072, patch\_size: 8, stride: 8, seq\_len: 336 \\
    PatchTST & CD-96 & lr: 6.07e-06, e\_layers: 3, d\_ff: 256, d\_model: 1024, dropout: 0.006, fc\_dropout: 0.642, patch\_size: 16, stride: 8, seq\_len: 192 \\
    TSMixer & CI-96 & lr: 0.0097, num\_blocks: 2, hidden\_size: 1024, dropout: 0.115, activation: relu, seq\_len: 192 \\
    TSMixer & CD-96 & lr: 0.0084, num\_blocks: 2, hidden\_size: 32, dropout: 0.264, activation: relu, seq\_len: 512 \\
    CrossFormer & CI-96 & lr: 0.0002, e\_layers: 6, d\_ff: 512, d\_model: 128, dropout: 0.266, seg\_len: 8, baseline: 1, cross\_factor: 5, seq\_len: 336 \\
    CrossFormer & CD-96 & lr: 0.0001, e\_layers: 6, d\_ff: 512, d\_model: 256, dropout: 0.272, seg\_len: 7, baseline: 1, cross\_factor: 17, seq\_len: 96 \\
    DLinear & CI-96 & lr: 0.0006, seq\_len: 336 \\
    iTransformers & CD-96 & lr: 0.0005, e\_layers: 5, d\_ff: 512, d\_model: 128, dropout: 0.462, seq\_len: 192 \\
    TimeMixer & CI-96 & lr: 0.0004, d\_ff: 256, d\_model: 128, e\_layers: 2, seq\_len: 192 \\
    TimeMixer & CD-96 & lr: 0.0009, d\_ff: 512, d\_model: 512, e\_layers: 8, seq\_len: 192 \\
    \midrule

    PatchTST & CI-192 & lr: 2.97e-06, e\_layers: 10, d\_ff: 256, d\_model: 1024, dropout: 0.081, fc\_dropout: 0.435, patch\_size: 16, stride: 8, seq\_len: 192 \\
    PatchTST & CD-192 & lr: 2.97e-06, e\_layers: 10, d\_ff: 256, d\_model: 1024, dropout: 0.081, fc\_dropout: 0.435, patch\_size: 16, stride: 8, seq\_len: 192 \\
    TSMixer & CI-192 & lr: 0.0097, num\_blocks: 2, hidden\_size: 1024, dropout: 0.115, activation: relu, seq\_len: 192 \\
    TSMixer & CD-192 & lr: 0.0005, num\_blocks: 6, hidden\_size: 64, dropout: 0.280, activation: relu, seq\_len: 192 \\
    CrossFormer & CI-192 & lr: 0.0002, e\_layers: 7, d\_ff: 512, d\_model: 256, dropout: 0.317, seg\_len: 7, baseline: 1, cross\_factor: 17, seq\_len: 96 \\
    CrossFormer & CD-192 & lr: 0.0002, e\_layers: 7, d\_ff: 512, d\_model: 256, dropout: 0.317, seg\_len: 7, baseline: 1, cross\_factor: 17, seq\_len: 96 \\
    DLinear & CI-192 & lr: 0.0010, seq\_len: 336 \\
    iTransformers & CD-192 & lr: 1.30e-05, e\_layers: 4, d\_ff: 512, d\_model: 1024, dropout: 0.440, seq\_len: 336 \\
    TimeMixer & CI-192 & lr: 0.0002, d\_ff: 256, d\_model: 256, e\_layers: 4, seq\_len: 192 \\
    TimeMixer & CD-192 & lr: 0.0002, d\_ff: 512, d\_model: 1024, e\_layers: 7, seq\_len: 192 \\
    \midrule

    PatchTST & CI-336 & lr: 0.0003, e\_layers: 9, d\_ff: 256, d\_model: 128, dropout: 0.228, fc\_dropout: 0.0716, patch\_size: 8, stride: 8, seq\_len: 336 \\
    PatchTST & CD-336 & lr: 1.44e-05, e\_layers: 10, d\_ff: 256, d\_model: 1024, dropout: 0.0137, fc\_dropout: 0.5389, patch\_size: 16, stride: 8, seq\_len: 192 \\
    TSMixer & CI-336 & lr: 0.0006, num\_blocks: 8, hidden\_size: 1024, dropout: 0.166, activation: relu, seq\_len: 192 \\
    TSMixer & CD-336 & lr: 0.0004, num\_blocks: 8, hidden\_size: 64, dropout: 0.215, activation: relu, seq\_len: 192 \\
    CrossFormer & CI-336 & lr: 0.0003, e\_layers: 1, d\_ff: 128, d\_model: 128, dropout: 0.265, seg\_len: 8, baseline: 0, cross\_factor: 9, seq\_len: 336 \\
    CrossFormer & CD-336 & lr: 0.0080, e\_layers: 1, d\_ff: 128, d\_model: 128, dropout: 0.189, seg\_len: 8, baseline: 0, cross\_factor: 9, seq\_len: 192 \\
    DLinear & CI-336 & lr: 0.0007, seq\_len: 336 \\
    iTransformers & CD-336 & lr: 0.0004, e\_layers: 1, d\_ff: 2048, d\_model: 1024, dropout: 0.388, seq\_len: 192 \\
    TimeMixer & CI-336 & lr: 0.0002, d\_ff: 256, d\_model: 256, e\_layers: 4, seq\_len: 192 \\
    TimeMixer & CD-336 & lr: 0.0021, d\_ff: 256, d\_model: 128, e\_layers: 4, seq\_len: 192 \\
    \midrule

    PatchTST & CI-720 & lr: 2.27e-06, e\_layers: 4, d\_ff: 1024, d\_model: 128, dropout: 0.198, fc\_dropout: 0.356, patch\_size: 16, stride: 8, seq\_len: 512 \\
    PatchTST & CD-720 & lr: 0.0019, e\_layers: 7, d\_ff: 1024, d\_model: 512, dropout: 0.471, fc\_dropout: 0.375, patch\_size: 8, stride: 8, seq\_len: 720 \\
    TSMixer & CI-720 & lr: 0.0097, num\_blocks: 2, hidden\_size: 1024, dropout: 0.115, activation: relu, seq\_len: 192 \\
    TSMixer & CD-720 & lr: 0.0011, num\_blocks: 6, hidden\_size: 1024, dropout: 0.339, activation: relu, seq\_len: 192 \\
    CrossFormer & CI-720 & lr: 0.0001, e\_layers: 6, d\_ff: 128, d\_model: 256, dropout: 0.272, seg\_len: 7, baseline: 1, cross\_factor: 20, seq\_len: 336 \\
    CrossFormer & CD-720 & lr: 0.0080, e\_layers: 1, d\_ff: 128, d\_model: 128, dropout: 0.189, seg\_len: 8, baseline: 0, cross\_factor: 9, seq\_len: 192 \\
    DLinear & CI-720 & lr: 0.0097, seq\_len: 192 \\
    iTransformers & CD-720 & lr: 0.0001, e\_layers: 8, d\_ff: 2048, d\_model: 128, dropout: 0.319, seq\_len: 720 \\
    TimeMixer & CI-720 & lr: 1.00e-05, d\_ff: 512, d\_model: 512, e\_layers: 1, seq\_len: 720 \\
    TimeMixer & CD-720 & lr: 0.0001, d\_ff: 1024, d\_model: 1024, e\_layers: 2, seq\_len: 192 \\
\end{xltabular}

\begin{xltabular}{\textwidth}{l l X}
    \caption{Hyperparameter settings for the \textbf{ETTm2} dataset across different forecasting horizons.} \label{tab:hyperparams_ettm2} \\
    \toprule
    \textbf{Model} & \textbf{Setting} & \textbf{Best Hyperparameters} \\
    \midrule
    \endfirsthead
    
    \multicolumn{3}{c}{{\bfseries \tablename\ \thetable{} -- continued from previous page}} \\
    \toprule
    \textbf{Model} & \textbf{Setting} & \textbf{Best Hyperparameters} \\
    \midrule
    \endhead
    PatchTST & CI-96 & lr: 0.0002, e\_layers: 7, d\_ff: 1024, d\_model: 1024, dropout: 0.648, fc\_dropout: 0.175, patch\_size: 8, stride: 4, seq\_len: 512 \\
    PatchTST & CD-96 & lr: 1.60e-06, e\_layers: 8, d\_ff: 256, d\_model: 1024, dropout: 0.307, fc\_dropout: 0.623, patch\_size: 8, stride: 4, seq\_len: 512 \\
    TSMixer & CI-96 & lr: 0.0097, num\_blocks: 5, hidden\_size: 32, dropout: 0.289, activation: relu, seq\_len: 512 \\
    TSMixer & CD-96 & lr: 0.0003, num\_blocks: 1, hidden\_size: 64, dropout: 0.077, activation: relu, seq\_len: 192 \\
    CrossFormer & CI-96 & lr: 0.0003, e\_layers: 6, d\_ff: 512, d\_model: 128, dropout: 0.298, seg\_len: 7, baseline: 1, cross\_factor: 5, seq\_len: 336 \\
    CrossFormer & CD-96 & lr: 0.0003, e\_layers: 4, d\_ff: 512, d\_model: 128, dropout: 0.382, seg\_len: 11, baseline: 1, cross\_factor: 11, seq\_len: 512 \\
    DLinear & CI-96 & lr: 0.0020, seq\_len: 720 \\
    iTransformers & CD-96 & lr: 1.20e-05, e\_layers: 2, d\_ff: 1024, d\_model: 512, dropout: 0.083, seq\_len: 512 \\
    TimeMixer & CI-96 & lr: 1.10e-05, d\_ff: 512, d\_model: 256, e\_layers: 4, seq\_len: 336 \\
    TimeMixer & CD-96 & lr: 5.00e-06, d\_ff: 512, d\_model: 512, e\_layers: 3, seq\_len: 720 \\
    \midrule

    PatchTST & CI-192 & lr: 7.02e-07, e\_layers: 2, d\_ff: 1024, d\_model: 512, dropout: 0.292, fc\_dropout: 0.673, patch\_size: 8, stride: 4, seq\_len: 336 \\
    PatchTST & CD-192 & lr: 1.82e-06, e\_layers: 4, d\_ff: 1024, d\_model: 1024, dropout: 0.471, fc\_dropout: 0.405, patch\_size: 16, stride: 8, seq\_len: 512 \\
    TSMixer & CI-192 & lr: 0.0010, num\_blocks: 1, hidden\_size: 64, dropout: 0.033, activation: relu, seq\_len: 192 \\
    TSMixer & CD-192 & lr: 0.0004, num\_blocks: 1, hidden\_size: 64, dropout: 0.077, activation: relu, seq\_len: 192 \\
    CrossFormer & CI-192 & lr: 0.0001, e\_layers: 3, d\_ff: 512, d\_model: 128, dropout: 0.180, seg\_len: 9, baseline: 1, cross\_factor: 14, seq\_len: 720 \\
    CrossFormer & CD-192 & lr: 0.0001, e\_layers: 3, d\_ff: 512, d\_model: 128, dropout: 0.180, seg\_len: 9, baseline: 1, cross\_factor: 14, seq\_len: 720 \\
    DLinear & CI-192 & lr: 0.0002, seq\_len: 512 \\
    iTransformers & CD-192 & lr: 1.40e-05, e\_layers: 1, d\_ff: 1024, d\_model: 512, dropout: 0.083, seq\_len: 512 \\
    TimeMixer & CI-192 & lr: 1.00e-05, d\_ff: 512, d\_model: 512, e\_layers: 1, seq\_len: 720 \\
    TimeMixer & CD-192 & lr: 0.0003, d\_ff: 512, d\_model: 128, e\_layers: 4, seq\_len: 720 \\
    \midrule

    PatchTST & CI-336 & lr: 0.0002, e\_layers: 7, d\_ff: 1024, d\_model: 1024, dropout: 0.6482, fc\_dropout: 0.1748, patch\_size: 8, stride: 4, seq\_len: 512 \\
    PatchTST & CD-336 & lr: 1.60e-06, e\_layers: 8, d\_ff: 256, d\_model: 1024, dropout: 0.307, fc\_dropout: 0.6229, patch\_size: 8, stride: 4, seq\_len: 512 \\
    TSMixer & CI-336 & lr: 0.0097, num\_blocks: 5, hidden\_size: 32, dropout: 0.328, activation: relu, seq\_len: 512 \\
    TSMixer & CD-336 & lr: 0.0087, num\_blocks: 6, hidden\_size: 64, dropout: 0.759, activation: gelu, seq\_len: 512 \\
    CrossFormer & CI-336 & lr: 0.0026, e\_layers: 1, d\_ff: 512, d\_model: 128, dropout: 0.897, seg\_len: 6, baseline: 1, cross\_factor: 5, seq\_len: 336 \\
    CrossFormer & CD-336 & lr: 0.0001, e\_layers: 6, d\_ff: 512, d\_model: 256, dropout: 0.679, seg\_len: 10, baseline: 1, cross\_factor: 17, seq\_len: 720 \\
    DLinear & CI-336 & lr: 0.0038, seq\_len: 720 \\
    iTransformers & CD-336 & lr: 8.00e-06, e\_layers: 6, d\_ff: 2048, d\_model: 128, dropout: 0.184, seq\_len: 512 \\
    TimeMixer & CI-336 & lr: 1.00e-05, d\_ff: 512, d\_model: 512, e\_layers: 1, seq\_len: 720 \\
    TimeMixer & CD-336 & lr: 1.00e-05, d\_ff: 512, d\_model: 128, e\_layers: 1, seq\_len: 720 \\
    \midrule

    PatchTST & CI-720 & lr: 2.27e-06, e\_layers: 4, d\_ff: 1024, d\_model: 128, dropout: 0.198, fc\_dropout: 0.356, patch\_size: 16, stride: 8, seq\_len: 512 \\
    PatchTST & CD-720 & lr: 0.0032, e\_layers: 3, d\_ff: 1024, d\_model: 128, dropout: 0.318, fc\_dropout: 0.402, patch\_size: 16, stride: 8, seq\_len: 512 \\
    TSMixer & CI-720 & lr: 0.0097, num\_blocks: 5, hidden\_size: 32, dropout: 0.289, activation: relu, seq\_len: 512 \\
    TSMixer & CD-720 & lr: 0.0086, num\_blocks: 9, hidden\_size: 32, dropout: 0.759, activation: relu, seq\_len: 512 \\
    CrossFormer & CI-720 & lr: 2.75e-05, e\_layers: 8, d\_ff: 512, d\_model: 256, dropout: 0.895, seg\_len: 10, baseline: 1, cross\_factor: 12, seq\_len: 720 \\
    CrossFormer & CD-720 & lr: 0.0028, e\_layers: 8, d\_ff: 512, d\_model: 256, dropout: 0.664, seg\_len: 10, baseline: 0, cross\_factor: 16, seq\_len: 512 \\
    DLinear & CI-720 & lr: 0.0022, seq\_len: 336 \\
    iTransformers & CD-720 & lr: 9.00e-06, e\_layers: 2, d\_ff: 2048, d\_model: 128, dropout: 0.280, seq\_len: 512 \\
    TimeMixer & CI-720 & lr: 1.40e-05, d\_ff: 512, d\_model: 256, e\_layers: 4, seq\_len: 336 \\
    TimeMixer & CD-720 & lr: 1.00e-05, d\_ff: 512, d\_model: 128, e\_layers: 1, seq\_len: 720 \\
\end{xltabular}

\begin{xltabular}{\textwidth}{l l X}
    \caption{Hyperparameter settings for the \textbf{Hopfield} dataset across different forecasting horizons.} \label{tab:hyperparams_hopfield} \\
    \toprule
    \textbf{Model} & \textbf{Setting} & \textbf{Best Hyperparameters} \\
    \midrule
    \endfirsthead
    
    \multicolumn{3}{c}{{\bfseries \tablename\ \thetable{} -- continued from previous page}} \\
    \toprule
    \textbf{Model} & \textbf{Setting} & \textbf{Best Hyperparameters} \\
    \midrule
    \endhead
    PatchTST & CI-96 & lr: 1.44e-05, e\_layers: 10, d\_ff: 256, d\_model: 1024, dropout: 0.014, fc\_dropout: 0.539, patch\_size: 16, stride: 8, seq\_len: 192 \\
    PatchTST & CD-96 & lr: 0.0003, e\_layers: 10, d\_ff: 256, d\_model: 128, dropout: 0.081, fc\_dropout: 0.390, patch\_size: 8, stride: 8, seq\_len: 336 \\
    TSMixer & CI-96 & lr: 0.0002, num\_blocks: 6, hidden\_size: 256, dropout: 0.334, activation: relu, seq\_len: 512 \\
    TSMixer & CD-96 & lr: 0.0097, num\_blocks: 5, hidden\_size: 32, dropout: 0.311, activation: relu, seq\_len: 512 \\
    CrossFormer & CI-96 & lr: 0.0003, e\_layers: 7, d\_ff: 256, d\_model: 512, dropout: 0.336, seg\_len: 6, baseline: 0, cross\_factor: 17, seq\_len: 512 \\
    CrossFormer & CD-96 & lr: 0.0001, e\_layers: 3, d\_ff: 256, d\_model: 512, dropout: 0.123, seg\_len: 8, baseline: 0, cross\_factor: 7, seq\_len: 336 \\
    DLinear & CI-96 & lr: 0.0078, seq\_len: 720 \\
    iTransformers & CD-96 & lr: 0.0002, e\_layers: 8, d\_ff: 2048, d\_model: 512, dropout: 0.114, seq\_len: 192 \\
    TimeMixer & CI-96 & lr: 0.0015, d\_ff: 256, d\_model: 256, e\_layers: 7, seq\_len: 192 \\
    TimeMixer & CD-96 & lr: 0.0023, d\_ff: 256, d\_model: 128, e\_layers: 5, seq\_len: 192 \\
    \midrule

    PatchTST & CI-192 & lr: 0.0003, e\_layers: 10, d\_ff: 256, d\_model: 128, dropout: 0.081, fc\_dropout: 0.390, patch\_size: 8, stride: 8, seq\_len: 336 \\
    PatchTST & CD-192 & lr: 0.0003, e\_layers: 10, d\_ff: 256, d\_model: 128, dropout: 0.081, fc\_dropout: 0.390, patch\_size: 8, stride: 8, seq\_len: 336 \\
    TSMixer & CI-192 & lr: 0.0002, num\_blocks: 8, hidden\_size: 64, dropout: 0.295, activation: relu, seq\_len: 512 \\
    TSMixer & CD-192 & lr: 0.0010, num\_blocks: 1, hidden\_size: 64, dropout: 0.033, activation: relu, seq\_len: 192 \\
    CrossFormer & CI-192 & lr: 0.0002, e\_layers: 7, d\_ff: 256, d\_model: 512, dropout: 0.248, seg\_len: 7, baseline: 0, cross\_factor: 20, seq\_len: 336 \\
    CrossFormer & CD-192 & lr: 0.0002, e\_layers: 7, d\_ff: 256, d\_model: 512, dropout: 0.248, seg\_len: 7, baseline: 0, cross\_factor: 20, seq\_len: 512 \\
    DLinear & CI-192 & lr: 0.0082, seq\_len: 512 \\
    iTransformers & CD-192 & lr: 0.0002, e\_layers: 8, d\_ff: 2048, d\_model: 512, dropout: 0.114, seq\_len: 192 \\
    TimeMixer & CI-192 & lr: 0.0023, d\_ff: 256, d\_model: 128, e\_layers: 5, seq\_len: 192 \\
    TimeMixer & CD-192 & lr: 0.0010, d\_ff: 1024, d\_model: 1024, e\_layers: 10, seq\_len: 192 \\
    \midrule

    PatchTST & CI-336 & lr: 1.44e-05, e\_layers: 10, d\_ff: 256, d\_model: 1024, dropout: 0.014, fc\_dropout: 0.539, patch\_size: 16, stride: 8, seq\_len: 192 \\
    PatchTST & CD-336 & lr: 0.0004, num\_blocks: 8, hidden\_size: 1024, dropout: 0.159, activation: relu, seq\_len: 512 \\
    TSMixer & CI-336 & lr: 0.0011, num\_blocks: 1, hidden\_size: 256, dropout: 0.322, activation: relu, seq\_len: 512 \\
    TSMixer & CD-336 & lr: 0.0011, num\_blocks: 1, hidden\_size: 256, dropout: 0.322, activation: relu, seq\_len: 512 \\
    CrossFormer & CI-336 & lr: 0.0006, e\_layers: 7, d\_ff: 128, d\_model: 256, dropout: 0.309, seg\_len: 7, baseline: 1, cross\_factor: 18, seq\_len: 336 \\
    CrossFormer & CD-336 & lr: 0.0011, e\_layers: 8, d\_ff: 512, d\_model: 128, dropout: 0.357, seg\_len: 7, baseline: 1, cross\_factor: 8, seq\_len: 512 \\
    DLinear & CI-336 & lr: 0.0002, seq\_len: 512 \\
    iTransformers & CD-336 & lr: 0.0001, e\_layers: 8, d\_ff: 2048, d\_model: 1024, dropout: 0.356, seq\_len: 192 \\
    TimeMixer & CI-336 & lr: 0.0041, d\_ff: 256, d\_model: 1024, e\_layers: 6, seq\_len: 96 \\
    TimeMixer & CD-336 & lr: 0.0018, d\_ff: 1024, d\_model: 128, e\_layers: 6, seq\_len: 192 \\
    \midrule

    PatchTST & CI-720 & lr: 0.0003, e\_layers: 10, d\_ff: 256, d\_model: 128, dropout: 0.081, fc\_dropout: 0.390, patch\_size: 8, stride: 8, seq\_len: 336 \\
    PatchTST & CD-720 & lr: 1.44e-05, e\_layers: 10, d\_ff: 256, d\_model: 1024, dropout: 0.014, fc\_dropout: 0.539, patch\_size: 16, stride: 8, seq\_len: 192 \\
    TSMixer & CI-720 & lr: 0.0002, num\_blocks: 8, hidden\_size: 1024, dropout: 0.205, activation: relu, seq\_len: 512 \\
    TSMixer & CD-720 & lr: 0.0010, num\_blocks: 1, hidden\_size: 64, dropout: 0.033, activation: relu, seq\_len: 192 \\
    CrossFormer & CI-720 & lr: 0.0001, e\_layers: 3, d\_ff: 256, d\_model: 512, dropout: 0.123, seg\_len: 8, baseline: 0, cross\_factor: 7, seq\_len: 336 \\
    CrossFormer & CD-720 & lr: 0.0002, e\_layers: 6, d\_ff: 512, d\_model: 128, dropout: 0.266, seg\_len: 8, baseline: 1, cross\_factor: 5, seq\_len: 336 \\
    DLinear & CI-720 & lr: 0.0034, seq\_len: 512 \\
    iTransformers & CD-720 & lr: 0.0001, e\_layers: 8, d\_ff: 2048, d\_model: 1024, dropout: 0.356, seq\_len: 192 \\
    TimeMixer & CI-720 & lr: 0.0021, d\_ff: 256, d\_model: 256, e\_layers: 8, seq\_len: 512 \\
    TimeMixer & CD-720 & lr: 0.0010, d\_ff: 1024, d\_model: 1024, e\_layers: 10, seq\_len: 192 \\
\end{xltabular}

\begin{xltabular}{\textwidth}{l l X}
    \caption{Hyperparameter settings for the \textbf{Lorenz} dataset across different forecasting horizons.} \label{tab:hyperparams_lorenz} \\
    \toprule
    \textbf{Model} & \textbf{Setting} & \textbf{Best Hyperparameters} \\
    \midrule
    \endfirsthead
    
    \multicolumn{3}{c}{{\bfseries \tablename\ \thetable{} -- continued from previous page}} \\
    \toprule
    \textbf{Model} & \textbf{Setting} & \textbf{Best Hyperparameters} \\
    \midrule
    \endhead
    PatchTST & CI-96 & lr: 0.0003, e\_layers: 9, d\_ff: 256, d\_model: 128, dropout: 0.228, fc\_dropout: 0.072, patch\_size: 8, stride: 8, seq\_len: 336 \\
    PatchTST & CD-96 & lr: 0.0003, e\_layers: 9, d\_ff: 256, d\_model: 128, dropout: 0.228, fc\_dropout: 0.072, patch\_size: 8, stride: 8, seq\_len: 336 \\
    TSMixer & CI-96 & lr: 0.0041, num\_blocks: 10, hidden\_size: 256, dropout: 0.415, activation: gelu, seq\_len: 96 \\
    TSMixer & CD-96 & lr: 0.0041, num\_blocks: 10, hidden\_size: 256, dropout: 0.415, activation: gelu, seq\_len: 96 \\
    CrossFormer & CI-96 & lr: 0.0002, e\_layers: 7, d\_ff: 256, d\_model: 512, dropout: 0.309, seg\_len: 3, baseline: 0, cross\_factor: 18, seq\_len: 336 \\
    CrossFormer & CD-96 & lr: 0.0004, e\_layers: 6, d\_ff: 512, d\_model: 256, dropout: 0.228, seg\_len: 7, baseline: 0, cross\_factor: 8, seq\_len: 336 \\
    DLinear & CI-96 & lr: 0.0093, seq\_len: 96 \\
    iTransformers & CD-96 & lr: 0.0001, e\_layers: 8, d\_ff: 2048, d\_model: 1024, dropout: 0.356, seq\_len: 192 \\
    TimeMixer & CI-96 & lr: 0.0028, d\_ff: 256, d\_model: 128, e\_layers: 4, seq\_len: 192 \\
    TimeMixer & CD-96 & lr: 0.0028, d\_ff: 256, d\_model: 128, e\_layers: 4, seq\_len: 192 \\
    \midrule

    PatchTST & CI-192 & lr: 1.44e-05, e\_layers: 10, d\_ff: 256, d\_model: 1024, dropout: 0.014, fc\_dropout: 0.539, patch\_size: 16, stride: 8, seq\_len: 192 \\
    PatchTST & CD-192 & lr: 1.44e-05, e\_layers: 10, d\_ff: 256, d\_model: 1024, dropout: 0.014, fc\_dropout: 0.539, patch\_size: 16, stride: 8, seq\_len: 192 \\
    TSMixer & CI-192 & lr: 0.0041, num\_blocks: 10, hidden\_size: 256, dropout: 0.415, activation: gelu, seq\_len: 96 \\
    TSMixer & CD-192 & lr: 0.0041, num\_blocks: 10, hidden\_size: 256, dropout: 0.415, activation: gelu, seq\_len: 96 \\
    CrossFormer & CI-192 & lr: 0.0004, e\_layers: 8, d\_ff: 256, d\_model: 256, dropout: 0.323, seg\_len: 10, baseline: 0, cross\_factor: 10, seq\_len: 336 \\
    CrossFormer & CD-192 & lr: 0.0002, e\_layers: 6, d\_ff: 128, d\_model: 256, dropout: 0.302, seg\_len: 7, baseline: 1, cross\_factor: 5, seq\_len: 336 \\
    DLinear & CI-192 & lr: 0.0093, seq\_len: 96 \\
    iTransformers & CD-192 & lr: 0.0011, e\_layers: 4, d\_ff: 2048, d\_model: 128, dropout: 0.255, seq\_len: 720 \\
    TimeMixer & CI-192 & lr: 0.0013, d\_ff: 512, d\_model: 1024, e\_layers: 8, seq\_len: 96 \\
    TimeMixer & CD-192 & lr: 0.0016, d\_ff: 256, d\_model: 128, e\_layers: 5, seq\_len: 192 \\
    \midrule

    PatchTST & CI-336 & lr: 0.0003, e\_layers: 9, d\_ff: 256, d\_model: 128, dropout: 0.2277, fc\_dropout: 0.0716, patch\_size: 8, stride: 8, seq\_len: 336 \\
    PatchTST & CD-336 & lr: 1.44e-05, e\_layers: 10, d\_ff: 256, d\_model: 1024, dropout: 0.0137, fc\_dropout: 0.5389, patch\_size: 16, stride: 8, seq\_len: 192 \\
    TSMixer & CI-336 & lr: 0.0041, num\_blocks: 10, hidden\_size: 256, dropout: 0.415, activation: gelu, seq\_len: 96 \\
    TSMixer & CD-336 & lr: 0.0080, num\_blocks: 7, hidden\_size: 64, dropout: 0.204, activation: relu, seq\_len: 192 \\
    CrossFormer & CI-336 & lr: 0.0003, e\_layers: 7, d\_ff: 256, d\_model: 512, dropout: 0.285, seg\_len: 7, baseline: 0, cross\_factor: 5, seq\_len: 336 \\
    CrossFormer & CD-336 & lr: 0.0003, e\_layers: 5, d\_ff: 512, d\_model: 256, dropout: 0.228, seg\_len: 6, baseline: 0, cross\_factor: 17, seq\_len: 336 \\
    DLinear & CI-336 & lr: 0.0093, seq\_len: 96 \\
    iTransformers & CD-336 & lr: 0.0011, e\_layers: 4, d\_ff: 2048, d\_model: 128, dropout: 0.255, seq\_len: 720 \\
    TimeMixer & CI-336 & lr: 0.0010, d\_ff: 1024, d\_model: 1024, e\_layers: 10, seq\_len: 192 \\
    TimeMixer & CD-336 & lr: 0.0009, d\_ff: 1024, d\_model: 256, e\_layers: 8, seq\_len: 192 \\
    \midrule

    PatchTST & CI-720 & lr: 0.0001, e\_layers: 3, d\_ff: 512, d\_model: 512, dropout: 0.518, fc\_dropout: 0.802, patch\_size: 16, stride: 4, seq\_len: 96 \\
    PatchTST & CD-720 & lr: 0.0001, e\_layers: 3, d\_ff: 512, d\_model: 512, dropout: 0.518, fc\_dropout: 0.802, patch\_size: 16, stride: 4, seq\_len: 96 \\
    TSMixer & CI-720 & lr: 0.0026, num\_blocks: 6, hidden\_size: 256, dropout: 0.126, activation: gelu, seq\_len: 96 \\
    TSMixer & CD-720 & lr: 0.0041, num\_blocks: 10, hidden\_size: 256, dropout: 0.415, activation: gelu, seq\_len: 96 \\
    CrossFormer & CI-720 & lr: 0.0006, e\_layers: 7, d\_ff: 128, d\_model: 256, dropout: 0.309, seg\_len: 7, baseline: 1, cross\_factor: 18, seq\_len: 336 \\
    CrossFormer & CD-720 & lr: 0.0006, e\_layers: 8, d\_ff: 512, d\_model: 128, dropout: 0.361, seg\_len: 6, baseline: 1, cross\_factor: 10, seq\_len: 336 \\
    DLinear & CI-720 & enc\_in: 3, lr: 0.0027, seq\_len: 96 \\
    iTransformers & CD-720 & lr: 0.0002, e\_layers: 7, d\_ff: 1024, d\_model: 128, dropout: 0.480, seq\_len: 720 \\
    TimeMixer & CI-720 & lr: 0.0010, d\_ff: 1024, d\_model: 1024, e\_layers: 10, seq\_len: 192 \\
    TimeMixer & CD-720 & lr: 0.0041, d\_ff: 256, d\_model: 1024, e\_layers: 6, seq\_len: 96 \\
\end{xltabular}

\begin{xltabular}{\textwidth}{l l X}
    \caption{Hyperparameter settings for the \textbf{LorenzCoupled} dataset.} \label{tab:hyperparams_lorenzcoupled} \\
    \toprule
    \textbf{Model} & \textbf{Setting} & \textbf{Best Hyperparameters} \\
    \midrule
    \endfirsthead
    
    \multicolumn{3}{c}{{\bfseries \tablename\ \thetable{} -- continued from previous page}} \\
    \toprule
    \textbf{Model} & \textbf{Setting} & \textbf{Best Hyperparameters} \\
    \midrule
    \endhead
    PatchTST & CI-96 & lr: 1.44e-05, e\_layers: 10, d\_ff: 256, d\_model: 1024, dropout: 0.014, fc\_dropout: 0.539, patch\_size: 16, stride: 8, seq\_len: 192 \\
    PatchTST & CD-96 & lr: 0.0003, e\_layers: 9, d\_ff: 256, d\_model: 128, dropout: 0.228, fc\_dropout: 0.072, patch\_size: 8, stride: 8, seq\_len: 336 \\
    TSMixer & CI-96 & lr: 0.0041, num\_blocks: 10, hidden\_size: 256, dropout: 0.415, activation: gelu, seq\_len: 96 \\
    TSMixer & CD-96 & lr: 0.0041, num\_blocks: 10, hidden\_size: 256, dropout: 0.415, activation: gelu, seq\_len: 96 \\
    CrossFormer & CI-96 & lr: 0.0003, e\_layers: 7, d\_ff: 256, d\_model: 512, dropout: 0.336, seg\_len: 10, baseline: 0, cross\_factor: 5, seq\_len: 336 \\
    CrossFormer & CD-96 & lr: 0.0004, e\_layers: 7, d\_ff: 512, d\_model: 256, dropout: 0.323, seg\_len: 6, baseline: 0, cross\_factor: 8, seq\_len: 336 \\
    DLinear & CI-96 & lr: 0.0097, seq\_len: 192 \\
    iTransformers & CD-96 & lr: 1.30e-05, e\_layers: 4, d\_ff: 2048, d\_model: 128, dropout: 0.008, seq\_len: 720 \\
    TimeMixer & CI-96 & lr: 0.0014, d\_ff: 1024, d\_model: 256, e\_layers: 8, seq\_len: 96 \\
    TimeMixer & CD-96 & lr: 0.0006, d\_ff: 1024, d\_model: 128, e\_layers: 5, seq\_len: 192 \\
    \midrule
    PatchTST & CI-192 & lr: 1.44e-05, e\_layers: 10, d\_ff: 256, d\_model: 1024, dropout: 0.014, fc\_dropout: 0.539, patch\_size: 16, stride: 8, seq\_len: 192 \\
    PatchTST & CD-192 & lr: 1.44e-05, e\_layers: 10, d\_ff: 256, d\_model: 1024, dropout: 0.014, fc\_dropout: 0.539, patch\_size: 16, stride: 8, seq\_len: 192 \\
    TSMixer & CI-192 & lr: 0.0041, num\_blocks: 10, hidden\_size: 256, dropout: 0.415, activation: gelu, seq\_len: 96 \\
    TSMixer & CD-192 & lr: 0.0097, num\_blocks: 6, hidden\_size: 256, dropout: 0.254, activation: relu, seq\_len: 512 \\
    CrossFormer & CI-192 & lr: 0.0003, e\_layers: 6, d\_ff: 256, d\_model: 512, dropout: 0.309, seg\_len: 8, baseline: 0, cross\_factor: 5, seq\_len: 336 \\
    CrossFormer & CD-192 & lr: 0.0003, e\_layers: 6, d\_ff: 512, d\_model: 128, dropout: 0.298, seg\_len: 7, baseline: 1, cross\_factor: 5, seq\_len: 336 \\
    DLinear & CI-192 & lr: 0.0080, seq\_len: 192 \\
    iTransformers & CD-192 & lr: 1.30e-05, e\_layers: 4, d\_ff: 2048, d\_model: 128, dropout: 0.008, seq\_len: 720 \\
    TimeMixer & CI-192 & lr: 0.0011, d\_ff: 256, d\_model: 256, e\_layers: 7, seq\_len: 96 \\
    TimeMixer & CD-192 & lr: 0.0024, d\_ff: 256, d\_model: 512, e\_layers: 6, seq\_len: 96 \\
    \midrule
    PatchTST & CI-336 & lr: 1.44e-05, e\_layers: 10, d\_ff: 256, d\_model: 1024, dropout: 0.0137, fc\_dropout: 0.539, patch\_size: 16, stride: 8, seq\_len: 192 \\
    PatchTST & CD-336 & lr: 1.44e-05, e\_layers: 10, d\_ff: 256, d\_model: 1024, dropout: 0.0137, fc\_dropout: 0.539, patch\_size: 16, stride: 8, seq\_len: 192 \\
    TSMixer & CI-336 & lr: 0.0041, num\_blocks: 10, hidden\_size: 256, dropout: 0.415, activation: gelu, seq\_len: 96 \\
    TSMixer & CD-336 & lr: 0.0010, num\_blocks: 1, hidden\_size: 64, dropout: 0.033, activation: relu, seq\_len: 192 \\
    CrossFormer & CI-336 & lr: 0.0001, e\_layers: 8, d\_ff: 256, d\_model: 256, dropout: 0.112, seg\_len: 10, baseline: 0, cross\_factor: 7, seq\_len: 336 \\
    CrossFormer & CD-336 & lr: 0.0002, e\_layers: 6, d\_ff: 512, d\_model: 128, dropout: 0.266, seg\_len: 8, baseline: 1, cross\_factor: 5, seq\_len: 336 \\
    DLinear & CI-336 & lr: 0.0093, seq\_len: 96 \\
    iTransformers & CD-336 & lr: 0.0003, e\_layers: 1, d\_ff: 512, d\_model: 512, dropout: 0.270, seq\_len: 720 \\
    TimeMixer & CI-336 & lr: 0.0015, d\_ff: 512, d\_model: 256, e\_layers: 8, seq\_len: 96 \\
    TimeMixer & CD-336 & lr: 0.0011, d\_ff: 256, d\_model: 256, e\_layers: 7, seq\_len: 96 \\
    \midrule
    PatchTST & CI-720 & lr: 0.0001, e\_layers: 2, d\_ff: 1024, d\_model: 1024, dropout: 0.463, fc\_dropout: 0.708, patch\_size: 16, stride: 4, seq\_len: 192 \\
    PatchTST & CD-720 & lr: 0.0018, e\_layers: 5, d\_ff: 1024, d\_model: 128, dropout: 0.097, fc\_dropout: 0.623, patch\_size: 16, stride: 8, seq\_len: 512 \\
    TSMixer & CI-720 & lr: 0.0005, num\_blocks: 6, hidden\_size: 64, dropout: 0.280, activation: relu, seq\_len: 192 \\
    TSMixer & CD-720 & lr: 0.0097, num\_blocks: 6, hidden\_size: 256, dropout: 0.254, activation: relu, seq\_len: 512 \\
    CrossFormer & CI-720 & lr: 0.0002, e\_layers: 6, d\_ff: 128, d\_model: 256, dropout: 0.302, seg\_len: 7, baseline: 1, cross\_factor: 5, seq\_len: 336 \\
    CrossFormer & CD-720 & lr: 0.0001, e\_layers: 3, d\_ff: 256, d\_model: 512, dropout: 0.123, seg\_len: 8, baseline: 0, cross\_factor: 7, seq\_len: 336 \\
    DLinear & CI-720 & lr: 0.0010, seq\_len: 96 \\
    iTransformers & CD-720 & lr: 0.0003, e\_layers: 5, d\_ff: 512, d\_model: 128, dropout: 0.305, seq\_len: 720 \\
    TimeMixer & CI-720 & lr: 0.0005, d\_ff: 256, d\_model: 256, e\_layers: 8, seq\_len: 96 \\
    TimeMixer & CD-720 & lr: 0.0041, d\_ff: 256, d\_model: 1024, e\_layers: 6, seq\_len: 96 \\
\end{xltabular}

\begin{xltabular}{\textwidth}{l l X}
    \caption{Hyperparameter settings for the \textbf{electricity} dataset.} \label{tab:hyperparams_electricity} \\
    \toprule
    \textbf{Model} & \textbf{Setting} & \textbf{Best Hyperparameters} \\
    \midrule
    \endfirsthead
    
    \multicolumn{3}{c}{{\bfseries \tablename\ \thetable{} -- continued from previous page}} \\
    \toprule
    \textbf{Model} & \textbf{Setting} & \textbf{Best Hyperparameters} \\
    \midrule
    \endhead
    PatchTST & CI-96 &  \\
    PatchTST & CD-96 &  \\
    TSMixer & CI-96 & lr: 0.0067, num\_blocks: 9, hidden\_size: 1024, dropout: 0.318, activation: relu, seq\_len: 512 \\
    TSMixer & CD-96 & lr: 0.0003, num\_blocks: 1, hidden\_size: 64, dropout: 0.166, activation: relu, seq\_len: 512 \\
    CrossFormer & CI-96 & lr: 0.0024, e\_layers: 1, d\_ff: 128, d\_model: 256, dropout: 0.305, seg\_len: 6, baseline: 0, cross\_factor: 7, seq\_len: 192 \\
    CrossFormer & CD-96 & lr: 0.0032, e\_layers: 1, d\_ff: 128, d\_model: 256, dropout: 0.294, seg\_len: 7, baseline: 0, cross\_factor: 8, seq\_len: 192 \\
    DLinear & CI-96 & lr: 0.0079, seq\_len: 512 \\
    iTransformers & CD-96 & lr: 0.0032, e\_layers: 4, d\_ff: 1024, d\_model: 128, dropout: 0.139, seq\_len: 720 \\
    TimeMixer & CI-96 & lr: 0.0010, d\_ff: 512, d\_model: 256, e\_layers: 1, seq\_len: 720 \\
    TimeMixer & CD-96 & lr: 0.0005, d\_ff: 512, d\_model: 256, e\_layers: 8, seq\_len: 512 \\
    \midrule
    PatchTST & CI-192 &  \\
    PatchTST & CD-192 &  \\
    TSMixer & CI-192 & lr: 0.0067, num\_blocks: 9, hidden\_size: 1024, dropout: 0.318, activation: relu, seq\_len: 512 \\
    TSMixer & CD-192 & lr: 0.0097, num\_blocks: 6, hidden\_size: 32, dropout: 0.315, activation: relu, seq\_len: 512 \\
    CrossFormer & CI-192 & OOM \\
    CrossFormer & CD-192 & lr: 0.0080, e\_layers: 1, d\_ff: 128, d\_model: 128, dropout: 0.189, seg\_len: 8, baseline: 0, cross\_factor: 9, seq\_len: 192 \\
    DLinear & CI-192 & lr: 0.0058, seq\_len: 512 \\
    iTransformers & CD-192 & lr: 0.0004, e\_layers: 1, d\_ff: 2048, d\_model: 1024, dropout: 0.080, seq\_len: 192 \\
    TimeMixer & CI-192 & lr: 0.00008, d\_ff: 512, d\_model: 512, e\_layers: 4, seq\_len: 192 \\
    TimeMixer & CD-192 & lr: 0.0015, d\_ff: 256, d\_model: 256, e\_layers: 7, seq\_len: 512 \\
    \midrule
    PatchTST & CI-336 & lr: 0.0067, num\_blocks: 9, hidden\_size: 1024, dropout: 0.318, activation: relu, seq\_len: 512 \\
    PatchTST & CD-336 &  \\
    TSMixer & CI-336 &  \\
    TSMixer & CD-336 & lr: 0.0097, num\_blocks: 5, hidden\_size: 32, dropout: 0.336, activation: relu, seq\_len: 512 \\
    CrossFormer & CI-336 & lr: 0.0080, e\_layers: 1, d\_ff: 128, d\_model: 128, dropout: 0.189, seg\_len: 8, baseline: 0, cross\_factor: 9, seq\_len: 192 \\
    CrossFormer & CD-336 & lr: 0.000176, e\_layers: 7, d\_ff: 128, d\_model: 512, dropout: 0.387, seg\_len: 10, baseline: 0, cross\_factor: 16, seq\_len: 192, affine: 1 \\
    DLinear & CI-336 & lr: 0.0098, seq\_len: 512 \\
    iTransformers & CD-336 & lr: 0.0003, e\_layers: 8, d\_ff: 2048, d\_model: 1024, dropout: 0.277, seq\_len: 192 \\
    TimeMixer & CI-336 & lr: 0.0001, d\_ff: 256, d\_model: 128, e\_layers: 2, seq\_len: 720 \\
    TimeMixer & CD-336 & lr: 0.0021, d\_ff: 256, d\_model: 128, e\_layers: 4, seq\_len: 192 \\
    \midrule
    PatchTST & CI-720 &  \\
    PatchTST & CD-720 &  \\
    TSMixer & CI-720 &  \\
    TSMixer & CD-720 & lr: 0.0090, num\_blocks: 5, hidden\_size: 64, dropout: 0.133, activation: relu, seq\_len: 192 \\
    CrossFormer & CI-720 & OOM \\
    CrossFormer & CD-720 & OOM \\
    DLinear & CI-720 & lr: 0.0098, seq\_len: 512 \\
    iTransformers & CD-720 & lr: 0.0012, e\_layers: 4, d\_ff: 2048, d\_model: 512, dropout: 0.360, seq\_len: 192 \\
    TimeMixer & CI-720 & lr: 0.0001, d\_ff: 256, d\_model: 512, e\_layers: 2, seq\_len: 720 \\
    TimeMixer & CD-720 & lr: 0.0021, d\_ff: 256, d\_model: 128, e\_layers: 4, seq\_len: 192 \\
\end{xltabular}

\begin{xltabular}{\textwidth}{l l X}
    \caption{Hyperparameter settings for the \textbf{weather} dataset.} \label{tab:hyperparams_weather} \\
    \toprule
    \textbf{Model} & \textbf{Setting} & \textbf{Best Hyperparameters} \\
    \midrule
    \endfirsthead
    
    \multicolumn{3}{c}{{\bfseries \tablename\ \thetable{} -- continued from previous page}} \\
    \toprule
    \textbf{Model} & \textbf{Setting} & \textbf{Best Hyperparameters} \\
    \midrule
    \endhead
    PatchTST & CI-96 & lr: 0.0003, e\_layers: 9, d\_ff: 256, d\_model: 128, dropout: 0.228, fc\_dropout: 0.072, patch\_size: 8, stride: 8, seq\_len: 336 \\
    PatchTST & CD-96 & lr: 0.0001, e\_layers: 1, d\_ff: 1024, d\_model: 128, dropout: 0.098, fc\_dropout: 0.465, patch\_size: 8, stride: 8, seq\_len: 336 \\
    TSMixer & CI-96 & lr: 0.0002, num\_blocks: 5, hidden\_size: 1024, dropout: 0.205, activation: relu, seq\_len: 512 \\
    TSMixer & CD-96 & lr: 0.0010, num\_blocks: 1, hidden\_size: 64, dropout: 0.033, activation: relu, seq\_len: 192 \\
    CrossFormer & CI-96 & lr: 1.43e-05, e\_layers: 5, d\_ff: 512, d\_model: 256, dropout: 0.098, seg\_len: 9, baseline: 0, cross\_factor: 11, seq\_len: 336 \\
    CrossFormer & CD-96 & lr: 0.0001, e\_layers: 3, d\_ff: 256, d\_model: 512, dropout: 0.123, seg\_len: 8, baseline: 0, cross\_factor: 7, seq\_len: 336 \\
    DLinear & CI-96 & lr: 0.0020, seq\_len: 720 \\
    iTransformers & CD-96 & lr: 0.0001, e\_layers: 8, d\_ff: 2048, d\_model: 1024, dropout: 0.356, seq\_len: 192 \\
    TimeMixer & CI-96 & lr: 6.00e-06, d\_ff: 512, d\_model: 256, e\_layers: 1, seq\_len: 720 \\
    TimeMixer & CD-96 & lr: 1.00e-05, d\_ff: 512, d\_model: 512, e\_layers: 1, seq\_len: 720 \\
    \midrule
    PatchTST & CI-192 & lr: 0.0001, e\_layers: 4, d\_ff: 256, d\_model: 128, dropout: 0.469, fc\_dropout: 0.405, patch\_size: 8, stride: 8, seq\_len: 512 \\
    PatchTST & CD-192 & lr: 0.0002, e\_layers: 1, d\_ff: 512, d\_model: 1024, dropout: 0.010, fc\_dropout: 0.333, patch\_size: 16, stride: 8, seq\_len: 192 \\
    TSMixer & CI-192 & lr: 0.0002, num\_blocks: 5, hidden\_size: 1024, dropout: 0.247, activation: relu, seq\_len: 512 \\
    TSMixer & CD-192 & lr: 0.0010, num\_blocks: 1, hidden\_size: 64, dropout: 0.033, activation: relu, seq\_len: 192 \\
    CrossFormer & CI-192 & lr: 0.0002, e\_layers: 6, d\_ff: 512, d\_model: 128, dropout: 0.266, seg\_len: 8, baseline: 1, cross\_factor: 5, seq\_len: 336 \\
    CrossFormer & CD-192 & lr: 0.0002, e\_layers: 1, d\_ff: 256, d\_model: 128, dropout: 0.326, seg\_len: 7, baseline: 0, cross\_factor: 9, seq\_len: 336 \\
    DLinear & CI-192 & lr: 0.0077, seq\_len: 720 \\
    iTransformers & CD-192 & lr: 1.60e-05, e\_layers: 9, d\_ff: 2048, d\_model: 1024, dropout: 0.012, seq\_len: 192 \\
    TimeMixer & CI-192 & lr: 0.00008, d\_ff: 512, d\_model: 512, e\_layers: 4, seq\_len: 192 \\
    TimeMixer & CD-192 & lr: 0.0005, d\_ff: 512, d\_model: 128, e\_layers: 8, seq\_len: 720 \\
    \midrule
    PatchTST & CI-336 & lr: 0.0007, e\_layers: 8, d\_ff: 1024, d\_model: 512, dropout: 0.150, fc\_dropout: 0.021, patch\_size: 8, stride: 8, seq\_len: 720 \\
    PatchTST & CD-336 & lr: 0.0007, e\_layers: 8, d\_ff: 1024, d\_model: 512, dropout: 0.150, fc\_dropout: 0.021, patch\_size: 8, stride: 8, seq\_len: 720 \\
    TSMixer & CI-336 & lr: 0.0002, num\_blocks: 5, hidden\_size: 1024, dropout: 0.265, activation: relu, seq\_len: 512 \\
    TSMixer & CD-336 & lr: 0.0001, num\_blocks: 6, hidden\_size: 256, dropout: 0.334, activation: gelu, seq\_len: 512 \\
    CrossFormer & CI-336 & lr: 0.0002, e\_layers: 6, d\_ff: 512, d\_model: 128, dropout: 0.266, seg\_len: 8, baseline: 1, cross\_factor: 5, seq\_len: 336 \\
    CrossFormer & CD-336 & OOM \\
    DLinear & CI-336 & lr: 0.0014, seq\_len: 720 \\
    iTransformers & CD-336 & lr: 1.60e-05, e\_layers: 9, d\_ff: 2048, d\_model: 1024, dropout: 0.012, seq\_len: 192 \\
    TimeMixer & CI-336 & lr: 0.0001, d\_ff: 256, d\_model: 128, e\_layers: 2, seq\_len: 720 \\
    TimeMixer & CD-336 & lr: 1.00e-05, d\_ff: 512, d\_model: 128, e\_layers: 1, seq\_len: 720 \\
    \midrule
    PatchTST & CI-720 & lr: 0.0005, e\_layers: 7, d\_ff: 256, d\_model: 256, dropout: 0.767, fc\_dropout: 0.878, patch\_size: 16, stride: 8, seq\_len: 720 \\
    PatchTST & CD-720 & lr: 6.07e-06, e\_layers: 3, d\_ff: 256, d\_model: 1024, dropout: 0.006, fc\_dropout: 0.642, patch\_size: 16, stride: 8, seq\_len: 192 \\
    TSMixer & CI-720 & lr: 0.0003, num\_blocks: 2, hidden\_size: 1024, dropout: 0.108, activation: relu, seq\_len: 512 \\
    TSMixer & CD-720 & lr: 0.0002, num\_blocks: 8, hidden\_size: 32, dropout: 0.281, activation: relu, seq\_len: 512 \\
    CrossFormer & CI-720 & lr: 0.0002, e\_layers: 1, d\_ff: 512, d\_model: 128, dropout: 0.228, seg\_len: 6, baseline: 0, cross\_factor: 17, seq\_len: 336 \\
    CrossFormer & CD-720 & lr: 0.0080, e\_layers: 1, d\_ff: 128, d\_model: 128, dropout: 0.189, seg\_len: 8, baseline: 0, cross\_factor: 9, seq\_len: 192 \\
    DLinear & CI-720 & lr: 0.0030, seq\_len: 720 \\
    iTransformers & CD-720 & lr: 0.0001, e\_layers: 8, d\_ff: 1024, d\_model: 1024, dropout: 0.356, seq\_len: 192 \\
    TimeMixer & CI-720 & lr: 0.0001, d\_ff: 256, d\_model: 512, e\_layers: 2, seq\_len: 720 \\
    TimeMixer & CD-720 & lr: 1.00e-05, d\_ff: 512, d\_model: 512, e\_layers: 1, seq\_len: 720 \\
\end{xltabular}

\begin{xltabular}{\textwidth}{l l X}
    \caption{Hyperparameter settings for the \textbf{Traffic} dataset.} \label{tab:hyperparams_traffic} \\
    \toprule
    \textbf{Model} & \textbf{Setting} & \textbf{Best Hyperparameters} \\
    \midrule
    \endfirsthead
    
    \multicolumn{3}{c}{{\bfseries \tablename\ \thetable{} -- continued from previous page}} \\
    \toprule
    \textbf{Model} & \textbf{Setting} & \textbf{Best Hyperparameters} \\
    \midrule
    \endhead
    PatchTST & CI-96 &  \\
    PatchTST & CD-96 &  \\
    TSMixer & CI-96 & lr: 0.0097, num\_blocks: 9, hidden\_size: 32, dropout: 0.271, activation: relu, seq\_len: 512 \\
    TSMixer & CD-96 & lr: 0.0097, num\_blocks: 5, hidden\_size: 32, dropout: 0.311, activation: relu, seq\_len: 512 \\
    CrossFormer & CI-96 & lr: 0.000516, e\_layers: 7, d\_ff: 128, d\_model: 256, dropout: 0.257, seg\_len: 11, baseline: 0, cross\_factor: 10, seq\_len: 96, affine: 1 \\
    CrossFormer & CD-96 & lr: 5.21e-05, e\_layers: 3, d\_ff: 256, d\_model: 512, dropout: 0.123, seg\_len: 8, baseline: 0, cross\_factor: 7, seq\_len: 336, affine: 0 \\
    DLinear & CI-96 & lr: 0.0078, seq\_len: 512 \\
    iTransformers & CD-96 & OOM \\
    TimeMixer & CI-96 & OOM \\
    TimeMixer & CD-96 & From TimeMixer Paper \\
    \midrule
    PatchTST & CI-192 &  \\
    PatchTST & CD-192 &  \\
    TSMixer & CI-192 & lr: 0.0067, num\_blocks: 9, hidden\_size: 1024, dropout: 0.318, activation: relu, seq\_len: 512 \\
    TSMixer & CD-192 & lr: 0.0097, num\_blocks: 5, hidden\_size: 32, dropout: 0.311, activation: relu, seq\_len: 512 \\
    CrossFormer & CI-192 & OOM \\
    CrossFormer & CD-192 & OOM \\
    DLinear & CI-192 & lr: 0.0078, seq\_len: 512 \\
    iTransformers & CD-192 & OOM \\
    TimeMixer & CI-192 & OOM \\
    TimeMixer & CD-192 & From TimeMixer Paper \\
    \midrule
    PatchTST & CI-336 &  \\
    PatchTST & CD-336 &  \\
    TSMixer & CI-336 & lr: 0.0067, num\_blocks: 9, hidden\_size: 1024, dropout: 0.318, activation: relu, seq\_len: 512 \\
    TSMixer & CD-336 & lr: 0.0097, num\_blocks: 5, hidden\_size: 32, dropout: 0.311, activation: relu, seq\_len: 512 \\
    CrossFormer & CI-336 & OOM \\
    CrossFormer & CD-336 & OOM \\
    DLinear & CI-336 & lr: 0.0038, seq\_len: 336 \\
    iTransformers & CD-336 & OOM \\
    TimeMixer & CI-336 & OOM \\
    TimeMixer & CD-336 & From TimeMixer Paper \\
    \midrule
    PatchTST & CI-720 &  \\
    PatchTST & CD-720 &  \\
    TSMixer & CI-720 &  \\
    TSMixer & CD-720 & lr: 0.0097, num\_blocks: 5, hidden\_size: 32, dropout: 0.311, activation: relu, seq\_len: 512 \\
    CrossFormer & CI-720 & OOM \\
    CrossFormer & CD-720 & OOM \\
    DLinear & CI-720 & lr: 0.0070, seq\_len: 336 \\
    iTransformers & CD-720 & OOM \\
    TimeMixer & CI-720 & OOM \\
    TimeMixer & CD-720 & From TimeMixer Paper \\
\end{xltabular}

\end{document}